\documentclass[pdflatex,sn-mathphys-num]{sn-jnl}

\usepackage{graphicx}%
\usepackage{multirow}%

\usepackage{amsmath,amssymb,amsfonts}%
\usepackage{amsthm}%
\usepackage{mathrsfs}%
\usepackage[title]{appendix}%
\usepackage{xcolor}%
\usepackage{textcomp}%
\usepackage{manyfoot}%
\usepackage{booktabs}%
\usepackage{algorithm}%
\usepackage{algorithmicx}%
\usepackage{algpseudocode}%
\usepackage{listings}%
\usepackage{caption}
\usepackage{multirow}
\usepackage{amsmath}
\usepackage{longtable}
\usepackage{hyperref}
\usepackage{graphicx}
\theoremstyle{thmstyleone}%
\theoremstyle{thmstyletwo}%

\theoremstyle{thmstylethree}%

\begin{document}

\title[Article Title]{FOLC-Net: A Federated-Optimized Lightweight Architecture for Enhanced MRI Disease Diagnosis across Axial, Coronal, and Sagittal Views}


\author*[1]{\fnm{Saif Ur Rehman} \sur{Khan}}\email{saif\_ur\_rehman.khan@dfki.de}

\author*[1,2]{\fnm{Muhammad Nabeel} \sur{Asim}}\email{muhammad\_nabeel.asim@dfki.de}
\author[1,2]{\fnm{Sebastian} \sur{ Vollmer}}\email{sebastian.vollmer@dfki.de}
\author[1,2,3]{\fnm{Andreas} \sur{Dengel}}\email{andreas.dengel@dfki.de}

\affil[1]{\orgdiv{German Research Center for Artificial Intelligence}, \orgaddress{ \city{Kaiserslautern}, \postcode{67663}, \country{Germany}}}
\affil[2]{\orgdiv{Intelligentx GmbH (intelligentx.com)}, \orgaddress{ \city{Kaiserslautern}, \country{Germany}}}
\affil[3]{\orgdiv{Department of Computer Science}, \orgname{Rhineland-Palatinate Technical University of Kaiserslautern-Landau} \orgaddress{ \city{Kaiserslautern}, \postcode{67663}, \country{Germany}}}


\abstract{
\textbf{Purpose:}
The framework is designed to improve performance in the analysis of combined as well as single anatomical perspectives for MRI disease diagnosis. It specifically addresses the performance degradation observed in state-of-the-art (SOTA) models, particularly when processing axial, coronal, and sagittal anatomical planes. The paper introduces the FOLC-Net framework, which incorporates a novel federated-optimized lightweight architecture with approximately 1.217 million parameters and a storage requirement of only 0.9 MB.

\textbf{Methods:}
FOLC-Net integrates Manta-ray foraging optimization (MRFO) mechanisms for efficient model structure generation, global model cloning for scalable training, and ConvNeXt for enhanced client adaptability. The model was evaluated on combined multi-view data as well as individual views, such as axial, coronal, and sagittal, to assess its robustness in various medical imaging scenarios. Moreover, FOLC-Net tests a ShallowFed model on different data to evaluate its ability to generalize beyond the training dataset.

\textbf{Results:}
The results show that FOLC-Net outperforms existing models, particularly in the challenging sagittal view. For instance, FOLC-Net achieved an accuracy of 92.44\% on the sagittal view, significantly higher than the 88.37\% accuracy of study method (DL + Residual Learning) and 88.95\% of DL models. Additionally, FOLC-Net demonstrated improved accuracy across all individual views, providing a more reliable and robust solution for medical image analysis in decentralized environments.

\textbf{Conclusion:}
FOLC-Net addresses the limitations of existing SOTA models by providing a framework that ensures better adaptability to individual views while maintaining strong performance in multi-view settings. The incorporation of MRFO, global model cloning, and ConvNeXt ensures that FOLC-Net performs better in real-world medical applications. }

\keywords{Global Synchronization, Cloning the Global , Federated knowledge , Comprehensive Disease Modality, Multi-View, Single-Views }



\maketitle

\section{Introduction}\label{sec1}
A tumor is an abnormal growth of tissue caused by the uncontrolled division and proliferation of cells \cite{sharma2025cancer}. When this uncontrolled growth of cells arises within the brain, it is referred to as a brain tumor. Brain tumors can originate from different cell types, such as neurons or glial cells, and are typically classified as benign tumor (non-cancerous) or malignant tumor (cancerous) \cite{biratu2021survey}. There are several common types of brain tumors including Gliomas, Meningiomas and pituitary \cite{salari2023global}. Gliomas are tumors that develop from glial cells and protect the nerve cells in the brain. Meningiomas are tumors that begin in the protective layers of tissue surrounding the brain. Pituitary tumors develop in the pituitary gland, which is a small organ at the bottom of the brain that controls various important functions in the body. These tumors arise in different parts of the brain and can cause various effects depending on their location. The symptoms of brain tumors vary significantly among patients and often include persistent seizures, cognitive impairment, visual or speech difficulties, and headaches issues \cite{park2022brain}. This variability in symptoms complicates the diagnostic process by highlighting the critical importance of precise and effective assessment techniques. The diagnosis of brain tumors typically relies on imaging techniques like magnetic resonance imaging (MRI) and computed tomography (CT) scans \cite{ali2022comprehensive}. Although these methods are commonly used and can be time-consuming due to the large volume of imaging data generated in clinical settings \cite{mouridsen2020artificial}. Analyzing such extensive data requires a significant amount of manual review, which can lead to variability in diagnostic accuracy depending on the experience of the radiologist conducting the analysis. Given these challenges, there is an urgent need for automated diagnostic tools that can speed up the diagnostic process to ensure more consistent tumor detection and improve the patient survival rate timely.

The advent of deep learning (DL) \cite{bilal2024differential, khan2025ensemble} technologies, especially Convolutional Neural Networks (CNNs) \cite{bilal2025amalgamation}, has revolutionized the field of medical image analysis among diverse types of disease. CNNs \cite{khan2025detection, khan2025leadcnn} are designed to automatically learn complex patterns from imaging data by resulting in significant classification improvements and faster diagnoses. These models have also been successfully applied to a wide range of medical imaging tasks for showing their ability to detect abnormalities that might be overlooked by traditional diagnostic methods. DL technologies have also significantly advanced medical image analysis by introducing transfer learning (TL) methods. TL \cite{hekmat2025brain, kumar2022hybrid} allows a model trained on one dataset to be fine-tuned for a new, related task, making it especially useful in medical imaging where annotated data is often limited. This method improves performance on new tasks by leveraging knowledge gained from large ImageNet dataset. However, while TL has proven valuable, it still faces challenges regarding data privacy and the need for regulatory compliance in clinical settings. In this context, federated learning (FL) \cite{chowdhury2021review, guan2024federated} has emerged as a powerful solution. By enabling models to be trained across decentralized devices without the need to transfer sensitive raw data, FL ensures patient confidentiality and helps to meet regulatory standards. This approach not only facilitates collaborative learning but also allows healthcare providers to develop advanced diagnostic tools that leverage insights from diverse data sources while maintaining the security of patient information. By integrating FL into the brain tumor diagnostics framework \cite{mahalool2022distributed}, we can address the limitations of conventional methods, thus both the accuracy and efficiency of diagnosing brain tumors can be improved.

\subsection{Problem formulation}
FL models are widely recognized for their ability to leverage decentralized data across heterogeneous clients, offering a promising solution for training models without the need to centralize sensitive data. However, the performance of FL models is highly sensitive to both the underlying model architecture and the methods used for training, particularly when applied multi-view to single-views representation of medical data in decentralized environments. Current methods, while effective in handling combined multi-view data, exhibit significant limitations when tested on individual views.

Let \( D = \{ D_1, D_2, \dots, D_n \} \) represent the set of decentralized datasets available at \( n \) heterogeneous clients. Each dataset \( D_i \) consists of data views such as axial, coronal, and sagittal views, denoted as \( D_i^{\text{axial}}, D_i^{\text{coronal}}, D_i^{\text{sagittal}} \). In conventional multi-view FL, a model \( f_{\theta} \) is trained on the aggregated feature set from all views:

\[
f_{\theta} = \arg \min_{\theta} \sum_{i=1}^{n} L(f_{\theta}(D_i), Y_i),
\tag{1} \]

where \( L \) is a loss function that compares the model predictions \( f_{\theta}(D_i) \) to the ground truth labels \( Y_i \). The goal is to optimize the model to handle the combined multi-view data for each client.

However, when the model is tested on individual views (axial, coronal, sagittal), significant performance degradation is observed. This issue arises from the fact that the model is typically optimized to capture global patterns across all views, neglecting the unique challenges presented by each individual view. The model performance on an individual view, e.g., the sagittal view, is often suboptimal compared to when all views are aggregated:

\[
f_{\theta}^{\text{sagittal}} = \arg \min_{\theta} L(f_{\theta}(D_i^{\text{sagittal}}), Y_i),
\tag{2} \]

where \( f_{\theta}^{\text{sagittal}} \) is the model performance on the sagittal view alone, and the accuracy tends to drop compared to the combined model performance. This issue stems from the previous proposed model failure to adapt to the specific characteristics of each view, resulting in a loss of accuracy and robustness.

Given that \( f_{\theta}^{\text{combined}} \) performs well on combined views but \( f_{\theta}^{\text{view}} \) underperforms on individual views, we aim to find a strategy to train a FL model that improves the performance on individual views without sacrificing the advantages of multi-view learning. Specifically, we seek to optimize the model in a way that it can:

\begin{enumerate}
    \item Effectively handle and adapt to the traits of each individual view \( D_i^{\text{axial}}, D_i^{\text{coronal}}, D_i^{\text{sagittal}} \),
    \item Maintain complexity or improve the overall performance on multi-view tasks while minimizing the drop in performance when tested on individual views.
\end{enumerate}

\[
\min_{\theta} L_{\text{view}}(f_{\theta}(D_i^{\text{view}}), Y_i)
\tag{3}  \]

where \( L_{\text{view}} \) represents the loss function for each individual view \( \text{view} \in \{\text{axial}, \text{coronal}, \text{sagittal}\} \), and the objective is to simultaneously optimize across all individual views without compromising the generalization to multi-view settings.

Main contribution of this work as follows:

\begin{itemize}
    \item \textbf{MRFO Mechanisms for Efficient Model Structure:} \\
    FOLC-Net integrates MRFO mechanisms, which optimize model structure generation by taking into account both global and local features. This mechanism allows the model to optimize more effectively to variations in representation across different views. While existing models like Hybrid-Net, BiGait, and other rely on a static model generation, FOLC-Net enables more targeted updates by adjusting to individual view characteristics, ensuring that performance is not compromised in any single view.
    \item \textbf{Cloning the Global Model for Scalable Training:} \\
    One of the critical features of FOLC-Net is the ability to clone the global model for more efficient scaling with less computation cost. This approach distributes the training load among clients, allowing for early convergence while avoiding the typical bottlenecks seen in large-scale FL systems. The global model cloning mechanism enhances the adaptability of client devices, particularly in scenarios where data is siloed into specific views like axial, coronal, or sagittal. By scaling the model more effectively, FOLC-Net can handle these specific conditions without overburdening the system.

    \item \textbf{ConvNeXt Integration for Enhanced Client Adaptability:} \\
    ConvNeXt, a previous learning framework, is incorporated into FOLC-Net, enhancing the adaptability of client devices. This feature allows client devices to contribute more meaningfully to the knowledge base, even in edge cases where data is sparse or view-specific. Existing models often struggle with such conditions, leading to poor performance in isolated views. ConvNeXt strengthens the capability of FOLC-Net to handle heterogeneous client environments, ensuring that each view is given adequate consideration during the training process. This results in improved accuracy across individual views, particularly for the axial and sagittal views, which tend to be more challenging for other models.

    \item \textbf{Comprehensive Validation Approach:} \\
    A unique aspect of FOLC-Net evaluation methodology is its multi-faceted validation approach, which not only assesses the model performance on combined multi-view representations but also on each individual view. Traditional studies tend to overlook the performance disparities that arise when model are evaluated on single views, often assuming that combined-view performance is sufficient. However, FOLC-Net focus on single-view validation ensures that the model performance under varied conditions is fully understood. This comprehensive approach helps identify specific areas of strength and weakness, allowing for more targeted improvements in individual view performance.
\end{itemize}
\section{Related work}
In several recent studies \cite{khan2025ethicsdesignlifecycleframework}, the integration of DL models and AI-driven algorithms has significantly enhanced the capabilities for detecting and classifying brain tumors by leading to improved patient outcomes and diagnostic precision. Among them, Saeedi et al. \cite{saeedi2023mri} designed a 2D CNN and a convolutional auto-encoder network to analyze a dataset of 3,264 MRI brain images among 4 distinct categories. Their study compared six different machine learning techniques and achieved an accuracy of 90.92\%. Rajput et al. \cite{rajput2024transfer} utilized TL with pre-trained CNN networks to diagnose brain tumors by fine-tuning the features for classifying different tumor types from MRI scans. Their study achieved an average accuracy of 90\% by outperforming conventional techniques, though it still faced challenges in precisely diagnosing brain tumors. Sajjad et al. \cite{sajjad2019multi} employed a deep CNN along with data augmentation methods to classify brain tumors for multi-grade classification. Tumor regions from an MRI image were first segmented using a DL technique and then classified by achieving an accuracy of 94.5\%. Building on the advancements in CNN and deep learning for tumor classification, recent studies have explored the use of FL to address privacy concerns while maintaining high accuracy in brain tumor identification from MRI images.

Islam et al. \cite{islam2023federated} addressed the issue of centralized data collection in brain tumor identification from MRI images using FL. The authors trained several CNN models and selected the top three to form ensemble classifiers. Their FL model was trained with local weights and attained 91.05\% accuracy compared to 96.68\% with the base ensemble model. Their method showed scalability and privacy protection with minimal accuracy loss on a larger dataset. Mastoi et al. \cite{mastoi2025explainable} proposed a collaborative FL model (CFLM) with explainable AI to address brain tumor  diagnosis challenges. By integrating GoogLeNet with FL, their model ensures data privacy while offering interpretability through Grad-CAM and saliency map visualizations for healthcare professionals for achieving 94\% accuracy. Sivakumar et al.\cite{sivakumar2025hybrid}  proposed a distributed learning framework based on FL to build a brain tumor classification system. Their designed framework combines Federated Averaging (FedAvg) and Federated Proximal (FedProx) to train CNNs on data hosted by multiple clients. The authors achieved 97.19\% accuracy by using the Brain Tumor MRI Dataset from Kaggle while maintaining patient data privacy through decentralized learning.

Rajit et al. \cite{rajit2024multi} proposed a system where each institution trains a DL classifier using local data by sharing only model parameters with the global model to maintain patient privacy. They used Blockchain to ensure secure transfer of these parameters. Their  system classified distinct tumor types and achieved 93.75\% accuracy with a weighted average aggregation technique. Jewalikar et al. \cite{jewalikar2025efficient} modified the Xception model for brain tumor classification and introduced Priority-based Client Selection Federated Learning (PC\_FedAvg). Their approach outperformed advanced collaborative learning algorithms, including FedAvg, FedAdam, and FedProx by achieving 91.6\% accuracy. Lhasnaoui et al. \cite{lhasnaoui2024decentralized} employed a FL methodology to train the model. The combination of FL and TL effectively preserves privacy to optimize model performance and encourages strong collaboration among healthcare entities. The author achieved a test accuracy of 88\% which needs to be improved further. Samsuzzaman et al.\cite{samsuzzaman2024optimizing} developed a federated DL model for brain tumor classification using MRI data by enhancing MobileNetV3 architecture with channel and spatial attention blocks. They proposed an innovative non-IID data partitioning approach to improve data distribution across clients. By leveraging the Flower framework and the FedOpt algorithm, their model reached an accuracy of 97.40\% in eight training rounds.

\section{Architecture design of FOLC-Net framework}\label{sec2}
In this section, we provide an in-depth exploration of the optimization of the Novel ShallowFed architecture using MRFO Mechanisms. We begin with an overview of the novel optimized ShallowFed structure, followed by a discussion on ConvNeXt learning and its role in enhancing federated knowledge to empower clients. We then introduce the concept of cloning the Global ShallowFed model as a scalable approach to improve system efficiency. The section further delves into client training and the process of global synchronization, culminating in an overview of the proposed FOLC-Net framework, which integrates these improvements to drive enhanced performance and scalability in federated learning environments.
\subsection{Optimizing the Novel ShallowFed Structure using Manta Ray Foraging Mechanisms}
In our study, we have adopted nature-inspired foraging \cite{khan2025optimize} methods, specifically the CYA and CHF methods. These techniques leverage principles from natural systems to enhance the optimization of model structure. By development a broader exploration of potential solutions and accelerating convergence, they provide a more effective and adaptable approach for optimization tasks. The body structure of a manta ray, depicted in Fig 1, illustrates the natural efficiency and flexibility that inspire CYA and CHF methods.
\begin{figure}[h]
    \centering
    \includegraphics[width = 8cm]{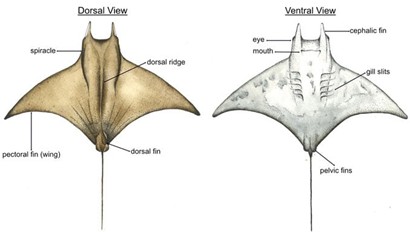}
    \caption{Manta-Ray Foraging structural (Dorsal | Ventral) overview }
    \label{fig:se.png}
\end{figure}
\subsubsection{Aging: Cyclone Method}
A random number, denoted by the symbol $\nu$, is selected from the interval $[0, 1]$ in Eq.~4-5. This randomness technique forms a flexible foundation for simulating various motion patterns across multiple dimensions. By leveraging this approach, it is possible to extend the simulation to $N$-dimensional spaces, enabling the modeling of more complex and diverse dynamic behaviors. This method provides a versatile framework that can be adapted to various applications requiring stochastic modeling, offering both scalability and robustness in simulating high-dimensional systems.
\begin{equation}
\text{FC}_i {(a+1)} = \text{FC}_{\text{best}} + r \times \left( \text{FC}_{(i-1)} {(a)} - \text{FC}_i {(a)} \right) + \beta^{uv} \times \sin(2 \pi \nu) \times \left( \text{FC}_{\text{best}} - \text{FC}_i{(a)} \right)
\tag{4}
\end{equation}

\begin{equation}
\text{CM}_i^{ts}(a+1) = \begin{cases}
\text{CM}_{\text{best}}^{ts}(a) + r \times \left( \text{CM}_{\text{best}}^{ts}(a) - \text{CM}_i^{ts(a)} \right) + \gamma \times \left( \text{CM}_{\text{best}}^{ts}(a) - \text{CM}_i^{ts}(a) \right) & \text{if } i = 1 \\
\text{CM}_{\text{best}}^{ts}(a) + r \times \left( \text{CM}_{(i-1)}^{ts}(a) - \text{CM}_i^{ts}(a) \right) + \gamma \times \left( \text{CM}_{\text{best}}^{ts}(a) - \text{CM}_i^{ts}(a) \right) & \text{if } i = 2, \dots, N
\end{cases}
\tag{5}
\end{equation}

Eq.~(6) presents the mathematical formulation of the CYF strategy, which simulates the spiraling search behavior observed in natural cyclonic patterns. In this context, the weight factor denoted as \( r_1 \) is a stochastic coefficient generated randomly within the interval \([0,1]\). This randomness introduces exploration capability into the algorithm, preventing premature convergence and promoting diversity in the search space. The variable \( R \) represents the maximum number of iterations allowed during the optimization process. It serves as a termination condition that controls the computational usage and ensures that the algorithm progresses within a defined timeframe. By adjusting the value of \( R \), the trade-off between solution accuracy and computational efficiency can be managed.
\begin{equation}
\gamma = 2 \beta^{r_1{\frac{ (R - s + 1)}{R}}} \times \sin(2\pi r_1) \tag{6}
\end{equation}

\begin{equation}
CYA_{\text{random}}^{ts} = L_{\text{lower}}^{ts} + r \times (L_{\text{upper}}^{ts} - L_{\text{lower}}^{ts}) \tag{7}
\end{equation}
The dimension in Eq.~(7) is denoted by the symbol $t_s$. We define $L_{\text{upper}}^{t_s}$ as the upper limit and $L_{\text{lower}}^{t_s}$ as the corresponding lower limit in the given search space. Additionally, $CYA_{\text{random}}^{t_s}$ represents a randomly generated point within this dimension, allowing for exploration across the entire search space. These components form the foundation of the optimization process by delineating the bounds and sampling mechanisms used during solution generation. This mechanism is central to the MRFO algorithm's balance between exploration and exploitation. By allowing manta to infrequently move toward the best-known positions, the algorithm avoids stagnation and promotes a comprehensive traversal of the search domain. Eq.~(8) formally characterizes this exploratory behavior, mathematically modeling the tendency of agents to deviate from the immediate best solution. This ensures robust performance across various optimization problems, particularly those involving multimodal landscapes or deceptive local minima, where a purely exploitative strategy might fail. 

\begin{equation}
\text{CM}_i^{ts} (a+1) = 
\begin{cases} 
\text{CYA}_{\text{random}}^{ts} (a) + r \times (\text{CYA}_{\text{random}}^{ts} (a) - \text{CM}_i^{ts} (a)) + \gamma \times (\text{CYA}_{\text{random}}^{ts} (a) - \text{CM}_i^{ts} (a)),  \\ i = 1\\
\text{CYA}_{\text{random}}^{ts} (a) + r \times (\text{CM}_{i-1}^{ts} (a) - \text{CM}_i^{ts} (a)) 
+ \gamma \times (\text{CYA}_{\text{random}}^{ts} (a) - m_k^{dr} (a)), 
\\ i = 2, \dots, N\\
\end{cases}
\tag{8}
\end{equation}
\subsubsection{Foraging: Chain Method}
CHF is an optimal activity of manta rays because of their synchronized alignment in a head-to-tail arrangement. People work together in this strategy to maximize feeding efficiency. Every ray, excluding the leader, looks for the next potential food source while focusing on the prey that is now within reach. The leader takes the first position and guides the chain toward the prey. This cooperative behavior demonstrates the species capacity to adapt and coordinate for effective foraging. The eating patterns of CHF patients demonstrate both their intelligence and social cooperation. This dynamic can be formalized using Eq.~9-10, which combined establish the CHF principles outlined. This setup guarantees that each manta ray can access not only the option directly ahead but also the most recent version of the best solution found so far, enabling them to make more informed decisions and enhance their search process.
\begin{equation}
\text{CM}_i^{ts}(a+1) = \begin{cases}
\text{CM}_i^{ts}(a) + r \times (\text{CM}_{\text{best}}^{ts}(a) - \text{CM}_i^{ts}(a)) + \nu \times (\text{CH}_{\text{best}}^{ts}(a) - \text{CM}_i^{ts}(a)), & \text{if } i=1, \\
\text{CM}_i^{ts}(a) + r \times (\text{CM}_{i-1}^{ts}(a) - \text{CM}_i^{ts}(a)) + \nu \times (\text{CH}_{\text{best}}^{ts}(a) - \text{CM}_i^{ts}(a)), & \text{if } i = 2, \dots, N.
\end{cases}
 \tag{9}
\end{equation}

\begin{equation}
\eta = 2 \times \alpha \times \sqrt{\left| \log(\alpha) \right|}. \tag{10}
\end{equation}
\bmhead{Outcome}
In the context of model optimization, ShallowFed structure tuning for medical image analysis benefits significantly from the integration of CYA and CHF techniques. CYA introduces a dynamic optimization paradigm by cyclically adjusting filters, kernel sizes, activation function, dropout rate and other. This modulation not only prevents premature convergence to local minima but also facilitates robust exploration of the loss landscape. The adaptability of CYA is particularly crucial for medical imaging, where data heterogeneity such as variations in resolution, contrast, or modality can easily lead to overfitting or undergeneralization. By continuously refining the optimization trajectory, CYA ensures the model remains resilient and generalizable across a wide array of diagnostic tasks. Moreover, ShallowFed optimal structure adaptation allows the model to self-adjust its architectural depth and representation power as new data or tasks are introduced, ensuring sustained relevance and performance.

Complementing CYA, CHF offers a layered strategy by optimizing the ShallowFed structure hierarchically. It progressively tunes architectural parameters such as convolutional kernel sizes, the number of filters, drop rate and the depth of layers based on the performance of earlier stages. This cascaded adjustment ensures that improvements in early-stage modules propagate effectively to deeper layers, yielding a compound enhancement in feature extraction. As a result, CHF enables the model to evolve with task complexity, optimizing both computational efficiency and diagnostic precision. When CYA and CHF are jointly applied, the ShallowFed framework becomes a highly adaptive and self-regulating system. It can learn from data evolution, maintain structural relevance, and optimize resource allocation, leading to enhanced diagnostic accuracy, reduced inference time, and more reliable clinical decision support in real-world medical imaging applications.
\subsubsection{ShallowFed optimized structure with MRFO}
Due to the inherent limitations of models \cite{khan2024boosting, khan2024hybrid} that lack the capability for illness-specific hierarchical feature learning, their application in medical disease diagnostic tasks can often lead to suboptimal performance. In light of these challenges, we have embarked on a comprehensive study of ShallowFed structure to address these shortcomings and optimize the predictive performance of our disease detection model. By focusing on the ShallowFed structure, we aim to enhance the model ability to capture the intricate patterns and dependencies present in medical imaging data, enabling more precise diagnostics. To achieve this, MRFO approach allows us to explore and fine-tune the search space for the optimal ShallowFed structure effectively. Our experiments are meticulously designed, starting with an initial population of 10 candidate solutions, each representing a unique ShallowFed structure. These solutions undergo eight iterative cycles, each lasting 40 epochs, during which the structure is continuously refined. This iterative process not only enhances the structural components of the model but also improves its overall performance, ensuring that the model can adapt to complex medical datasets and provide accurate predictions. Table 1 presents the candidate ShallowFed model structure settings.
\begin{table}[h!]
\centering
\caption{Model ShallowFed structure parameters to optimized by MRFO}
\begin{tabular}{cc}
\hline
\textbf{Component} & \textbf{Structure Settings} \\ \hline
Filters & 8, 16, 32, 64, 128, 256, 512 \\ 
Kernel Sizes & 3x3, 5x5, 7x7, 9x9 \\ 
Activation Function & ReLU, LeakyReLU, Tanh, ELU \\ 
Dropout & 0.1, 0.2, 0.3, 0.4, 0.5 \\ 
Number of Neurons & 16, 32, 64, 128 \\ \hline
\end{tabular}
\label{table:shallowfed_structure}
\end{table}
\bmhead{Outcome}
To ensure a more efficient and thorough structure optimization process, we have established a population exploration range with clear boundaries, limiting the search within a specified region of the structure settings space. This approach ensures that the MFRO remains focused on a relevant subset of potential solutions, thus enhancing computational efficiency. Additionally, we have incorporated a patience mechanism with a value of 10 to prevent premature termination, allowing the algorithm to explore the solution space in greater detail. This technique helps to safeguard against the possibility of early convergence to suboptimal solutions, ensuring a more robust exploration of the structure settings landscape. The solution space in our optimization process is designed with five distinct dimensions, representing the various configurations of the model structure, culminating in the optimal ShallowFed architecture. This exploration accommodates the wide range of potential structural adjustments, which are essential for optimizing the model performance.
\subsubsection{Objective function}
To enhance the precision of disease diagnosis, we simultaneously refined the CHF and CYA objective functions within the MRFO algorithm. This dual optimization strategy allowed for more comprehensive exploration and exploitation of the structure settings. By iterating through a pre-defined number of cycles, the MRFO algorithm adjusted the structure settings of the ShallowFed model, using standard evaluation metrics for each population in the process. During this optimization, the structure settings of the model, denoted as \( \text{ShallSS}_{ij} \), were continuously modified. The optimization procedure, constrained to a maximum of ten iterations, efficiently narrowed the search space, aiming to identify the optimal combination of model settings while minimizing the computational cost. Once the CYA and CHF no longer yielded improvements after the set number of iterations, it was deemed that the current model configurations had reached the optimal solution for the given search problem. This approach helped balance between computational efficiency and diagnostic performance.

The solution is a collection of structure settings, represented by \( \text{ShallSS}_{i} \), that define various parameters, such as filters (\( \text{fil}_{i} \)), kernel sizes (\( \text{ks}_{i} \)), activation functions (\( \text{act}_{i} \)), dropout (\( \text{dro}_{i} \)), and the number of neurons (\( \text{NN}_{i} \)). These parameters serve as the building blocks of the model configuration. Within each iteration \( a \) of the population \( \text{IP}_a^{\text{shallss}}(\text{structure}) \), these parameters \( \text{ShallSS}_{ij} \) are dynamically adjusted, leveraging optimization techniques to refine and improve the model architecture. The primary objective is to optimize the search for the most effective ShallowFed model setting by using two key methods: CHF and CYA. Both methods work by fine-tuning the structure parameters iteratively, ensuring that the model performance is continually refined. After training the ShallowFed model with the validation data, standard evaluation metrics are employed to assess its performance. The best-performing settings, \( \text{IP}_{\text{best}}^{\text{shallSS}}(\text{structure}) \), are recorded and represent the ideal model configuration for each population iteration \( x \). These best scores are then aggregated into a final score, which serves as the benchmark for determining the optimal model structure. The overarching goal of the objective function is to maximize validation accuracy, driving the optimization process toward configurations that deliver the highest levels of performance. By refining these parameters over successive iterations, the solution continuously improves, approaching the ideal ShallowFed model configuration that yields the best results.
The fitness function can be expressed as Eq.~11:

\[
\text{Fitness} = \text{IP}_{\text{best}}^{\text{shallSS}(\text{structure})} = \sum_{i=1}^{N} \left( \sum_{j=1}^{N} \left( \text{ShallSS}_{ij} \left( \max(\text{Validation\_Acc}) \right) \right) \right)
\]

where

\[
\text{ShallSS}_{ij} = (\text{fil}_{ij}, \text{ks}_{ij}, \text{act}_{ij}, \text{dro}_{ij}, \text{NN}_{ij}) \tag{11}
\]
\subsection{Novel optimized ShallowFed architecture overview}
The ShallowFed architecture is a robust DL model designed for analyzing medical images to aid in disease diagnosis. By leveraging optimal layers order such as convolutional, pooling, activation, and regularization, the model extracts meaningful features and delivers accurate classifications essential for medical decision-making. Below is a breakdown of the layer-by-layer structure and its benefits for diagnosing medical conditions. Fig 2 presents the visualization of the Layer-by-layer ShallowFed architecture diagram. 
\bmhead{1. Input Image}
\begin{itemize}[left=2em] 
    \item \textbf{Description:} The architecture begins with a preprocessed input image, typically a medical image. This study has utilized three image modalities, namely MRI, CT scan, and Histogram.
    \item \textbf{Benefit for Diagnosis:} The model processes detailed medical images to extract concern information, such as abnormalities or diseases.
\end{itemize}

\bmhead{2. Feature Learning (Convolutional Layer)}
\begin{itemize}[left=2em]
    \item \textbf{Description:} The convolutional layers apply filters (32) with kernel sizes (3x3) to the input image to detect low-level features (edges, textures) that are crucial in medical images.
    \item \textbf{Benefit for Diagnosis:} The convolutional layers help identify key features like edges of tumors or abnormal structures, essential for initial pattern recognition in medical diagnosis.
\end{itemize}

\bmhead{3. Activation Function (LeakyReLU (\( \alpha \)=0.1))}
\begin{itemize}[left=2em]
    \item \textbf{Description:} A LeakyReLU activation function is applied element-wise to introduce non-linearity to the model. The parameter (\( \alpha \)=0.1) controls the slope for negative inputs.
    \item \textbf{Benefit for Diagnosis:} LeakyReLU prevents dead neurons and helps the network capture complex, subtle patterns in disease images that could indicate symptoms.
\end{itemize}

\bmhead{4. Spatial Dimension Control (Maxpool 2D (2x2))}
\begin{itemize}[left=2em]
    \item \textbf{Description:} Maxpool reduces the spatial dimensions of the input image by taking the maximum value from each 2x2 region.
    \item \textbf{Benefit for Diagnosis:} This layer preserves important features while reducing computational complexity, making the network more efficient and focusing on critical features for disease identification (e.g., tumors or lesions).
\end{itemize}

\bmhead{5. Feature Learning (Convolutional Layer) (Repetition)}
\bmhead{6. Activation Function (LeakyReLU (\( \alpha \)=0.1)) (Repetition)}
\bmhead{7. Spatial Dimension Control (Maxpool 2D (2x2)) (Repetition)}

\bmhead{8. Handle Overfitting (Dropout (\( \alpha \) = 0.25))}
\begin{itemize}[left=2em]
    \item \textbf{Description:} Dropout regularization is applied to prevent overfitting by randomly disabling neurons during training.
    \item \textbf{Benefit for Diagnosis:} Dropout helps the model generalize better, ensuring that it is less likely to overfit to specific features of the training data, which improves performance on unseen medical images.
\end{itemize}

\bmhead{9. Feature Learning (Convolutional Layer) (Repetition)}
\bmhead{10. Activation Function (LeakyReLU (\( \alpha \)=0.1)) (Repetition)}
\bmhead{11. Feature Learning (Convolutional Layer) (Repetition)}
\bmhead{12. Activation Function (LeakyReLU (\( \alpha \)=0.1)) (Repetition)}
\bmhead{13. Spatial Dimension Control (Maxpool 2D (2x2)) (Repetition)}
\bmhead{14. Handle Overfitting (Dropout (\( \alpha \) = 0.25)) (Repetition)}

\bmhead{15. Feature Learning (Convolutional Layer) (Repetition)}
\bmhead{16. Activation Function (LeakyReLU (\( \alpha \)=0.1)) (Repetition)}
\bmhead{17. Feature Learning (Convolutional Layer) (Repetition)}
\bmhead{18. Activation Function (LeakyReLU (\( \alpha \)=0.1)) (Repetition)}

\bmhead{19. Normalize Generated Features (Batch Norm)}
\begin{itemize}[left=2em]
    \item \textbf{Description:} Batch normalization normalizes the outputs of previous layers to improve training stability and speed up convergence.
    \item \textbf{Benefit for Diagnosis:} This step helps in reducing the impact of data variance, improving the model’s ability to generalize across different datasets, ensuring robust disease classification even in noisy data.
\end{itemize}

\bmhead{20. Feature Learning (Convolutional Layer) (Repetition)}
\bmhead{21. Normalize Generated Features (Batch Norm) (Repetition)}
\bmhead{22. Activation Function (LeakyReLU (\( \alpha \)=0.1)) (Repetition)}

\bmhead{23. 1-Dimension Vector (Flatten)}
\begin{itemize}[left=2em]
    \item \textbf{Description:} The output from the convolution and pooling layers is flattened into a one-dimensional vector.
    \item \textbf{Benefit for Diagnosis:} Flattening transforms the multi-dimensional feature maps into a format that can be processed by fully connected layers, enabling the model to learn high-level decision rules for diagnosis.
\end{itemize}

\bmhead{24. Leveling Feature (Feature Uniform)}
\begin{itemize}[left=2em]
    \item \textbf{Description:} This block helps to adjust or level the feature representations to make them more uniform and stable.
    \item \textbf{Benefit for Diagnosis:} It ensures that the features used for classification are standardized, improving the model’s ability to recognize patterns across varying datasets and helping it focus on the most relevant aspects of the medical images.
\end{itemize}

\bmhead{Dense Layer (Neurons-128)}
\begin{itemize}[left=2em]
    \item \textbf{Description:} A fully connected dense layer is applied to the flattened output. This layer connects every neuron from the previous layer to each neuron in this layer.
    \item \textbf{Benefit for Diagnosis:} Dense layers help to combine all the features learned so far and produce an output that is used for final prediction. This is crucial for high-level decision-making, leading to the final diagnosis.
\end{itemize}

\bmhead{Activation Function (LeakyReLU (\( \alpha \)=0.1)) (Repetition)}

\bmhead{Normalize Generated Features (Batch Norm (Momentum (0.99), Epsilon (0.01))) (Repetition)}

\bmhead{Handle Overfitting (Dropout (\( \alpha \) = 0.25)) (Repetition)}

\bmhead{Dense Layer (Neurons-64)}
\begin{itemize}[left=2em]
    \item \textbf{Description:} A fully connected dense layer is applied to the output from the previous layers.
    \item \textbf{Benefit for Diagnosis:} This layer refines the learned representations and combines them for the final output.
\end{itemize}

\bmhead{Activation Function (LeakyReLU (\( \alpha \)=0.1)) (Repetition)}
\bmhead{Normalize Generated Features (Batch Norm (Momentum (0.99), Epsilon (0.01))) (Repetition)}
\bmhead{Handle Overfitting (Dropout (\( \alpha \)= 0.5)) (Repetition)}
\begin{figure}[h!]
    \centering
    \includegraphics[width = 12cm]{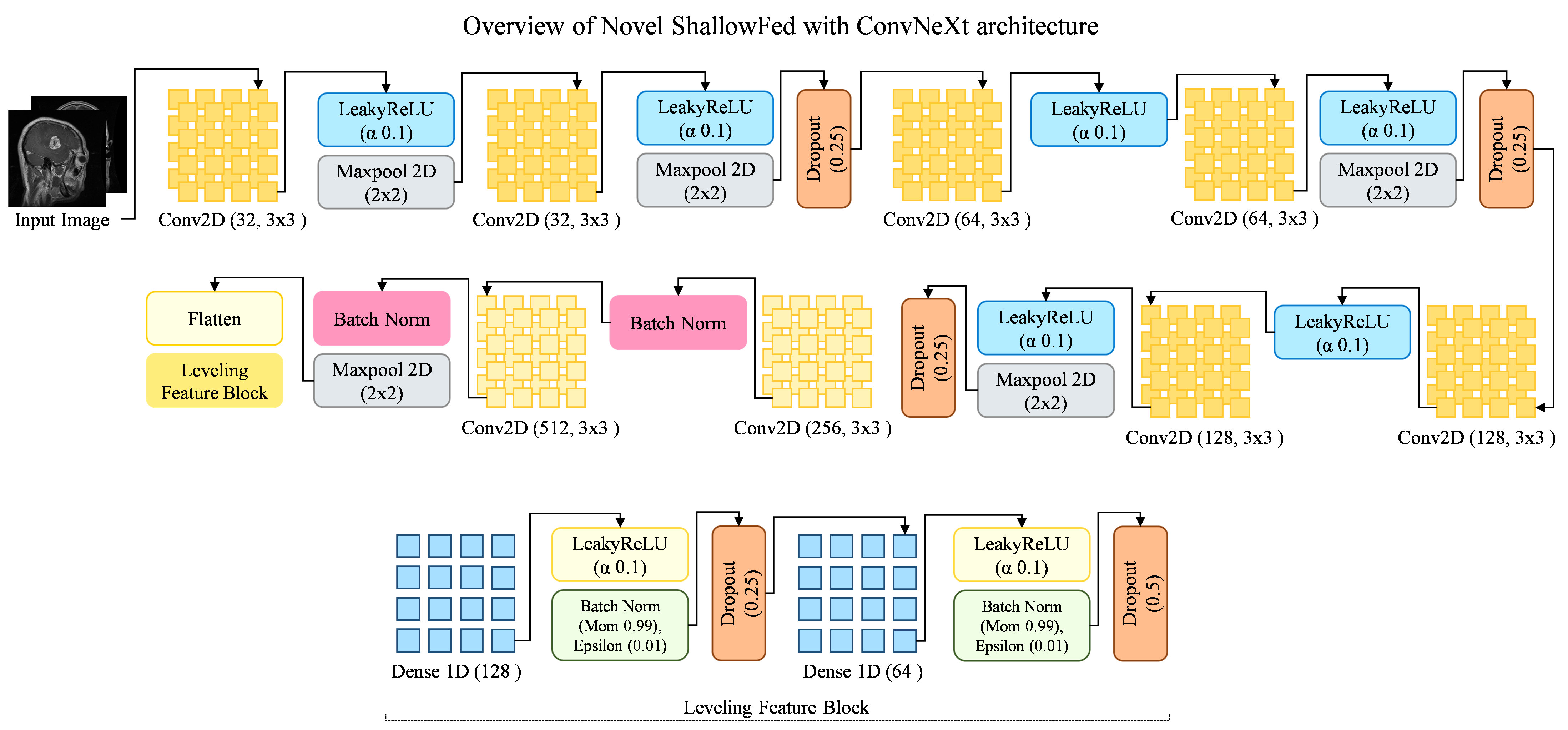}
    \caption{An architectural overview of the novel ShallowFed model, optimized with the MRFO algorithm and utilizing the ConvNeXt learning}
    \label{fig:se.png}
\end{figure}
\subsection{Federated knowledge to empowering clients}
Federated knowledge \cite{samsuzzaman2024optimizing} plays a pivotal role in enhancing the capabilities of clients, particularly in our context of multiclass disease diagnosis. By leveraging distributed data sources while ensuring privacy, it enables models to train on diverse datasets across multiple clients, improving generalization and performance. In this approach, our ShallowFed models can benefit from the collective knowledge shared in a federated manner, ultimately enhancing diagnostic accuracy across various disease classes. The weight scaling factor in a FL setup is essential for ensuring that each client contribution to the model update is proportional to the amount of data they hold. In the context of multiclass disease diagnosis, this scaling mechanism allows for the fair aggregation of model updates from various clients, each having different amounts of training data.

In this study, the weight scaling factor is calculated by first determining the total number of data points across all clients (\texttt{global\_count}) and the number of data points held by a specific client (\texttt{local\_count}). The ratio of these two values gives the weight scaling factor, ensuring that clients with more data have a proportionally higher influence on the global model update. This approach helps maintain the balance between clients with varying amounts of data, thereby enhancing the model's ability to generalize across diverse datasets for improved disease diagnosis performance.
The Weight Scaling Factor is given by Eq.~12-14:

\[
\text{Weight Scaling Factor} = \frac{N_{\text{local}}}{N_{\text{global}}} \tag{12}
\]

where:

\[
N_{\text{local}} = \text{cardinality}(D_{\text{client}}) \times \text{batch size} \tag{13}
\]
Here, $\text{cardinality}(D_{\text{client}})$ represents the total number of batches in the client dataset, and the batch size is the number of data points in each batch.

\[
N_{\text{global}} = \sum_{i=1}^{n} \left( \text{cardinality}(D_{\text{client}}) \times \text{batch size} \right) \tag{14}
\]

where $n$ is the total number of clients, and $D_{\text{client}}$ is the dataset for the $i$-th client. Thus, the weight scaling factor ensures that each client’s model update is scaled according to the proportion of data they contribute to the global dataset.
\subsection{Cloning the Global Novel ShallowFed: A Scalable approach}
Cloning the Global Novel ShallowFed model introduces a scalable and computationally efficient solution for multiclass disease diagnosis. In this approach, the same shallow model is utilized both globally and by each individual client, ensuring minimal computational overhead while maintaining the integrity of the global model. This strategy allows the global model to be replicated at the client level, enabling each client to perform local training without the need for complex, resource-intensive architectures. By using a shared model across both the global and client levels, the system achieves a balance between scalability, efficiency, and high-performance multiclass classification for disease diagnosis. The use of a lightweight model for both global and client-level tasks ensure faster training times, reduced memory usage, and lower energy consumption, making it ideal for resource-constrained environments. This approach also minimizes the need for frequent data transfers between clients and the global model, further reducing the computational burden and improving system efficiency.
\subsection{Client Training and Global Synchronization}
Client Training and Global Synchronization is a critical process in FL, particularly for applications like multiclass disease diagnosis. In this framework, individual clients (e.g., medical institutions, hospitals, or devices) train local models using their own datasets, which may be diverse in terms of disease types and patient data. These local models are then periodically synchronized with a global model to aggregate the learned parameters while maintaining data privacy.

The process follows several steps:
\bmhead{Client Training}

Each client trains a local model using its own data. In the context of disease diagnosis, this model is typically shallow to balance performance with efficiency. The training happens on the client-side, and only model updates (such as gradients) are shared, ensuring sensitive patient data remains private.

Each client \( C_i \) trains a shallow model \( \Phi_i \) on its local dataset \( D_i \). Here, the model is a shallow CNN, and the training objective is to minimize a loss function \( \rho(\Phi_i; D_i) \). The update for the local model parameters \( \Phi_i \) after training on client \( C_i \) data is given by Eq.~15:

\[
\Phi_i^{\text{new}} = \Phi_i - \eta \nabla \rho(\Phi_i; D_i) \tag{15}
\]

Where:
\begin{itemize}
  \item \( \Phi_i \) is the model parameters at client \( C_i \),
  \item \( \eta \) is the learning rate,
  \item \( \nabla \rho(\Phi_i; D_i) \) is the gradient of the loss function with respect to the model parameters \( \Phi_i \) using data from \( D_i \).
\end{itemize}
\bmhead{Global Synchronization}
After local training, the model updates from all clients are sent to a central server. The server performs a global aggregation, where it combines these updates (usually by averaging the model parameters). This allows the global model to learn from the collective knowledge of all clients, enhancing its ability to generalize across diverse data sources.

After local training, the global model parameters \(\Phi_{\text{global}}\) are updated based on the weighted average of the local models \(\Phi_i\) from each client. The global model is represented as Eq.~16:

\[
\Phi_{\text{global}}^{\text{new}} = \frac{1}{N} \sum_{i=1}^{n} w_i \Phi_i^{\text{new}} \tag{16}
\]

Where:

\begin{itemize}
    \item \(N\) is the total number of clients. In our case, it is \(C_5\).
    \item \(w_i\) is the weight assigned to the \(i\)-th client, typically proportional to the number of data points \(\lvert D_i \rvert\) the client has, so
    \[
    w_i = \frac{\lvert D_i \rvert}{\sum_{i=1}^{n} \lvert D_i \rvert}
    \]
    \item \(\Phi_i^{\text{new}}\) are the updated parameters from each client \(C_i\).
    \item \(\Phi_{\text{global}}^{\text{new}}\) is the new global model parameters after aggregation.
\end{itemize}
This averaging process ensures that the global model benefits from the collective knowledge of all the clients while maintaining privacy.
\bmhead{Scalable Approach:} For the model to scale across multiple clients with varying amounts of data, the process must be efficient. The equation for the weight scaling factor is given by Eq.~17:

\begin{equation}
\text{Weight Scaling Factor} = \frac{|D_i|}{\sum_{i=1}^{n} |D_i|} \tag{17}
\end{equation}

Where: 
\begin{itemize}
    \item $|D_i|$ is the number of data points in client $C_i$ dataset.
    \item $\sum_{i=1}^{n} |D_i|$ is the total number of data points across all clients.
\end{itemize}

This weight scaling ensures that clients with more data contribute more to the global model.
\subsection{Overview of proposed FOLC-Net framework}
The proposed FOLC-Net framework is designed to significantly enhance the performance, scalability, and efficiency of FL systems. By integrating a combination of architectural optimizations and efficient training methodologies, FOLC-Net builds upon the ShallowFed architecture, which has been further refined with the inclusion of MRFO mechanisms. These innovations allow for more efficient model updates, leading to a higher degree of accuracy within federated environments, where communication constraints and heterogeneous client conditions are critical factors. A unique feature of the FOLC-Net is its ability to clone the Global ShallowFed model, which enables the system to scale more effectively. This strategic approach to training not only boosts system efficiency but also mitigates the typical communication overhead associated with large-scale FL systems. By allowing for the cloning of the global model and distributing the training load, the framework offers an innovative solution to overcome one of the major bottlenecks in FL. Another significant contribution introduced by FOLC-Net is the incorporation of ConvNeXt, a leading-edge learning framework, which further strengthens the capabilities of client devices. This feature enhances the adaptability of client devices, allowing them to contribute more effectively to the knowledge base and improving the overall FL process at both the client and global levels. The result is a more robust and responsive FL environment where client devices are empowered to make more meaningful updates without overburdening the network. However, the validation of FOLC-Net requires a multi-faceted evaluation approach to ensure its effectiveness. Our hypothesis emphasizes the need to not only evaluate models based on combined multi-view representations, but also on individual view representations, which are often overlooked in traditional studies \cite{bilal2025amalgamation, bilal2024differential}. By examining each view independently, FOLC-Net offers deeper insights into the model performance under varied conditions, revealing performance disparities or strengths that combined evaluations may fail to capture. This comprehensive validation approach ensures a more thorough understanding of the model capabilities, ultimately improving its applicability to real-world scenarios where different data views may be more relevant in certain contexts. Fig 3 presents the overview of proposed framework diagram. 
\begin{figure}[h]
    \centering
    \includegraphics[width = 14cm]{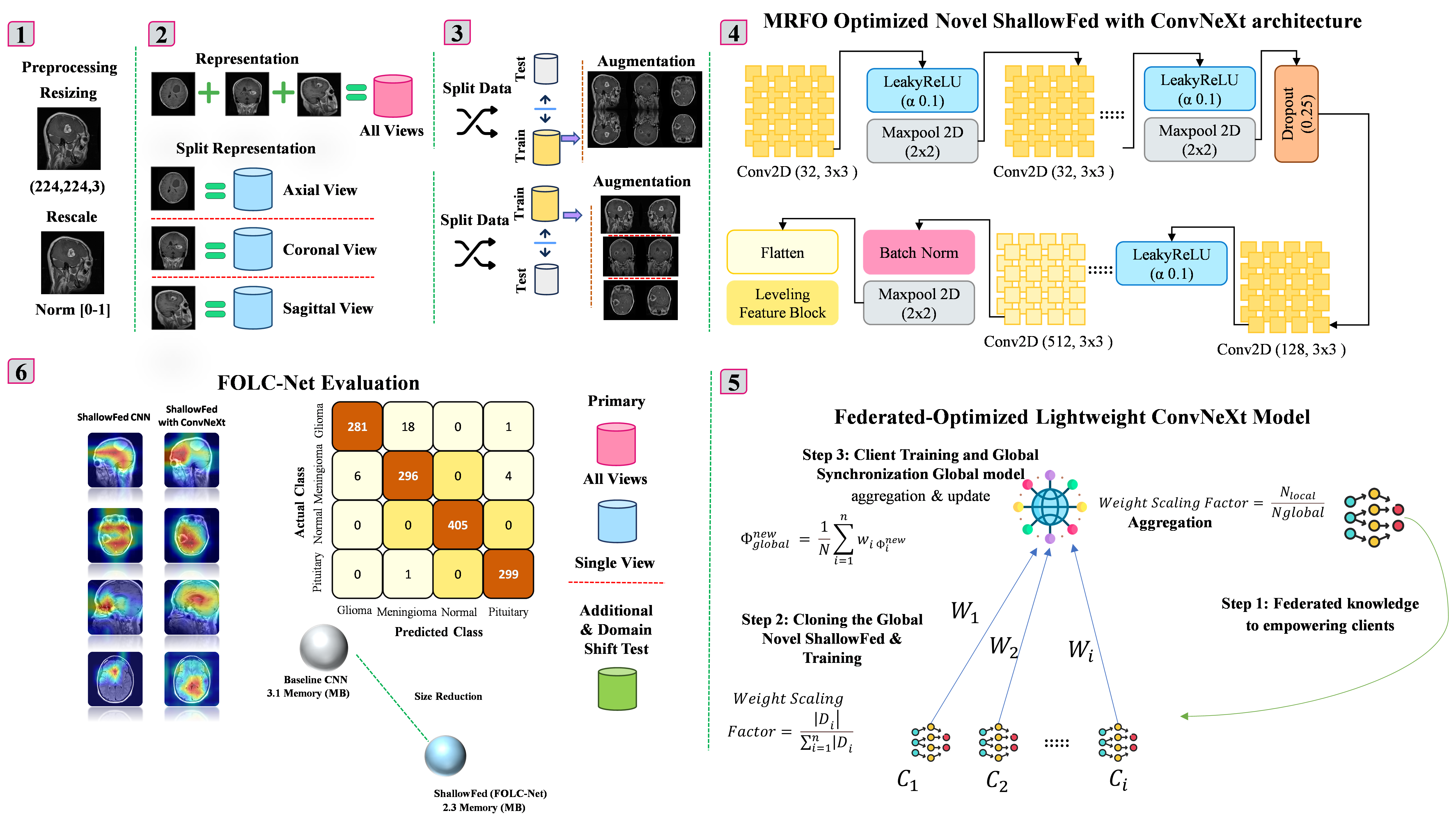}
    \caption{An overview of the proposed FOLC-Net framework, which integrates MRFO to optimize cancer prediction performance through FL: Multi-view to individual-views}
    \label{fig:se.png}
\end{figure}
\section{Result and implementation}
This section provides an in-depth explanation of the various components and evaluations involved in the proposed FOLC-Net model. It covers the utilized hyperparameters and metrics for performance evaluation, the experimental setup, and an overview of the dataset used, specifically focusing on the transformation from multi-view to single-view representation. The section also discusses data preprocessing and balancing techniques, followed by a detailed performance analysis of FOLC-Net in comparison to existing models. Additionally, it explores the interpretability of the model using GRADCAN and GRADCAN++ for explainable diagnosis, as well as the t-distributed Stochastic Neighbor Embedding (t-SNE) test for dimensionality reduction. The robustness of the model is further validated through statistical analysis using Chi-Square tests. Lastly, the section includes additional dataset testing and a comprehensive evaluation across different disease modalities, reinforcing the model performance and versatility.
\subsection{Utilized hyperparameter and metrices to evaluate FOLC-Net}
We trained the proposed model with a carefully selected set \cite{saleem2023fnreq, asim2022circ} of hyperparameters to ensure consistency and robust performance. Specifically, a learning rate of 0.001 was chosen to optimize convergence, with the model being trained over 30 epochs to allow sufficient learning while preventing overfitting. A batch size of 64 was used, striking a balance between computational efficiency and model accuracy. The Adam optimizer played a pivotal role in achieving enhanced performance, providing faster convergence and better generalization compared to other optimization techniques. Its ability to adapt the learning rate for each parameter individually led to more stable training and reduced the risk of overshooting, making it highly effective in our ShallowFed case. Additionally, the categorical cross-entropy loss function was employed, which proved essential for our disease classification task. This loss function helped ensure a proper evaluation of the model accuracy, guiding the optimization process effectively. Overall, the combination of these carefully chosen hyperparameters and techniques contributed to significant improvements in the ShallowFed model performance, showcasing the robustness and efficiency of FOLC-Net methodology.

We conducted a comprehensive evaluation of the proposed FOLC-Net architecture to assess its performance, employing widely recognized evaluation metrics commonly used in DL tasks. The results were consistent with findings from previous studies, affirming the validity of our approach. Each diagnosed sample was classified into one of four categories: True Positive, True Negative, False Positive, and False Negative. The performance evaluation was conducted using these classifications, with the corresponding mathematical formulations for calculating key performance metrics such as accuracy (Eq.~18), precision (Eq.~19), recall (Eq.~20), and F1-score (Eq.~21), which provide a quantitative understanding of the model's effectiveness in distinguishing between correct and incorrect diagnoses.

\[
\text{Accuracy} = \frac{\text{True}_\text{Positive} + \text{True}_\text{Negative}}{\text{True}_\text{Positive} + \text{False}_\text{Positive} + \text{True}_\text{Negative} + \text{False}_\text{Negative}} \tag{18}
\]

\[
\text{Precision} = \frac{\text{True}_\text{Positive}}{\text{True}_\text{Positive} + \text{False}_\text{Positive}} \tag{19}
\]

\[
\text{Recall} = \frac{\text{True}_\text{Positive}}{\text{True}_\text{Positive} + \text{False}_\text{Negative}} \tag{20}
\]

\[
\text{F1 Score} = 2 \times \frac{\text{Precision} \times \text{Recall}}{\text{Precision} + \text{Recall}} \tag{21}
\]
\subsection{Experimental setup}
The study employed the Keras framework along with Python programming to develop and train the FOLC-Net framework. The Keras framework, known for its simplicity and flexibility, was utilized to construct the ShallowFed model architecture, while Python provided an efficient environment for the implementation and execution of various tasks. To maximize computational efficiency and speed, all experiments were conducted within the Python environment, leveraging the capabilities of a GPU to accelerate both training and testing processes. Specifically, an NVIDIA RTX350 Tesla GPU was employed, coupled with 16 GB of RAM, ensuring sufficient hardware resources for the intensive operations involved. This setup was integral in efficiently handling the large-scale data processing required for the model performance evaluation.
\subsection{Ensuring Reporting Compliance in Medical Research: The Role of the TRIPOD Checklist in Disease Diagnosis}
Accurate and transparent reporting of medical research is essential for confirming our understanding of disease mechanisms, treatment effectiveness, and patient outcomes. Ensuring compliance with established guidelines, such as the TRIPOD checklist (Table 2), is crucial in medical disease diagnosis. The TRIPOD checklist provides a structured framework for reporting prediction models, promoting consistency, reliability, and reproducibility in research findings. The standing of following to reporting standards is especially significant in disease diagnosis, where the outcomes of predictive FOLC-Net framework directly impact patient care. By improving the transparency of model development, validation, and interpretation, our study can ensure that findings are both trustworthy and actionable in clinical practice. This section explores the pivotal role of the TRIPOD checklist in ensuring high-quality reporting, enhancing the reliability of diagnostic models, and fostering better clinical decision-making.
\begin{table}[h!]
\centering
\caption{TRIPOD Checklist for ShallowFed with ConvNeXt (FOLC-Net)}
\begin{tabular}{p{3cm}  p{4cm}  p{5cm}  p{1cm}}
\hline
\textbf{TRIPOD Section} & \textbf{Description} & \textbf{Our Paper Mapping} & {Checklist} \\
\hline
\textbf{Title} & A clear and concise title that reflects the objective and scope of the study. & A Novel Federated Learning Framework for Medical Diagnosis Using the FOLC-Net Framework & Pass \\
\textbf{Abstract} & Summarizes the study, including the model purpose, methodology, outcomes, and results. & Abstract provides a brief summary of FOLC-Net's purpose, the methodology used, and its effectiveness in multi-view to single-view brain tumor diagnosis. & Pass \\
\textbf{Introduction} & Describes the problem, objectives, model purpose, and potential outcomes. & Introduction explains the need for FOLC-Net, its relevance to medical diagnosis, and the expected improvements over existing models. & Pass \\
\textbf{Related Work} & Review of existing models, methods, and relevant research, and their limitations. & Compare FOLC-Net with previous federated learning models and their use in medical diagnostics, highlighting the unique contributions of the work. & Fair \\
\textbf{Architecture Design of FOLC-Net Framework} & Detailed description of the model architecture. Includes model components, algorithms, and design principles. & Overview of FOLC-Net's structure and components such as ShallowFed, MRFO, and FL integration. & Pass \\
\textbf{Model Development and Validation} & Details about how the model was developed, trained, and validated. & Sections 4.6, 4.7, and 4.10 describe the validation using classification performance, statistical tests, and interpretability analysis. & Pass \\
\textbf{Objective Function} & Clear description of the model objective function and performance metrics used in training and validation. & Section 3.1.3.1 describes the objective function for optimizing ShallowFed with MRFO. & Pass \\
\textbf{Model Calibration and Tuning} & Discussion of hyperparameters, model calibration, and tuning methods. & Section 4.1 covers hyperparameters and metrics used to evaluate the model, ensuring they align with TRIPOD guidelines. & Pass \\
\textbf{Evaluation Metrics} & Use of relevant evaluation metrics (accuracy, AUC, etc.) to assess model performance. & Sections 4.6 and 4.7 provide metrics for classification performance and outperformance compared to existing models. & Pass \\
\textbf{Results} & Clear presentation of model results, including comparisons with existing methods and statistical significance. & Sections 4.6 to 4.13 cover various evaluation aspects, including performance comparisons, statistical tests, and the interpretability of results. & Pass \\
\textbf{Discussion} & Interpretation of the results and why the proposed model performs better than existing models. & Section 5 discusses why FOLC-Net outperforms existing models and addresses views where other models struggle. & Pass \\
\textbf{Model Limitations} & Acknowledgment of the limitations of the model, including potential biases, generalizability, and data-related issues. & Discuss any limitations of FOLC-Net, such as data constraints, model complexity, or edge cases where it might underperform. & Unclear \\
\textbf{Conclusion} & Summarizes the model impact, potential future work, and broader applicability. & Conclusion should summarize the significance of FOLC-Net, its impact on medical diagnostics, and potential for future work or improvements. & Pass \\
\textbf{Data and Code Availability} & Provide access to the data and code used in the study for transparency and reproducibility. & Ensure that data (e.g., medical datasets) and code are made available for reproducibility. & Unclear \\
\textbf{Ethical Considerations} & Address any ethical concerns, particularly in relation to patient data and privacy. & Ensure that ethical considerations are discussed, including data privacy and compliance with ethical guidelines for medical research. & Pass \\
\textbf{Reporting Compliance} & Compliance with reporting guidelines, e.g., TRIPOD for medical research studies. & Ensure compliance with TRIPOD guidelines by following the checklist for transparent reporting in your study. & Pass \\
\hline
\end{tabular}
\end{table}

\subsection{Dataset overview: Multi to Single view Representation }
The dataset consists of MRI images of brain tumors, which are publicly available at the Kaggle directory 
\footnote{https://www.kaggle.com/datasets/masoudnickparvar/brain-tumor-mri-dataset} MRI Brain Tumor Dataset. The images are segmented according to different views and types of cancer.The images are displayed fig 4 in multiple orientations: Axial, Coronal, and Sagittal views. Each view corresponds to a different plane through which the MRI scan captures the brain, providing a comprehensive understanding of the tumor characteristics from various angles. The dataset \cite{siddique2020deep} includes images representing several brain cancers types, with each type being shown in all three views. It is important to assess each type of cancer in these views as it provides more precise data for ShallowFed model training and can improve detection accuracy. For each cancer type, images are split based on their respective views. The total number of views per cancer type is presented to ensure that model receives a consistent set of data for learning. We utilized a dataset consisting of 7,023 brain MRI images sourced from the Kaggle directory. The dataset covers four distinct types of MRI images: pituitary, glioma, meningioma, and normal brain. These images are provided in three different anatomical views which offer comprehensive perspectives of brain structures. This diverse set of imaging modalities enables a multi-view approach to the analysis, ensuring a more robust understanding of various brain conditions. The inclusion of multiple views enhances the dataset ability to capture the spatial variations within the brain efficiently as compared to single-view, making it an invaluable resource for training DL models that can aid in the diagnosis and classification of brain abnormalities in multi-view.
   
For example, the images of each cancer type will be represented across all three views, providing a more robust dataset for analysis. This view-based approach is essential because different orientations can highlight unique tumor characteristics, which might be missed when considering combine view. Therefore, splitting cancer types by multi-view to single-view representation for comprehensive modeling enables the development of more accurate and generalized FOLC-Net framework.
\begin{figure}[h]
    \centering
    \includegraphics[width = 8cm]{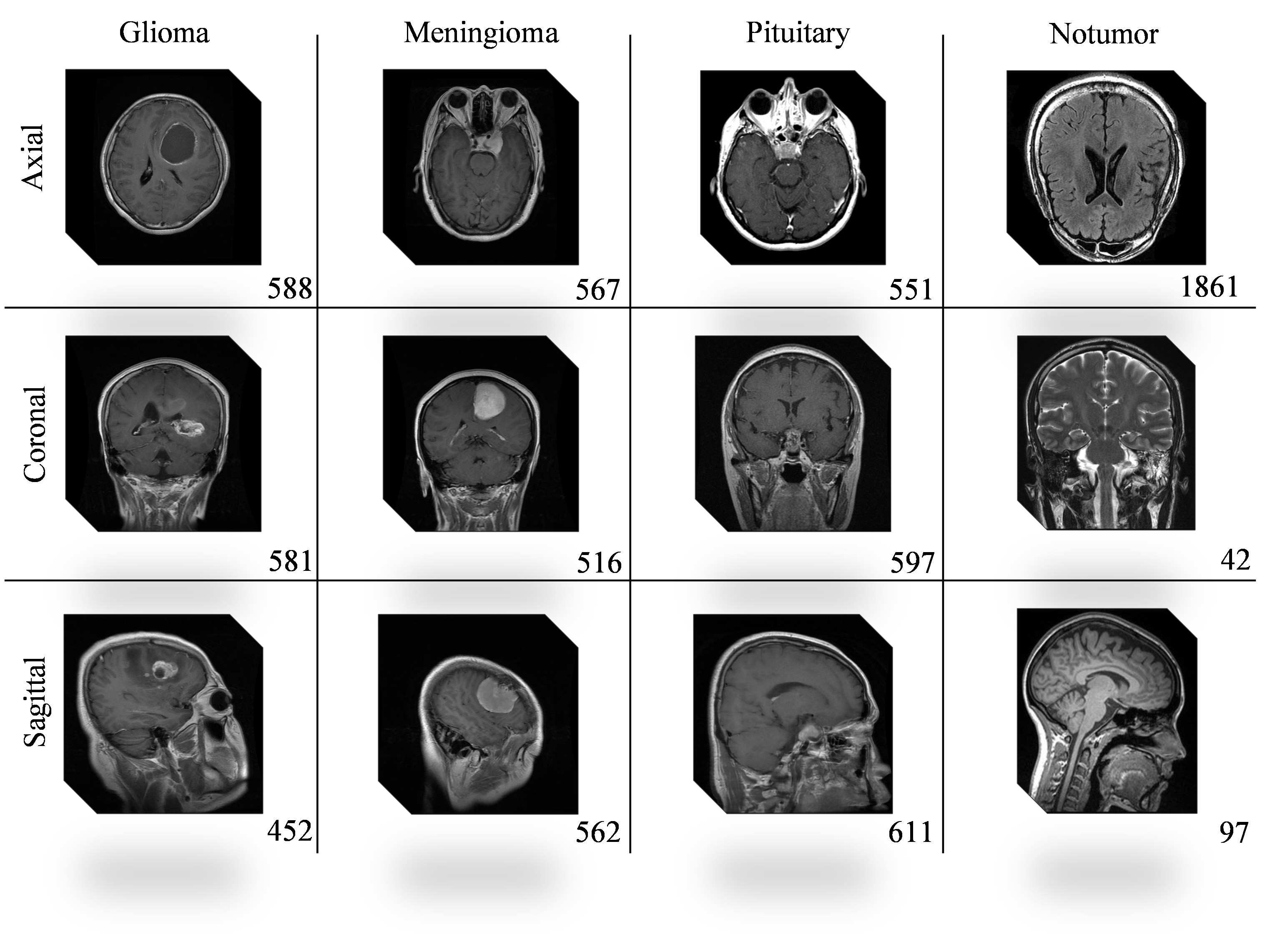}
    \caption{Sample images of each cancer type along each view (Axial || Coronal || Sagittal) overview}
    \label{fig:se.png}
\end{figure}
To ensure a fair and unbiased evaluation of the ShallowFed model performance across all tumor categories, we maintained a consistent distribution of the test dataset in multi-views, which was originally around 20\% per class, throughout all model assessments \cite{ali2022comprehensive, hekmat2025brain} Moreover, to optimize the model training process and prevent overfitting, we strategically partitioned the dataset into separate subsets: a training set (approximately 70\%) for ShallowFed model fitting, and a validation set (10\%) to evaluate model performance during training. 

For the single-view representation, we organized each view into separate folders and then allocated 20\% of the total dataset for each cancer type to serve as a dedicated test set, distinct from the training and validation datasets. This approach ensures a reasonable and balanced evaluation by preserving the integrity of each cancer type representation across all views. Table 3 presents the distribution of both multi-view and single-view representations across the three sets: training, validation, and test. This division allowed for effective tuning and early stopping, which further enhanced the model ability to generalize to new, unseen data. This systematic organization guarantees that the ShallowFed model performance is assessed in a manner that accurately reflects its generalization ability while maintaining consistency across all cancer types.

\begin{table}[h!]
\centering
\caption{Presents the distribution of class samples across four distinct image types available in the Kaggle dataset. These image types correspond to different anatomical views or angles from which the brain images were captured. Specifically, the axial view represents images taken from a top-down perspective of the brain, the coronal view captures the brain from the back, and the sagittal view displays images taken from the side, highlighting the left and right regions.  Similarly, in all views, all three types of same image are taken simultaneously}
\begin{tabular}{lllll}
\hline
\textbf{Different anatomical views} & \textbf{Class} & \textbf{Train} & \textbf{Validation} & \textbf{Test} \\ \hline
  All views & Glioma & 1189 & 132 & 300 \\  
   & Meningioma & 1205 & 134 & 306 \\  
   & Normal & 1435 & 160 & 405 \\ \ 
   & Pituitary & 1311 & 146 & 300 \\ \hline
  Axial View & Glioma & 476 & 53 & 59 \\  
   & Meningioma & 485 & 51 & 58 \\  
   & Normal & 1509 & 166 & 186 \\  
   & Pituitary & 446 & 50 & 55 \\ \hline
  Coronal View & Glioma & 470 & 53 & 58 \\  
   & Meningioma & 418 & 46 & 52 \\  
   & Normal & 34 & 4 & 4 \\  
   & Pituitary & 483 & 54 & 60 \\ \hline
  Sagittal View & Glioma & 367 & 40 & 45 \\  
   & Meningioma & 455 & 51 & 56 \\  
   & Normal & 78 & 9 & 10 \\  
   & Pituitary & 495 & 55 & 61 \\ \hline
\end{tabular}
\label{tab:distribution}
\end{table}
\subsection{Preprocessing and dataset balancing}
Before training the ShallowFed model, preprocessing of the images is essential to ensure optimal performance. This step involves removing irrelevant areas from the MRI images, allowing the model to focus solely on the relevant regions, which in turn enhances the accuracy of tumor detection. To avoid bias and distortion caused by variations in image dimensions, all images are standardized to a consistent input size of 224 × 224 pixels. This standardization ensures that the ShallowFed can extract features uniformly across multi-views to single-view, improving its ability to generalize across different datasets. Additionally, preprocessing helps to mitigate common issues associated with MRI scans, such as noise in image intensity, thus contributing to more stable and reliable ShallowFed performance. To further enhance the model robustness, the pixel values are normalized to a range of [0, 1], facilitating more effective training. These preprocessing steps collectively ensure that the model is trained on high-quality, consistent data, which is crucial for achieving precise tumor detection and improving the model overall reliability.

In this study, we employed three distinct augmentation strategies to enhance the diversity of our dataset. Table 4 presents the distribution of images after applying dataset balancing, ensuring an equitable representation across multi-views to single-view for each category. Additionally, Fig 5 illustrates the visual comparison between the original and augmented images, providing a clear depiction of the augmentation processes. This approach not only helps in increasing the model exposure to varied data but also supports its ability to generalize effectively across different tumor categories. By preserving equal class proportions, we mitigated the risk of drawing skewed that could arise from class imbalance, thereby offering a more reliable indication of the model generalization capabilities. This approach ensures that the model performance reflects its ability to handle diverse cases rather than being disproportionately influenced by overrepresented or underrepresented classes.
\begin{figure}[h]
    \centering
    \includegraphics[width = 8cm]{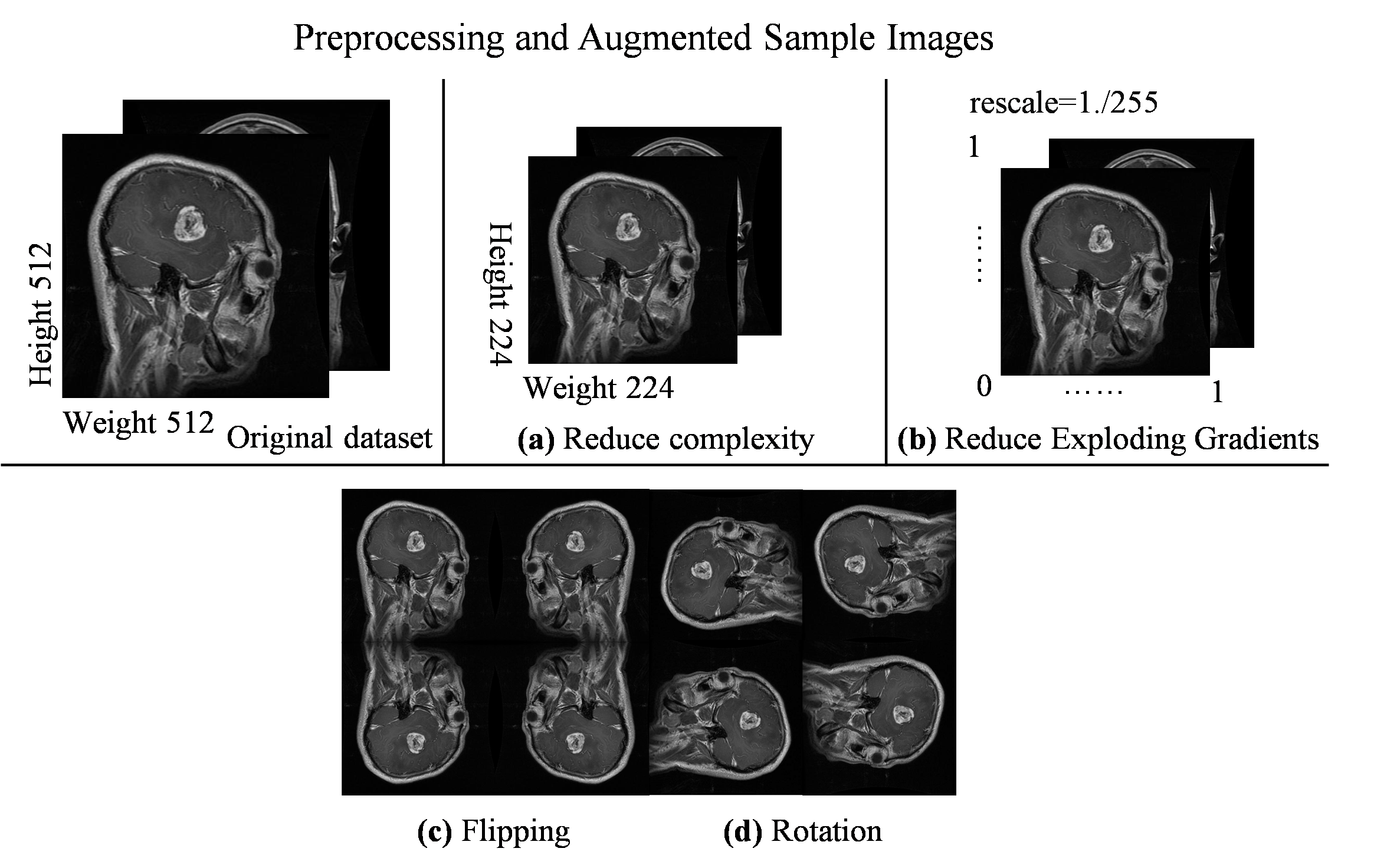}
    \caption{An overview of the MRI sample images following preprocessing and augmentation steps}
    \label{fig:se.png}
\end{figure}
\begin{table}[h!]
\centering
\caption{Shows the balanced distribution of class samples in the train and validation datasets across four distinct image types from the Kaggle dataset. These image types represent different anatomical views: axial (top-down), coronal (back), and sagittal (side, left and right regions). All three types of images are captured simultaneously in each view.}
\begin{tabular}{lllcc}
\hline
\textbf{Anatomical views}  && \textbf{Class} & \textbf{Train} & \textbf{Validation} \\
\hline
\multirow{4}{*}{All views} & \multirow{4}{*}{} & Glioma & 1435 & 160 \\
 &  & Meningioma & 1435 & 160 \\
 &  & Normal & 1435 & 160 \\
 &  & Pituitary & 1435 & 160 \\
\hline
\multirow{4}{*}{Axial View} & \multirow{4}{*}{} & Glioma & 1509 & 166 \\
 &  & Meningioma & 1509 & 166 \\
 &  & Normal & 1509 & 166 \\
 &  & Pituitary & 1509 & 166 \\
\hline
\multirow{4}{*}{Coronal View} & \multirow{4}{*}{} & Glioma & 483 & 54 \\
 &  & Meningioma & 483 & 54 \\
 &  & Normal & 483 & 54 \\
 &  & Pituitary & 483 & 54 \\
\hline
\multirow{4}{*}{Sagittal View} & \multirow{4}{*}{} & Glioma & 495 & 55 \\
 &  & Meningioma & 495 & 55 \\
 &  & Normal & 495 & 55 \\
 &  & Pituitary & 495 & 55 \\
\hline
\end{tabular}
\label{tab:distribution}
\end{table}
\subsection{Classification Performance of proposed FOLC-Net: Multi-View to Single-View Representation}
In this section, we present the results of the FOLC-Net framework across both multi-view and single-view evaluations. By evaluating the model in both combined and individual views, we not only provide a comprehensive assessment of its performance but also highlight its potential in diverse representations. This approach offers a more nuanced understanding of the model capabilities, especially considering that most previous studies \cite{guan2024federated, islam2023federated} have focused solely on multi-view evaluations, neglecting the valuable insights that single-view assessments can provide. 
\bmhead{Multi-View Performance Comparison} 
The FOLC-Net performance across all views demonstrates impressive results (Table 5), with the highest accuracy achieved in the Axial View (99.44\%), closely followed by Coronal View (98.27\%) and Sagittal View (92.44\%). The Axial View outperforms the other views, showing the highest F1-score (1.00) for both Glioma and Normal classes, as well as consistent high performance in all tumor categories. While the Axial View shows the best overall accuracy, each view excels in specific tumor categories, indicating the model ability to capture distinct features depending on the orientation of the images. For instance: Glioma and Normal classes have best precision and recall in the Axial and Coronal views, suggesting that these views are particularly beneficial for classifying these categories. The Pituitary class also performs consistently well across all views, with high recall and precision. Moreover, Sagittal View, while showing slightly lower overall accuracy (92.44\%), excels in the Meningioma and Pituitary categories, with the highest recall for Pituitary (1.00). This indicates that the Sagittal View, though lower in overall accuracy, may be particularly effective in distinguishing certain tumor types, especially Pituitary and Meningioma.
\bmhead{View-by-View Comparison}
The Axial View achieved the highest overall results for the Glioma, Normal, and Pituitary categories, with best precision and recall for most classes. It consistently performs the best for these categories, making it the most reliable view for these tumor types. The Coronal View shows a slight reduce in accuracy compared to the Axial View (98.27\% vs. 99.44\%). However, it still provides strong performance for most categories, with the highest F1-score for Meningioma (0.97) and efficient recall for Pituitary (1.00). This suggests that while the Axial View may perform better overall, the Coronal View is still a solid alternative, especially for Meningioma. The Sagittal View demonstrates the lowest accuracy (92.44\%) among the three views. However, it performs extremely well for Pituitary (1.00 recall, 0.99 F1-score) and Meningioma (0.87 F1-score). Although its accuracy is lower, it still provides valuable insights for specific tumor types and is especially effective in cases where other views may not perform as well.
\begin{table}[h!]
\centering
\caption{Class-wise performance comparison of ShallowFed model across different anatomical: Kaggle Multi}
\begin{tabular}{lllllll}
\hline
\textbf{Model} & \textbf{Anatomical views} & \textbf{Class}   & \textbf{Precision} & \textbf{Recall} & \textbf{F1-Score} & \textbf{Accuracy} \\ \hline
\multirow{4}{*}{FOLC-Net} & \multirow{4}{*}{All views} & Glioma   & 0.99  & 0.96  & 0.97  & 98.01  \\ 
 &  & Meningioma & 0.96  & 0.96  & 0.96  &       \\  
 &  & Normal   & 0.99  & 1.00  & 1.00  &      \\ 
 &  & Pituitary & 0.98  & 0.99  & 0.99  &       \\ \hline
\multirow{4}{*}{FOLC-Net} & \multirow{4}{*}{Axial View} & Glioma   & 1.00  & 1.00  & 1.00  & 99.44  \\  
 &  & Meningioma & 0.98  & 0.98  & 0.98  &       \\ 
 &  & Normal   & 1.00  & 1.00  & 1.00  &       \\ 
 &  & Pituitary & 0.98  & 0.98  & 0.98  &       \\ \hline
\multirow{4}{*}{FOLC-Net} & \multirow{4}{*}{Coronal View} & Glioma   & 1.00  & 1.00  & 1.00  & 98.27  \\ 
 &  & Meningioma & 0.98  & 0.96  & 0.97  &       \\  
 &  & Normal   & 1.00  & 0.75  & 0.86  &       \\ 
 &  & Pituitary & 0.97  & 1.00  & 0.98  &       \\ \hline
\multirow{4}{*}{FOLC-Net} & \multirow{4}{*}{Sagittal View} & Glioma   & 0.87  & 1.00  & 0.93  & 92.44  \\ 
 &  & Meningioma & 1.00  & 0.77  & 0.87  &       \\  
 &  & Normal   & 0.67  & 1.00  & 0.80  &       \\ 
 &  & Pituitary & 0.98  & 1.00  & 0.99  &       \\ \hline
\end{tabular}
\label{tab:shallowfed_performance}
\end{table}
The confusion matrix in fig 6, a clear and detailed evaluation of the ShallowFed model performance, highlighting the accuracy, misclassifications, and overall effectiveness for each class in combine and separate view representation.

All Views (Fig a): Best result has achieved in Normal (405) and Pituitary (299) classifications have the highest correct predictions across all views, with very few misclassifications (errors).
\bmhead{Key Observations}
\begin{enumerate}
    \item The ShallowFed model performs well in identifying the Normal and Pituitary classes, with high accuracy.
    \item Glioma (281 correctly predicted) and Meningioma (296 correctly predicted) also perform well but with slightly higher misclassification compared to Normal and Pituitary.
\end{enumerate}
Axial View (Fig b): Best result has highlighted in the Normal class (186 correct predictions) and Pituitary (54 correct) are well predicted in this view.
\bmhead{Key Observations}
\begin{enumerate}
    \item Glioma and Meningioma show better accuracy than in the All-Views matrix.
    \item The Pituitary class has the most significant number of misclassifications in this view (1 misclassified).
\end{enumerate}
\bmhead{Comparative Insight: }
Axial view improves the detection of Glioma and Meningioma compared to All Views, while the Pituitary class remains less accurate. Moreover, Axial view yields the most accurate results for Normal and Meningioma classifications, making it stand out compared to sagittal-view and coronal-view.

Coronal View (Fig c): Best result has presents in Glioma (58 correct) and Pituitary (60 correct) classifications are solid here, but Normal (50 correct) and Meningioma predictions suffer from misclassifications (1 and 3 misclassified, respectively).
\bmhead{Key Observations}
\begin{enumerate}
    \item The Glioma and Pituitary predictions perform best in the Coronal view.
    \item Meningioma suffers with relatively poor prediction accuracy in this view.
\end{enumerate}
\bmhead{Comparative Insight: }
This view is better at classifying Pituitary but not as effective as others in classifying Normal. In addition to it. Coronal view shows strong performance in Glioma predictions but suffers in Meningioma, with relatively lower accuracy for these classes as compared to axial-view and sagittal-view.
Sagittal View (Fig d): In this view Glioma (75 correct) performs significantly well understand by ShallowFed model, with Meningioma and Pituitary suffering slight errors.
\bmhead{Key Observations}
\begin{enumerate}
    \item The Meningioma and Pituitary classes have lower accuracy here, with notable misclassification errors (7 and 1, respectively).
    \item The Normal class performs the worst in this view (43 correct predictions, 5 misclassified).
\end{enumerate}
\bmhead{Comparative Insight: }
Sagittal view performs well for Glioma with the highest accuracy in this class, but struggles with Normal and Pituitary as compared to others’ views. 
\bmhead{Reasons for Accuracy Differences}
Axial view excels in understanding the brain structure, which aids in better classifying Normal and Meningioma. Sagittal view strong performance in Glioma can be attributed to its ability to capture frontal sections of the brain effectively, though it underperforms in other classes. Coronal view, while effective in detecting Pituitary, is less accurate in Normal and Meningioma, possibly due to the complexity of classifying these regions from a side perspective. Considering all views, the Axial view provides the most balanced and accurate classification. However, combining all views assumes that the model is robust and unbiased, offering a comprehensive evaluation of its performance across different brain regions still missing in literature.
\begin{figure}[!h]
\centering
\begin{minipage}[]{0.46\textwidth}
  \centering
  \includegraphics[width = \textwidth]{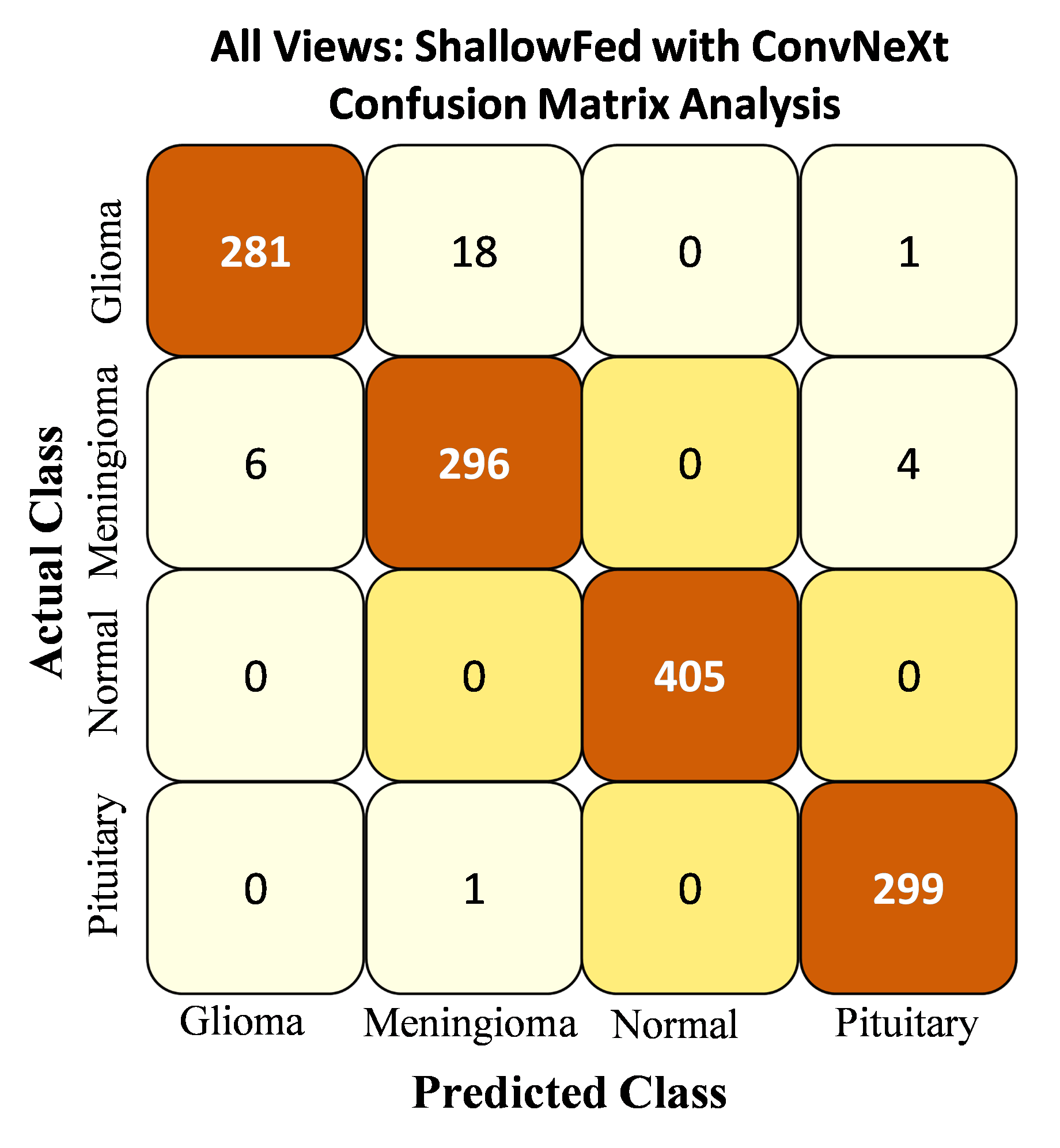}
  
    \begin{center}
     \textbf{a}    
     \end{center}
\end{minipage}
\begin{minipage}[]{0.46\textwidth}
  \centering
  \includegraphics[width = \textwidth]{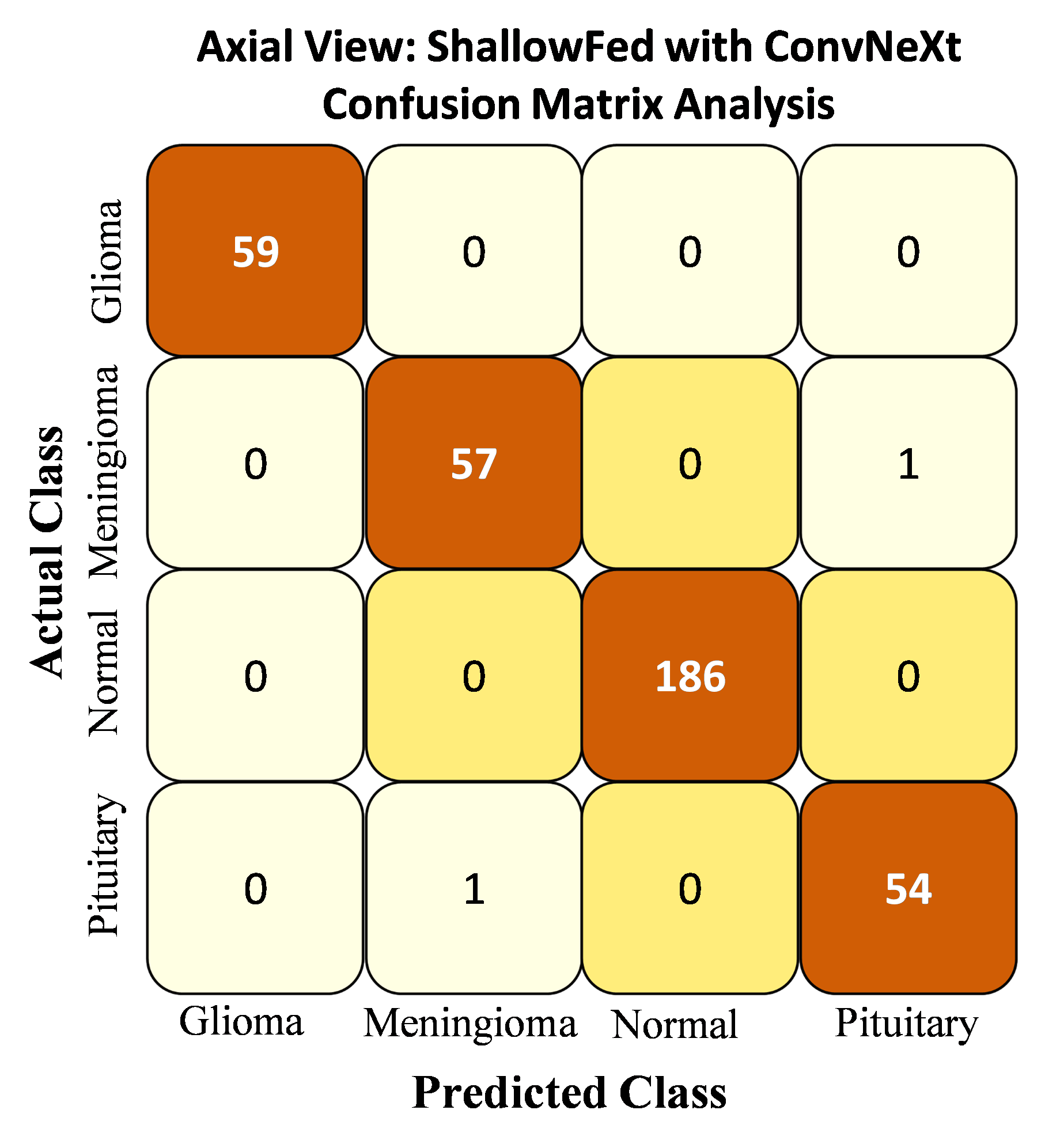}
 
     \begin{center}
     \textbf{b}    
     \end{center}
\end{minipage}
\begin{minipage}[]{0.46\textwidth}
  \centering
  \includegraphics[width = \textwidth]{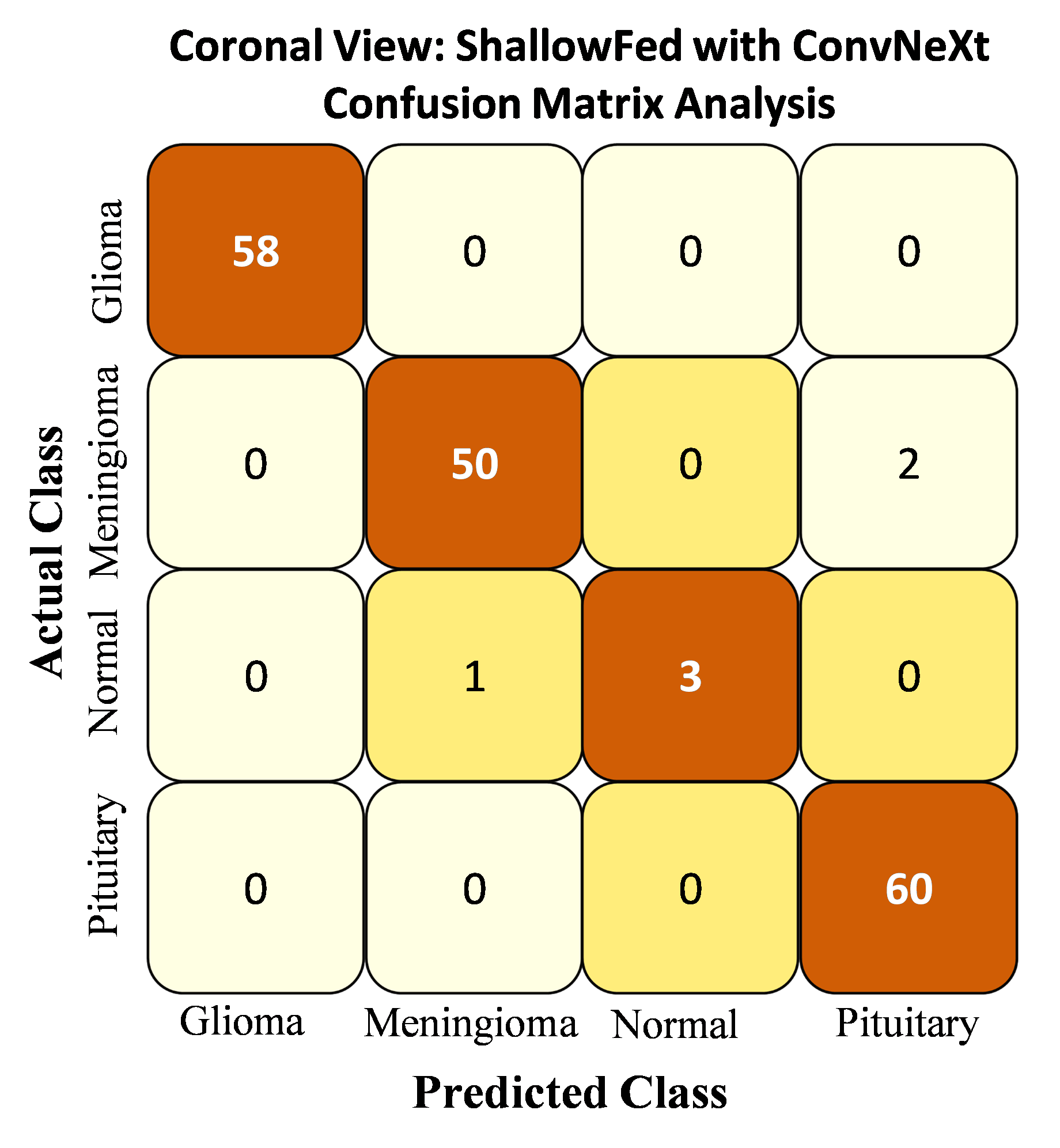}
  
    \begin{center}
     \textbf{c}    
     \end{center}
\end{minipage}
\begin{minipage}[]{0.46\textwidth}
  \centering
  \includegraphics[width = \textwidth]{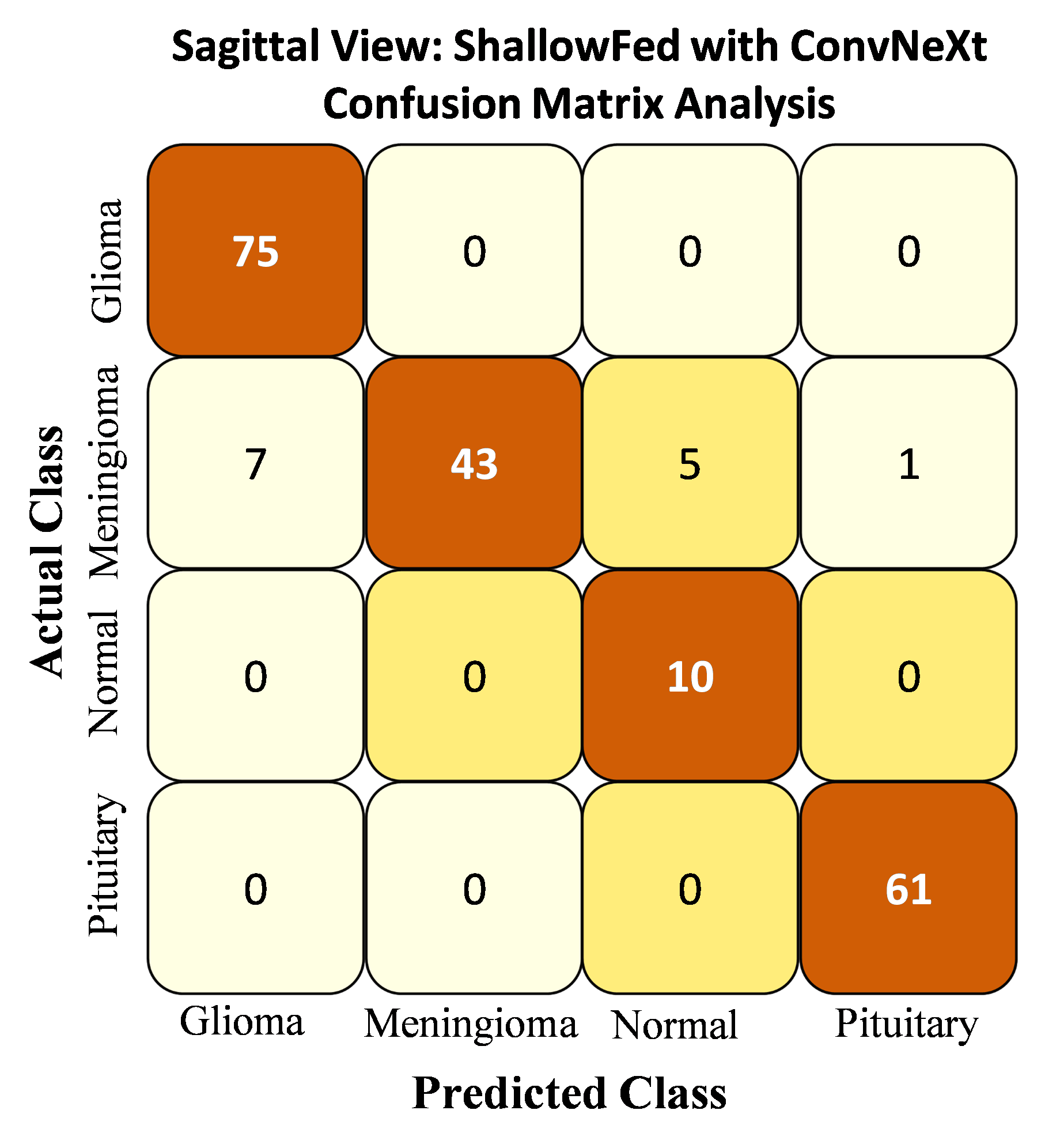}
 
     \begin{center}
     \textbf{d}    
     \end{center}
\end{minipage}
  \caption{An overview of the confusion matrices for proposed ShallowFed with ConvNeXt (FOLC-Net): (a) All-Views, (b) Axial View, (c) Coronal View, and (d) Sagittal View }
  \label{fig:11.png}
\end{figure}
\bmhead{Comparison of All-Views vs Individual Views}
All Views performance analysis clearly shows that ShallowFed with ConvNeXt outperforms the ShallowFed CNN Model in both the Receiver Operating Characteristic (ROC) and Precision-Recall (P-R) Curves. For the All Views (fig 7 a and b), ShallowFed with ConvNeXt achieves significantly higher performance with an AUC of 0.9987 and an AP of 0.9958 compared to the ShallowFed CNN Model, which has an AUC of 0.9985 and an AP of 0.9952. This suggests that incorporating ConvNeXt into the ShallowFed model improves the overall performance in multi-view evaluation, with both ROC and P-R curves showing a more pronounced area under the curve.
\bmhead{Comparison Between Individual Views (Axial, Coronal, and Sagittal)}
When examining the Axial, Coronal, and Sagittal views individually, the ShallowFed with ConvNeXt model continues to show superior performance. In the Axial View (fig 7 c and d), ShallowFed with ConvNeXt achieves an AUC of 1.0000 and an AP of 0.9999, outperforming the ShallowFed CNN Model with an AUC of 0.9999 and AP of 0.9997. Similarly, in the Coronal View (fig 7 e and f), the ShallowFed with ConvNeXt model achieves an AUC of 0.9998 and an AP of 0.9970, surpassing the ShallowFed CNN Model with an AUC of 0.9945 and AP of 0.9795. Lastly, for the Sagittal View (fig 7 g and h), both models show slightly lower performance compared to the Axial and Coronal views, but ShallowFed with ConvNeXt still outperforms the baseline ShallowFed CNN Model with an AUC of 0.9842 and AP of 0.9970, while the ShallowFed CNN Model achieves an AUC of 0.9762 and an AP of 0.9240. This consistent trend of higher AUC and AP values in ShallowFed with ConvNeXt across all views demonstrates the effectiveness of the ConvNeXt architecture in improving model performance, particularly for finer anatomical distinctions in the Axial, Coronal, and Sagittal views.
\begin{figure}[h!]
\centering
\begin{minipage}[]{0.40\textwidth}
  \centering
  \includegraphics[width = \textwidth]{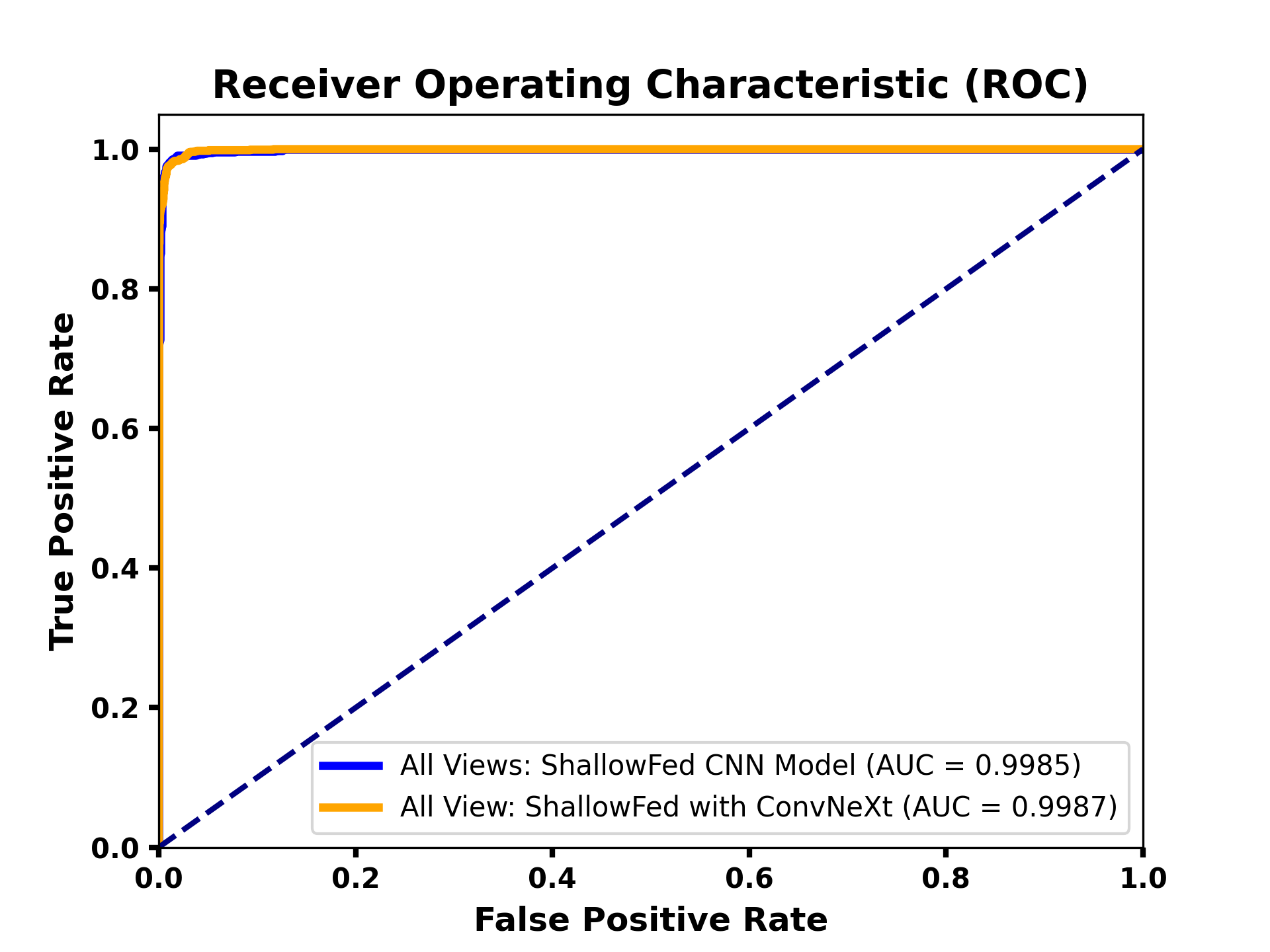}
  
    \begin{center}
     \textbf{a}    
     \end{center}
\end{minipage}
\begin{minipage}[]{0.40\textwidth}
  \centering
  \includegraphics[width = \textwidth]{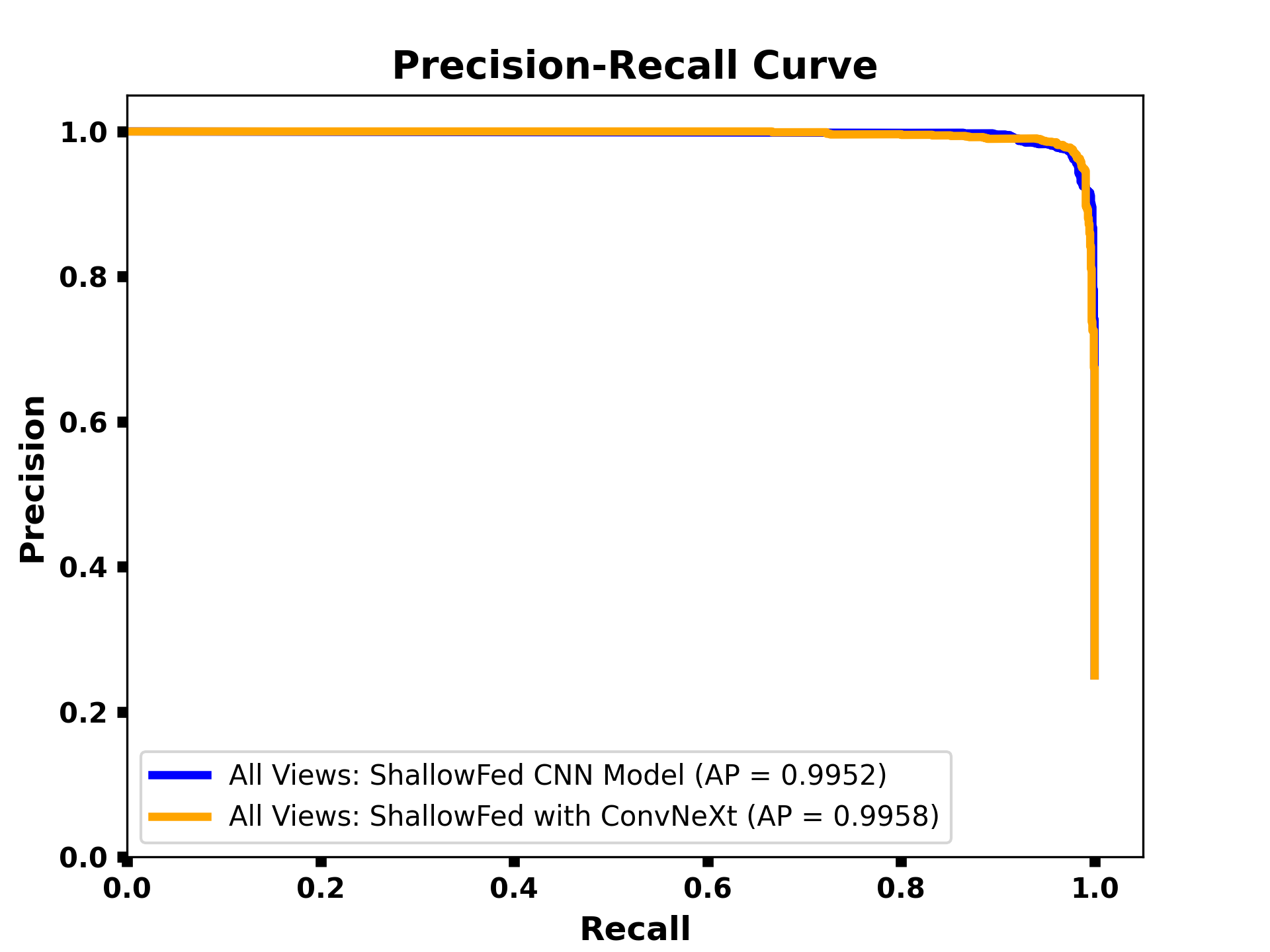}
 
     \begin{center}
     \textbf{b}    
     \end{center}
\end{minipage}
\begin{minipage}[]{0.40\textwidth}
  \centering
  \includegraphics[width = \textwidth]{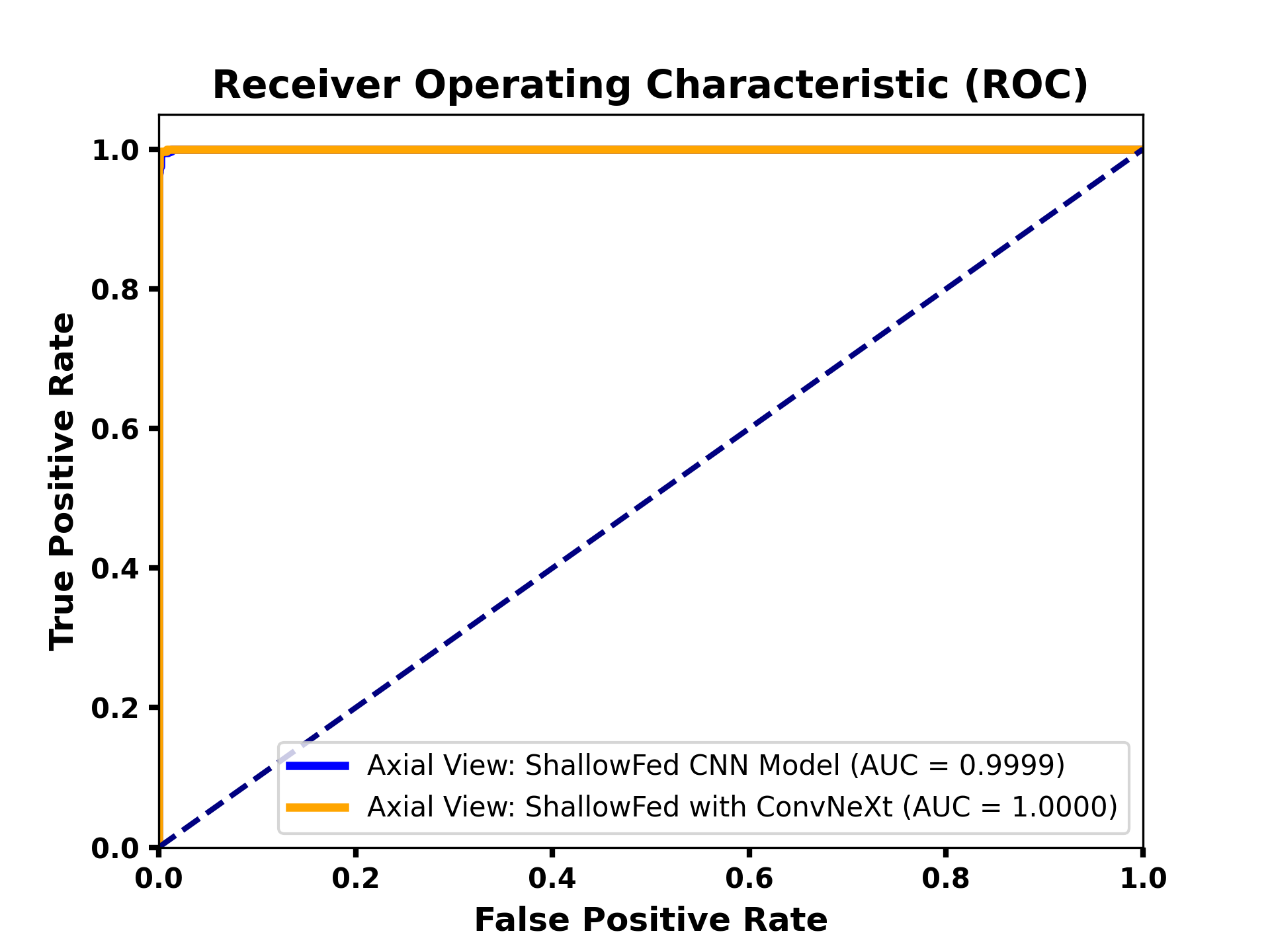}
  
    \begin{center}
     \textbf{c}    
     \end{center}
\end{minipage}
\begin{minipage}[]{0.40\textwidth}
  \centering
  \includegraphics[width = \textwidth]{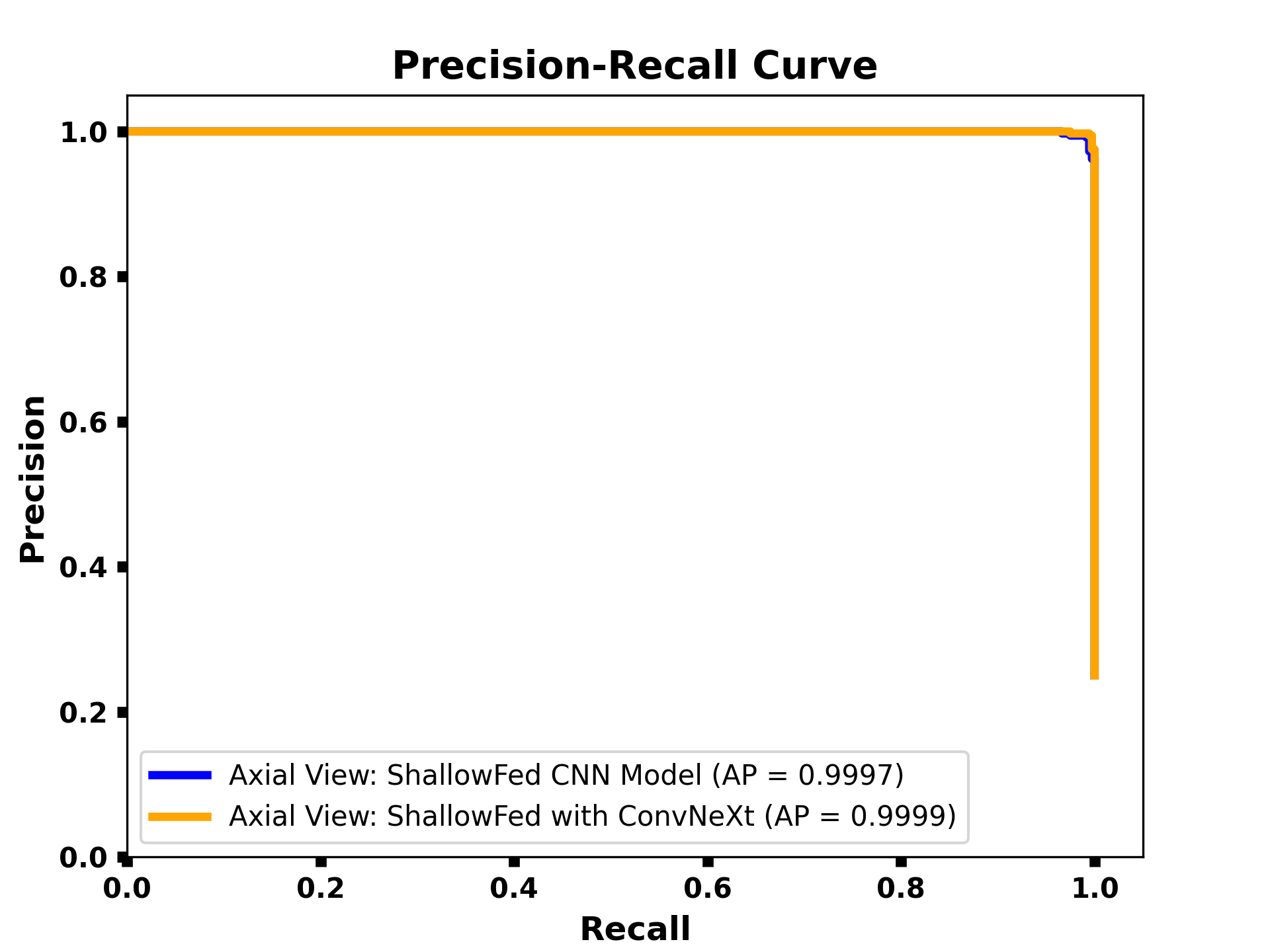}
 
     \begin{center}
     \textbf{d}    
     \end{center}
\end{minipage}
\begin{minipage}[]{0.40\textwidth}
  \centering
  \includegraphics[width = \textwidth]{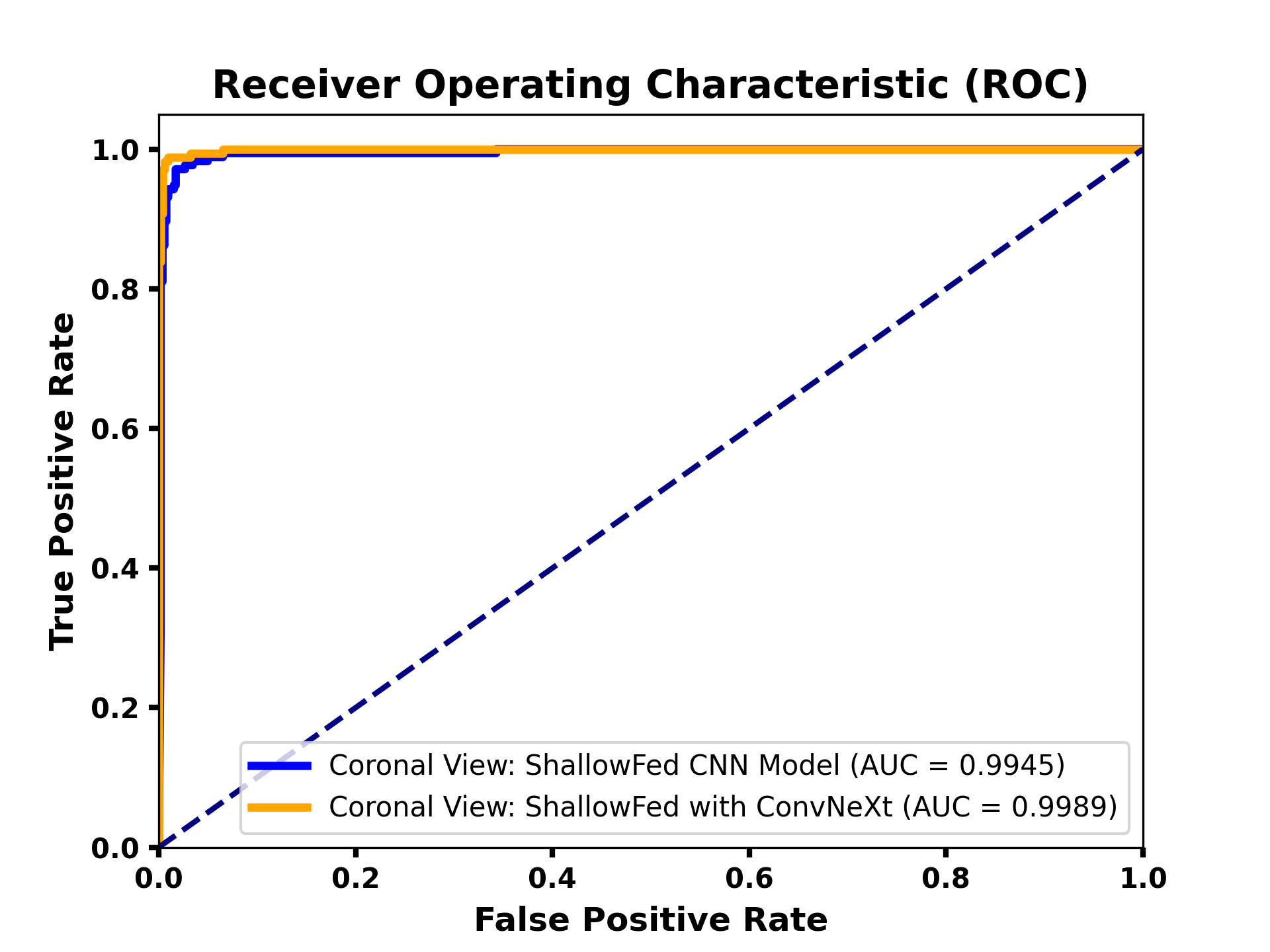}
  
    \begin{center}
     \textbf{e}    
     \end{center}
\end{minipage}
\begin{minipage}[]{0.40\textwidth}
  \centering
  \includegraphics[width = \textwidth]{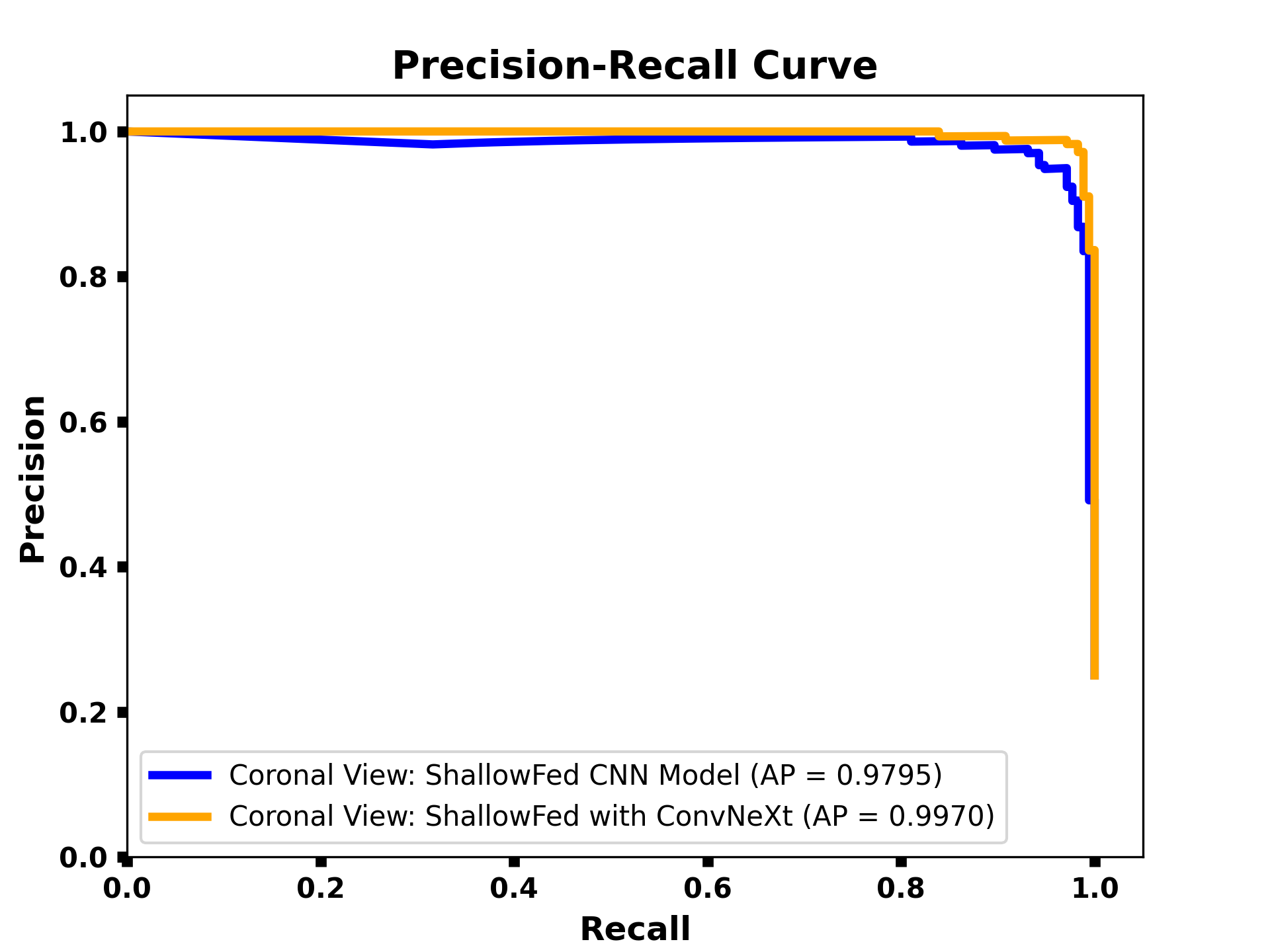}
 
     \begin{center}
     \textbf{f}    
     \end{center}
\end{minipage}
\begin{minipage}[]{0.40\textwidth}
  \centering
  \includegraphics[width = \textwidth]{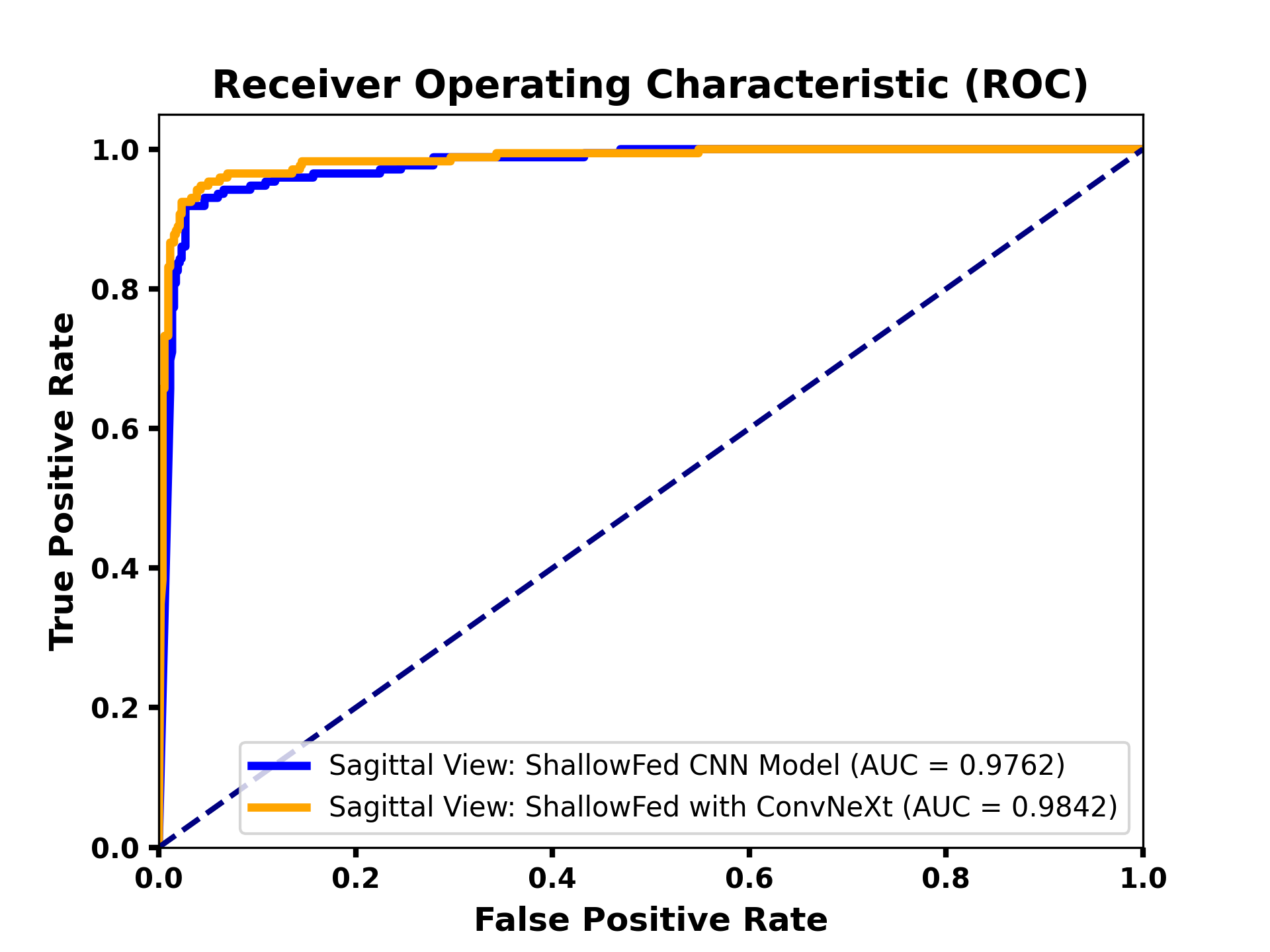}
  
    \begin{center}
     \textbf{g}    
     \end{center}
\end{minipage}
\begin{minipage}[]{0.40\textwidth}
  \centering
  \includegraphics[width = \textwidth]{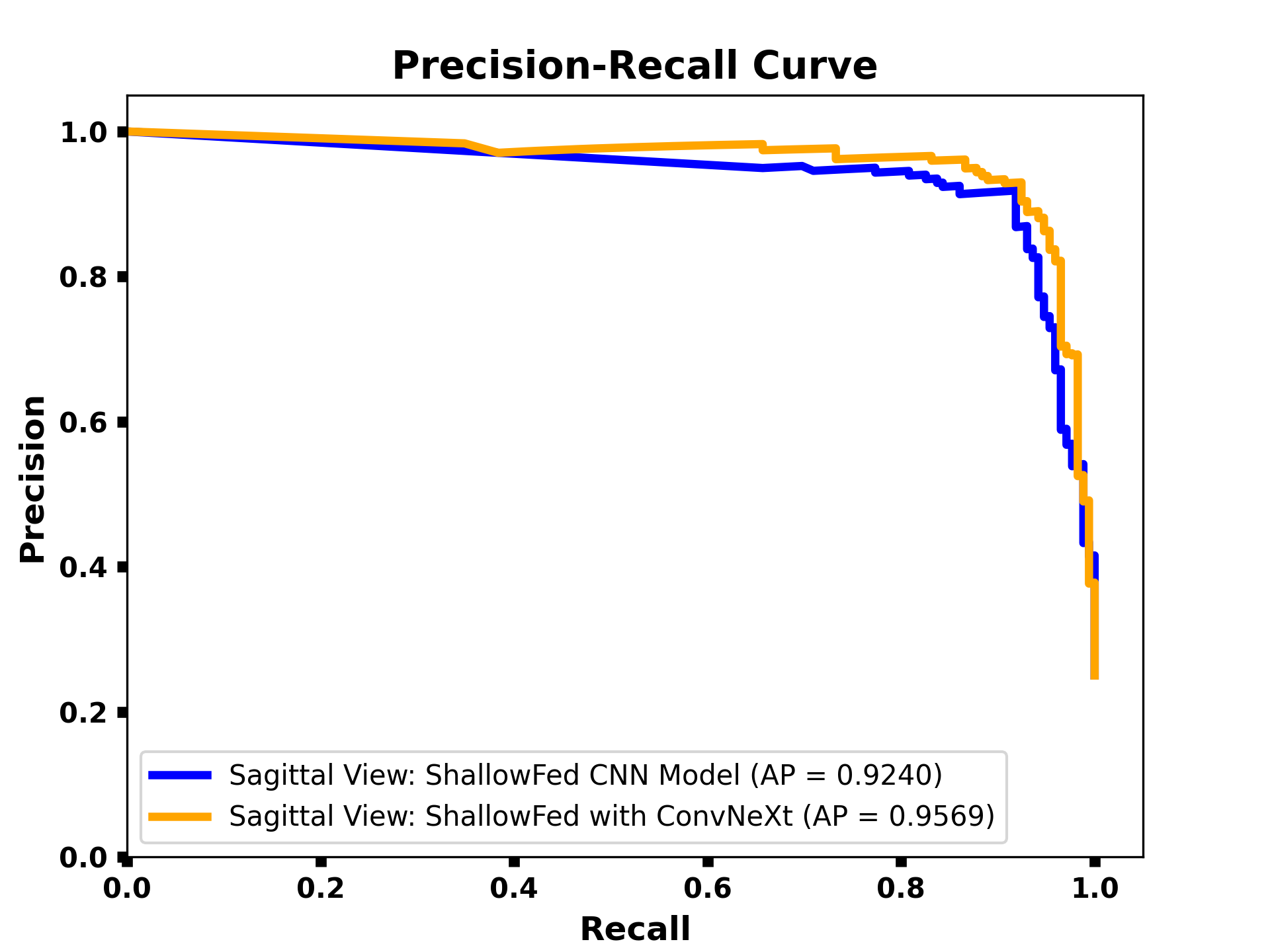}
     \begin{center}
     \textbf{h}    
     \end{center}
\end{minipage}
  \caption{An overview of the Baseline CNN and the proposed ShallowFed with ConvNeXt (FOLC-Net), evaluated using ROC and PR curves: (a-b) All Views, (c-d) Axial View, (e-f) Coronal View, and (g-h) Sagittal View}
  \label{fig:11.png}
\end{figure}
\bmhead{Comparative Insight: }
The performance comparison between All Views and the individual Axial, Coronal, and Sagittal views highlight the superiority of the ShallowFed with ConvNeXt model across all scenarios. While the All-Views analysis demonstrates a strong overall performance with an AUC of 0.9987 and an AP of 0.9958, individual views provide more coarse insights. The Axial View achieves perfect AUC and near-perfect AP (1.0000 and 0.9999), indicating exceptional model performance. Similarly, the Coronal View follows closely with an AUC of 0.9998 and an AP of 0.9970. The Sagittal View, although showing slightly lower performance than the other two, still exhibits a marked improvement over the baseline ShallowFed CNN Model, with an AUC of 0.9842 and an AP of 0.9970. In all views, the ShallowFed with ConvNeXt model consistently outperforms the baseline, underscoring its enhanced ability to capture complex features across multiple anatomical perspectives.
\bmhead{Outcome: }
The performance analysis reinforces our hypothesis that validating a model with individual views, rather than solely relying on an all-encompassing MRI representation, is crucial for a comprehensive evaluation. The results clearly demonstrate that while the All-Views approach offers a solid performance metric, individual views Axial, Coronal, and Sagittal reveal important variations in model performance, with some views showing higher accuracy than others. For instance, the Axial View achieved near-perfect results, while the Sagittal View showed relatively lower performance, despite still outperforming the baseline model. These findings suggest that testing models on individual views provides a more detailed and nuanced assessment, ensuring that the model performance is fully validated across different perspectives rather than assuming it will generalize equally well across all views. This approach is vital for ensuring robustness and identifying potential weaknesses in model performance specific to certain MRI orientations.
\subsection{Superior Performance Across Multi-View to Single-Views, Outperforming Existing Models}
This section presents a comprehensive performance comparison between the proposed ShallowFed (FOLC-Net) model and a range of existing CNN architectures. Evaluations are conducted across All Views as well as individual Axial, Coronal, and Sagittal views. Metrics such as accuracy, precision, recall, and F1-score are used to assess classification performance on the Kaggle Multi dataset (Table 6).
\bmhead{Performance Comparison across All Views}
The ShallowFed (FOLC-Net) model achieves the highest overall performance across all existing CNN model in the All-Views configuration. It achieves an accuracy of 98.01\%, outperforming the best-performing CNN models such as MobileNetV2 (97.10\%), DenseNet169 (97.40\%), and VGG16 (97.10\%). In terms of precision, recall, and F1-score, ShallowFed also leads with 97.92\%, 97.85\%, and 97.88\% respectively. These results highlight the superior generalization and feature extraction capabilities of ShallowFed when aggregating information from all views, confirming its robustness compared to traditional CNN architectures.
\bmhead{Performance Comparison across Individual Views}
For the Axial View, ShallowFed achieves the best performance with 99.44\% accuracy and a notably high F1-score of 99.11\%, surpassing all existing CNN including MobileNetV2 (99.16\%) and DenseNet201 (99.16\%). In the Coronal View, while several existing CNN like VGG16 (95.40\%) and MobileNetV1 (95.97\%) show strong results, ShallowFed maintains a leading accuracy of 98.27\% with a competitive F1-score of 95.29\%, reflecting its consistency across more challenging views. In the Sagittal View, which generally exhibits lower performance across all models, ShallowFed still outperforms the others with 92.44\% accuracy and a strong F1-score of 89.70\%, exceeding the next best NASNetLarge (91.86\% accuracy, 88.05\% F1-score). These results confirm ShallowFed effectiveness in handling individual view modalities and its capability to maintain high classification performance even in less favorable conditions.
\bmhead{Comparative Insight: }
Existing CNN architectures tend to underperform on single-view inputs, particularly in Coronal and Sagittal views, due to their deep and rigid structures that often overfit to dominant features present in multi-view or axial-rich datasets. These models lack adaptability when faced with the sparse or less discriminative features typically found in single-view disease images. In contrast, the proposed ShallowFed (FOLC-Net) leverages a novel shallow and lightweight architecture optimized for distributed environments through FL. This design enhances generalization by preventing overfitting to view-specific biases and promotes robust feature learning across heterogeneous data sources. Federated optimization ensures that the model captures diverse structural patterns from decentralized clients, enabling consistent and superior performance across all individual views, including those where conventional CNNs struggle.
\begin{table}[h!]
\centering
\caption{Performance comparison of ShallowFed and existing CNN model across different anatomical views: Kaggle Multi}
\begin{tabular}{llcccc}
\hline
\textbf{Model} & \textbf{Anatomical views} & \textbf{Accuracy} & \textbf{Precision} & \textbf{Recall} & \textbf{F1-Score} \\
\hline
Base- DensNet121 & All views & 97.10 & 96.86 & 96.80 & 96.86 \\
Base- DensNet169 & All views & 97.40 & 97.30 & 97.21 & 97.21 \\
Base- DensNet201 & All views & 97.33 & 97.16 & 97.10 & 97.12 \\
Base- MobileNetV1 & All views & 96.33 & 96.17 & 96.02 & 96.08 \\
Base- MobileNetV2 & All views & 97.10 & 97.55 & 97.51 & 97.51 \\
Base- ResNet50 & All views & 83.29 & 82.56 & 82.67 & 82.39 \\
Base- ResNet101 & All views & 88.40 & 87.63 & 87.56 & 87.57 \\
Base- ResNet152 & All views & 88.71 & 88.66 & 87.90 & 87.96 \\
Base-VGG16 & All views & 97.10 & 96.91 & 96.85 & 96.88 \\
Base-VGG19 & All views & 96.10 & 95.80 & 95.79 & 95.78 \\
Base- NASNetMobile & All views & 96.18 & 95.96 & 95.86 & 95.91 \\
Base- NASNetLarge & All views & 95.49 & 95.26 & 95.10 & 95.13 \\
ShallowFed (FOLC-Net) & All views & 98.01 & 97.92 & 97.85 & 97.88 \\
Base- DensNet121 & Axial View & 99.16 & 98.69 & 98.69 & 98.69 \\
Base- DensNet169 & Axial View & 99.16 & 98.97 & 98.66 & 98.69 \\
Base- DensNet201 & Axial View & 99.16 & 98.98 & 98.71 & 98.84 \\
Base- MobileNetV1 & Axial View & 99.16 & 98.69 & 98.69 & 98.69 \\
Base- MobileNetV2 & Axial View & 99.16 & 98.70 & 99.01 & 98.85 \\
Base- ResNet50 & Axial View & 96.92 & 95.86 & 95.85 & 95.76 \\
Base- ResNet101 & Axial View & 96.64 & 95.73 & 96.03 & 95.82 \\
Base- ResNet152 & Axial View & 94.97 & 93.15 & 94.01 & 93.49 \\
Base-VGG16 & Axial View & 98.60 & 98.14 & 98.13 & 98.13 \\
Base-VGG19 & Axial View & 98.60 & 98.13 & 98.12 & 98.12 \\
Base- NASNetMobile & Axial View & 97.76 & 96.78 & 97.67 & 97.16 \\
Base- NASNetLarge & Axial View & 98.32 & 97.50 & 98.61 & 98.01 \\
ShallowFed (FOLC-Net) & Axial View & 99.44 & 99.11 & 99.11 & 99.14 \\
Base- DensNet121 & Coronal View & 94.82 & 86.51 & 95.90 & 89.72 \\
Base- DensNet169 & Coronal View & 94.83 & 88.49 & 95.88 & 91.33 \\
Base- DensNet201 & Coronal View & 96.55 & 92.73 & 97.27 & 94.72 \\
Base- MobileNetV1 & Coronal View & 96.76 & 96.78 & 97.67 & 97.16 \\
Base- MobileNetV2 & Coronal View & 95.97 & 96.94 & 96.96 & 96.87 \\
Base- ResNet50 & Coronal View & 81.60 & 70.26 & 73.63 & 70.64 \\
Base- ResNet101 & Coronal View & 86.78 & 74.24 & 77.73 & 74.90 \\
Base- ResNet152 & Coronal View & 86.20 & 71.83 & 71.71 & 71.66 \\
Base-VGG16 & Coronal View & 95.40 & 97.14 & 90.57 & 93.00 \\
Base-VGG19 & Coronal View & 95.40 & 96.42 & 90.58 & 93.02 \\
Base- NASNetMobile & Coronal View & 96.55 & 97.30 & 97.26 & 97.27 \\
Base- NASNetLarge & Coronal View & 93.10 & 90.11 & 94.44 & 91.92 \\
ShallowFed (FOLC-Net) & Coronal View & 98.27 & 98.70 & 92.78 & 95.29 \\
Base- DensNet121 & Sagittal View & 90.11 & 84.80 & 92.39 & 86.22 \\
Base- DensNet169 & Sagittal View & 88.37 & 83.55 & 91.18 & 84.82 \\
Base- DensNet201 & Sagittal View & 90.69 & 85.91 & 90.84 & 87.27 \\
Base- MobileNetV1 & Sagittal View & 91.86 & 85.65 & 93.75 & 87.35 \\
Base- MobileNetV2 & Sagittal View & 88.95 & 83.31 & 91.55 & 84.42 \\
Base- ResNet50 & Sagittal View & 73.83 & 67.52 & 71.22 & 66.45 \\
Base- ResNet101 & Sagittal View & 79.65 & 74.45 & 79.97 & 73.41 \\
Base- ResNet152 & Sagittal View & 75.00 & 70.97 & 72.44 & 66.62 \\
Base-VGG16 & Sagittal View & 90.11 & 84.85 & 92.41 & 86.27 \\
Base-VGG19 & Sagittal View & 84.30 & 80.71 & 87.98 & 80.30 \\
Base- NASNetMobile & Sagittal View & 91.27 & 85.04 & 93.01 & 87.06 \\
Base- NASNetLarge & Sagittal View & 91.86 & 86.09 & 91.66 & 88.05 \\
ShallowFed (FOLC-Net) & Sagittal View & 92.44 & 87.89 & 94.19 & 89.70 \\
\hline
\end{tabular}
\end{table}
\subsection{Do Existing Studies Fully Explore Brain Tumor Views? A Performance Comparison with FOLC-Net}
The following analysis compares the performance of existing models in the literature (Table 7), which were validated only using the all-views approach. These existing studies did not assess performance on individual views (Axial, Coronal, and Sagittal), which may limit the understanding of model robustness in different anatomical perspectives. To ensure a fair evaluation and further validate the performance of these models, we re-build and test them on each view separately. This allows us to better assess the accuracy and generalizability of the models when applied to specific views, which is essential for real-world applications in medical imaging.
   
When comparing the Proposed Model (ShallowFed (FOLC-Net)) with existing studies, the following observations can be made:
\bmhead{1. All-Views Performance}
The ShallowFed (FOLC-Net) model achieves the highest accuracy at 98.01\%, surpassing the closest competitor, Muneeb et al.~\cite{khan2025adaptive} at 97.52\% by approximately 0.5\%. This indicates the proposed model's overall superiority in capturing patterns across all views.

\bmhead{2. Axial View}
The proposed model demonstrates exceptional performance in the axial view, with an accuracy of 99.44\%. This is a significant improvement over Khan et al.~\cite{khan2024hybrid} at 97.76\%, yielding a 1.68\% increase. The axial view accuracy is notably higher than any other model tested, reflecting the effectiveness of the proposed method in handling this view.

\bmhead{3. Coronal View}
In the coronal view, the proposed model again excels, achieving 98.27\% accuracy. This is notably higher than the second-best model, Saif et al.~\cite{khan2024boosting}, which achieved 95.97\%, representing a 2.3\% improvement. This highlights the robustness of the proposed model in handling the coronal view.

\bmhead{4. Sagittal View}
The proposed model performs strongly in the sagittal view with an accuracy of 92.44\%, outperforming the best-performing model, Saif et al.~\cite{khan2024boosting} at 90.11\%, by a margin of 2.33\%. This consistent performance across all views underscores the versatility and generalizability of the ShallowFed model.
\begin{table}[h!]
\centering
\caption{Performance comparison of ShallowFed and existing CNN model across different representations: Kaggle Multi}
\begin{tabular}{llcccc}
\hline
\textbf{Reference} & \textbf{Method} & \textbf{All-Views} & \textbf{Axial View} & \textbf{Coronal View} & \textbf{Sagittal View} \\ \hline
Khan et al \cite{khan2025detection}      & DL + Residual Learning & 96.95 & 97.76 & 95.40 & 88.37 \\ \hline
Khan et al \cite{khan2024hybrid}     & Hybrid‐NET            & 95.10 & 96.64 & 96.55 & 88.95 \\ \hline
Saif et al \cite{khan2024boosting}      & BiGait                & 96.33 & 97.76 & 95.97 & 90.11 \\ \hline
Muneeb et al \cite{khan2025adaptive}    & DL                     & 97.52 & 98.60 & 96.55 & 88.95 \\ \hline
Isunuri et al \cite{isunuri2022three}  & SC-NN                  & 97.52 & 98.32 & 95.40 & 90.11 \\ \hline
Shanthi et al \cite{shanthi2022efficient}  & CNN-LSTM Classifier    & 97.50 & 96.76 & 96.55 & 90.69 \\ \hline
Noreen et al \cite{noreen2021brain}   & Xception with KNN Ensemble & 94.34 & 94.97 & 95.40 & 84.30 \\ \hline
Proposed (Our)      & ShallowFed (FOLC-Net)  & 98.01 & 99.44 & 98.27 & 92.44 \\ \hline
\end{tabular}
\end{table}
\bmhead{Outcome}
ShallowFed (FOLC-Net) model consistently outperforms existing models in all views. It shows particularly significant improvements in the Axial and Sagittal views, suggesting that the model is not only robust but also more accurate across a range of different perspectives. The proposed model ability to provide such reliable and consistent performance across views positions it as a strong contender compared to previously proposed methods.
\subsection{Interpretable Analysis of FOLC-Net using GRADCAN and GRADCAN++ for Explainable Diagnosis}
A critical aspect of evaluating FOLC-Net model, especially in medical imaging, is not only their classification accuracy but also their interpretability. Understanding how a model arrives at its predictions is essential for building trust and ensuring clinical applicability. Model interpretability provides insights into the decision-making process by highlighting the features and regions that influence predictions. Techniques such as GRADCAM and GRADCAM++ offer valuable tools for visualizing the internal workings of convolutional networks by generating class-discriminative heatmaps. These visual explanations are instrumental in assessing whether the model is focusing on relevant anatomical regions in MRI data. In the case of the proposed ShallowFed model, GRADCAM and GRADCAM++ visualization (shown in Fig 8) clearly demonstrates its ability to accurately localize critical regions associated with tumor presence as compared to ShallowFed CNN. The resulting heatmaps indicate strong activation in areas that align with pathological findings, highlighting the model’s capability to extract meaningful spatial features. This interpretability not only confirms the effectiveness of the model shallow, federated structure but also enhances transparency, enabling clinicians and researchers to validate predictions and potentially identify regions that help further investigation.
\begin{figure}[h]
    \centering
    \includegraphics[width = 8cm]{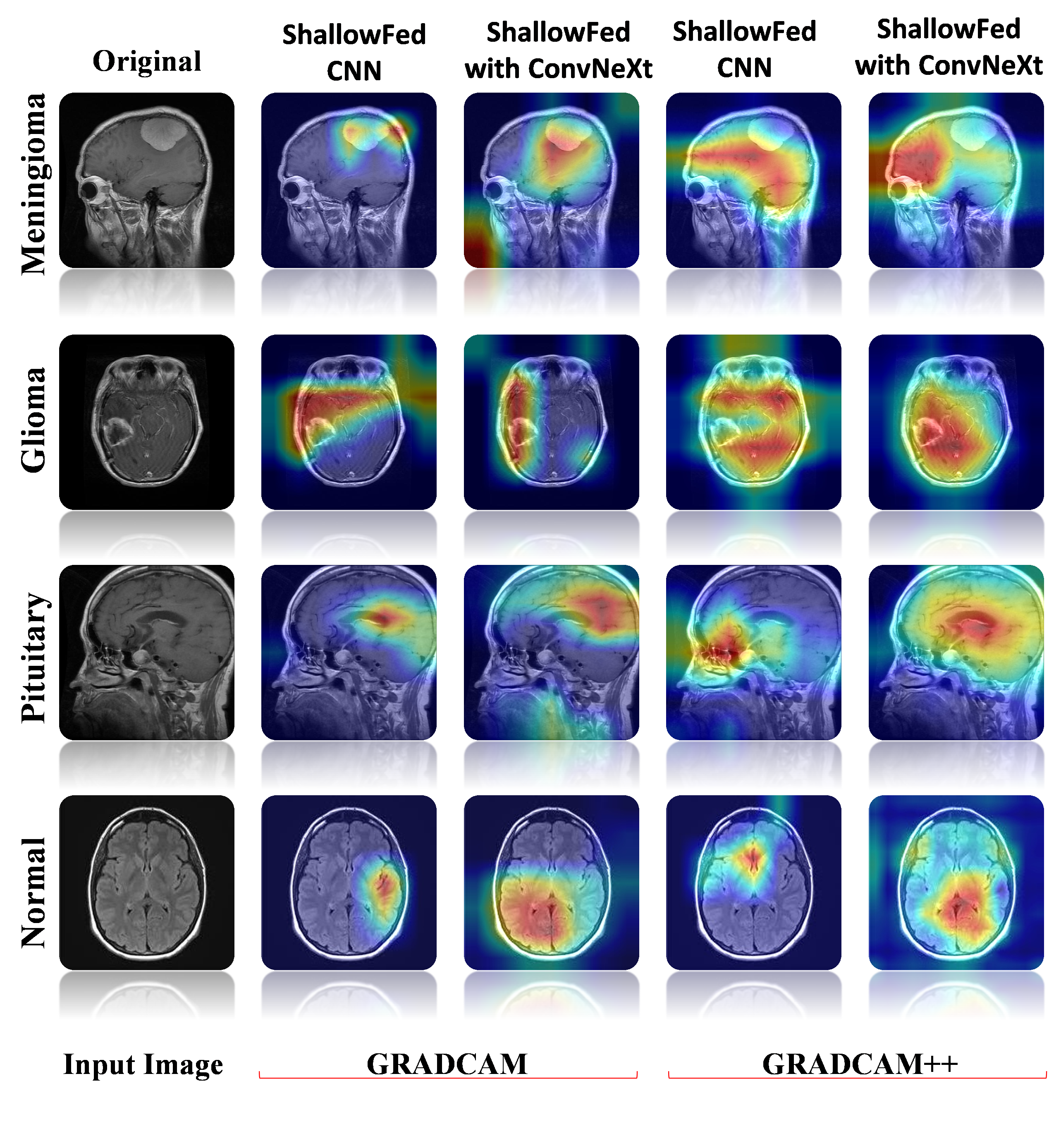}
    \caption{An overview of the Baseline CNN and the proposed ShallowFed with ConvNeXt (FOLC-Net), evaluated using GRADCAM and GRADCAM++}
    \label{fig:se.png}
\end{figure}
\subsection{t-distributed Stochastic Neighbor Embedding test of FOLC-Net: Multi-View to Single-View}
The t-SNE visualization in fig 9 subfigure (a) demonstrates the feature separability of FOLCO-Net when trained on All Views vs Axial, Coronal, and Sagittal. The visual clusters for the four tumor classes Glioma, Meningioma, Notumor, and Pituitary are well-separated and compact, indicating that combining multi-view and single-views data improves the model ability to learn discriminative and robust feature representations. The black, purple, orange, and yellow clusters distinctly represent the different classes, showing clear differentiation across the All-Views vs Axial, Coronal, and Sagittal approach.

When focusing on individual views, we see the following:

\bmhead{Axial View (subfigure b)}: The separation of the tumor classes is still clear. The Glioma and Meningioma clusters are well-separated, but the Notumor and Pituitary classes appear somewhat more merged, reflecting a slight loss in the clarity of the feature separability in this view alone.

\bmhead{Coronal View (subfigure c)}: The separation of classes is again observed, but the Notumor class overlaps with the Meningioma class more noticeably compared to the Axial View. However, the clusters for Glioma and Pituitary remain well-defined, highlighting the effective feature extraction in the Coronal view despite some challenges with class overlap.

\bmhead{Sagittal View (subfigure d)}: In this view, the Notumor class is less distinguishable from other classes, with a notable overlap between Meningioma and Notumor. The Pituitary class also exhibits some spread, but the Glioma class is clearly identifiable. This highlights that the Sagittal view has some limitations in feature discrimination, especially for certain tumor types.
\bmhead{Comparative Insight: }
Each view provides level of discriminative power as well as the All-Views configuration yields the robust feature separation, allowing for superior classification performance across All-Views vs Axial, Coronal, and Sagittal. The individual views, though useful, each exhibit some degree of class overlap and reduced separability compared to the multi-view approach.
\begin{figure}[!h]
\centering
\begin{minipage}[]{0.46\textwidth}
  \centering
  \includegraphics[width = \textwidth]{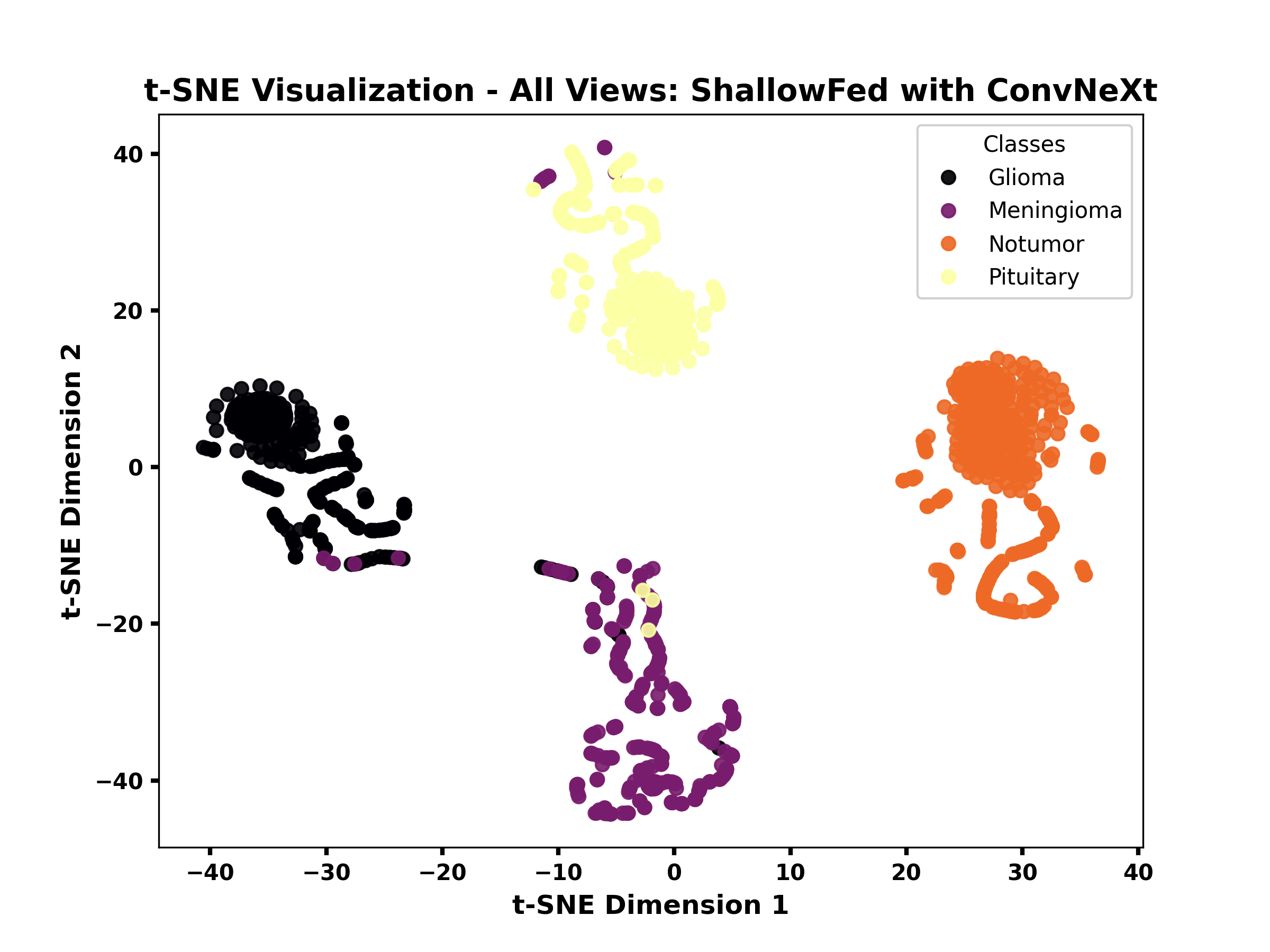}
  
    \begin{center}
     \textbf{a}    
     \end{center}
\end{minipage}
\begin{minipage}[]{0.46\textwidth}
  \centering
  \includegraphics[width = \textwidth]{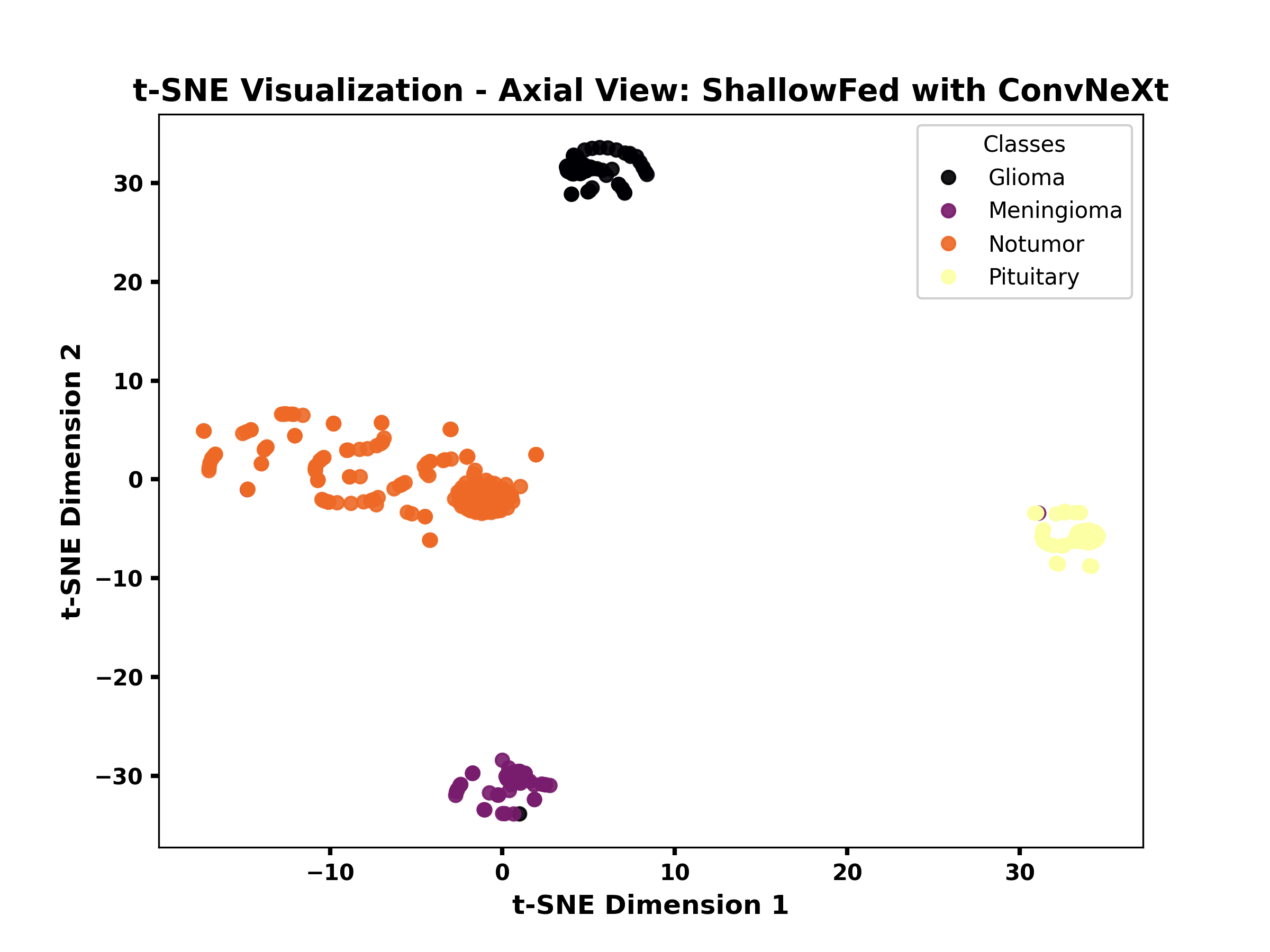}
 
     \begin{center}
     \textbf{b}    
     \end{center}
\end{minipage}
\begin{minipage}[]{0.46\textwidth}
  \centering
  \includegraphics[width = \textwidth]{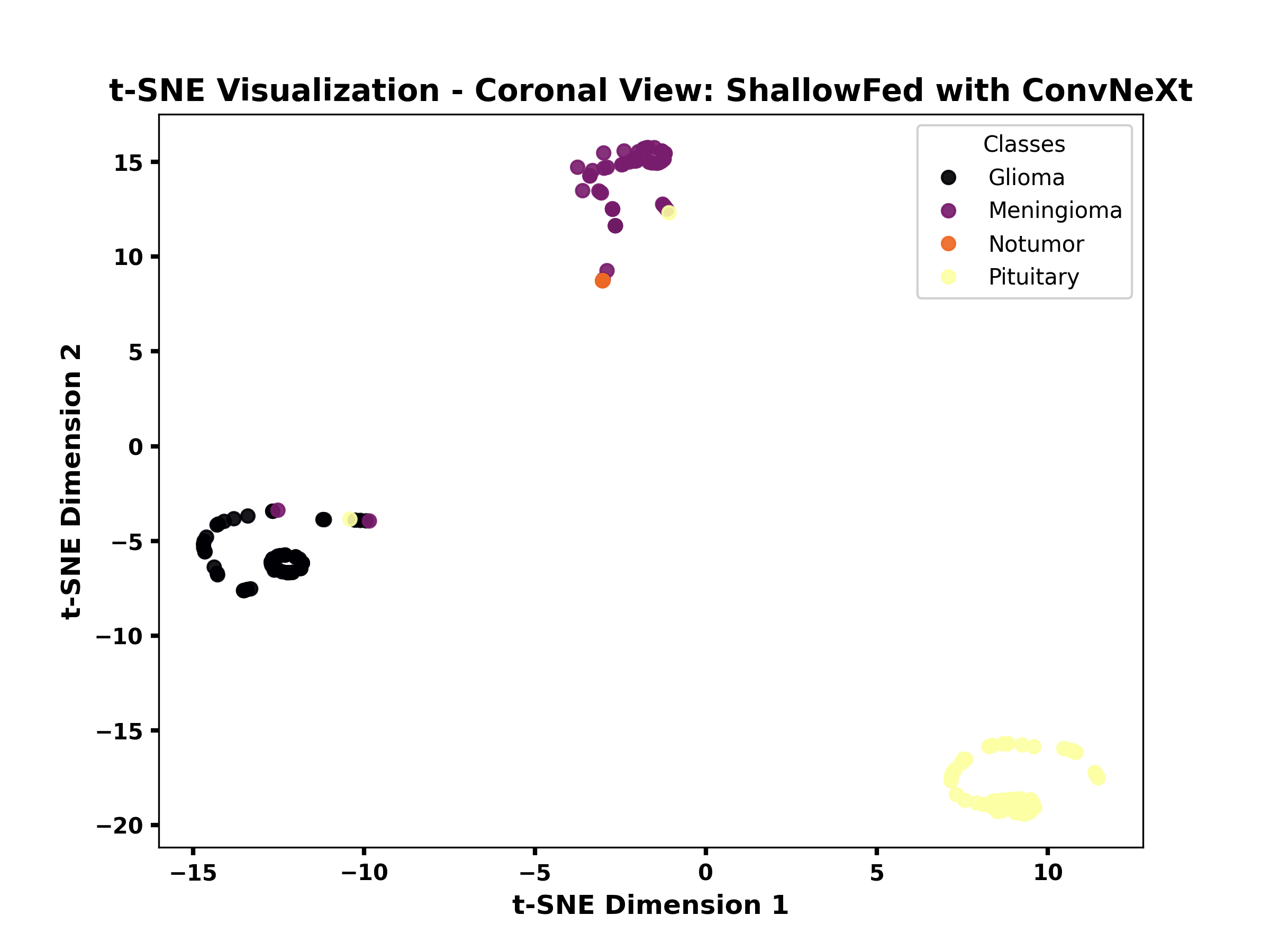}
  
    \begin{center}
     \textbf{c}    
     \end{center}
\end{minipage}
\begin{minipage}[]{0.46\textwidth}
  \centering
  \includegraphics[width = \textwidth]{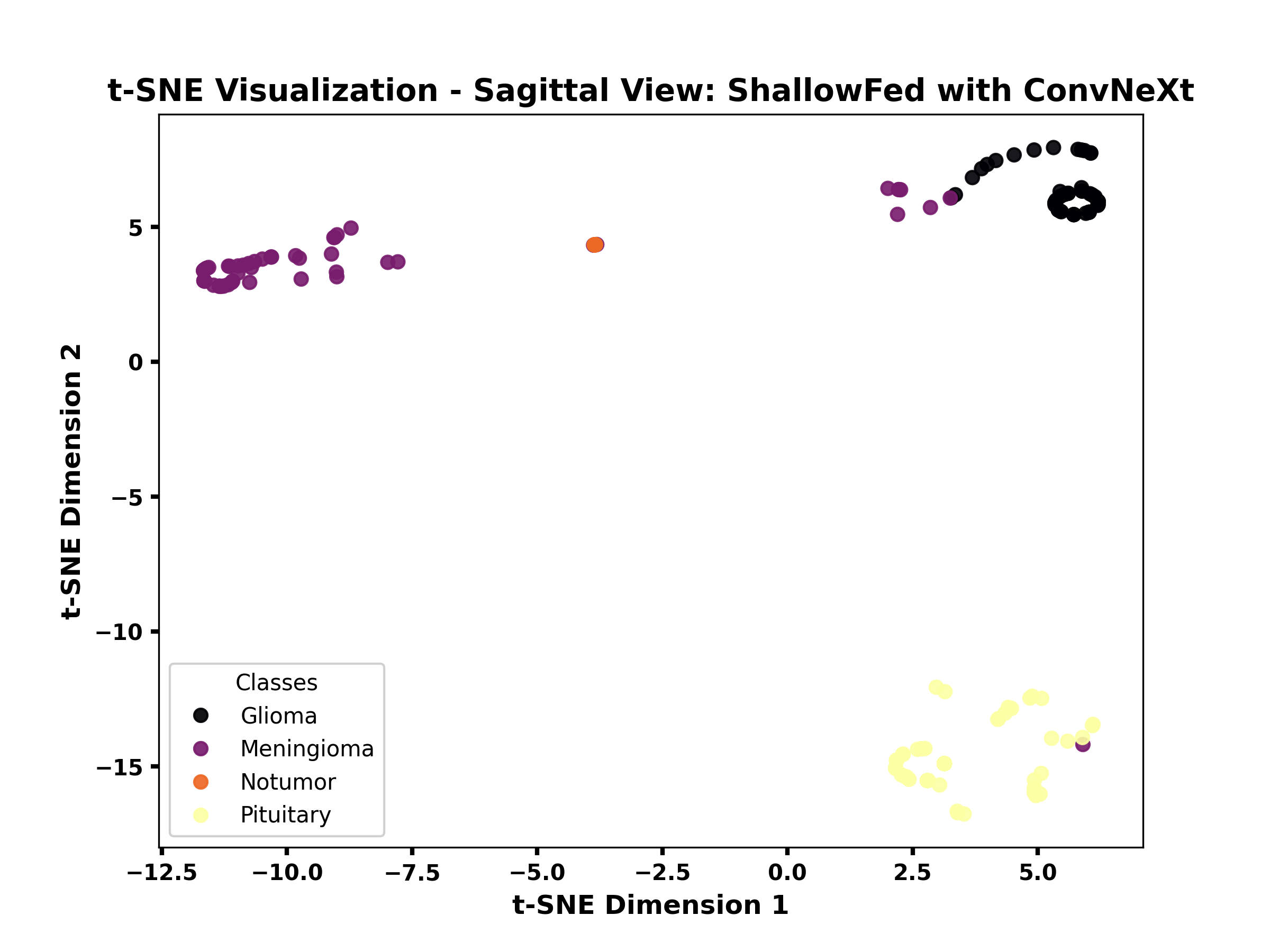}
 
     \begin{center}
     \textbf{d}    
     \end{center}
\end{minipage}
  \caption{An overview of the proposed ShallowFed with ConvNeXt (FOLC-Net), evaluated using t-SNE visualization: (a) All Views, (b) Axial View, (c) Coronal View, and (d) Sagittal View }
  \label{fig:11.png}
\end{figure}
\subsection{Statistical Analysis with Chi-Square Tests of FOLC-Net: Multi-View to Single-View for Robust Model Evaluation and Validation}
The Chi-square test is essential for validating proposed model used in disease diagnosis as it statistically assesses whether the observed categorization of disease predictions significantly deviates from what would be expected by chance. This ensures that the model is reliably distinguishing between different disease classes and not simply generating random results, thus validating the model practical applicability for accurate and consistent diagnoses in clinical settings. 

If the observed categorization results in the confusion matrix differ significantly from what would be expected by chance, the association between predicted and actual class labels can be statistically validated using the Chi-square test with Degrees of freedom (DOF = 9). In our study, this test produced a highly significant result with a p-value of less than 0.001 (Table 8), indicating that the ShallowFed with ConvNeXt (FOLC-Net) predictions are strongly correlated with the true class labels across all representation, rather than being random. This outcome not only reinforces the validity of the model but also demonstrates that the model has captured meaningful patterns that go beyond mere chance. The Chi-square test (Eq.~22) serves as an additional, quantitative measure of the model effectiveness, showing that the observed categorization frequencies align well with the actual distributions. This finding suggests that the model can be generalized to real-world diagnostic scenarios, providing a robust tool for identifying brain tumors with high reliability. The results from the Chi-square test thus lend substantial support to the model applicability in clinical settings, where accuracy and consistency are paramount for effective diagnosis.

\[
\chi^2 = \sum \frac{(X - Y)^2}{Y} \tag{22}
\]
\begin{table}[h!]
\centering
\caption{Chi-Square Evaluation of FOLC-Net for Multi-View to Single-Views anatomical views}
\begin{tabular}{ccc}
\hline
\textbf{Anatomical views} & \textbf{$\chi^2$} & \textbf{DOF} \\
\hline
All-Views        & 3731.3 & 9 \\
Axial View       & 3800.2 & 9 \\
Coronal View     & 3500.4 & 9 \\
Sagittal         & 3500.4 & 9 \\
\hline
\end{tabular}
\end{table}
\bmhead{Outcome: }In all these cases, the p-value would be extremely small, indicating that the observed categorization significantly deviates from what would be expected by chance. This reinforces the reliability of the model predictions, confirming that it is effectively capturing meaningful patterns in the data rather than producing random results. Such a strong association between predicted and actual outcomes demonstrates the model's robustness and its potential for accurate, real-world application in medical diagnosis.
\subsection{Complexity analysis}
This section illustrates the complexity analysis (Fig 10) between the Baseline CNN and the Proposed ShallowFed with ConvNeXt (FOLC-Net) models, providing insights into the computational efficiency of the models. The analysis highlights the reduction in computational complexity, where the Baseline CNN requires approximately 2,710,000 million parameters along 3.1 MB storage memory, while the Proposed ShallowFed with ConvNeXt reduces this to about 1,217,000 million parameters with 0.9 MB storage space. This reduction, shown by the green line, indicates that the proposed model is significantly more efficient. Performing complexity analysis is essential to assess the computational cost of FOLC-Net model, helping identify trade-offs between model accuracy and resource consumption, which is particularly critical for deployment in resource-constrained environments like edge computing.
\begin{figure}[h!]
    \centering
    \includegraphics[width = 10cm]{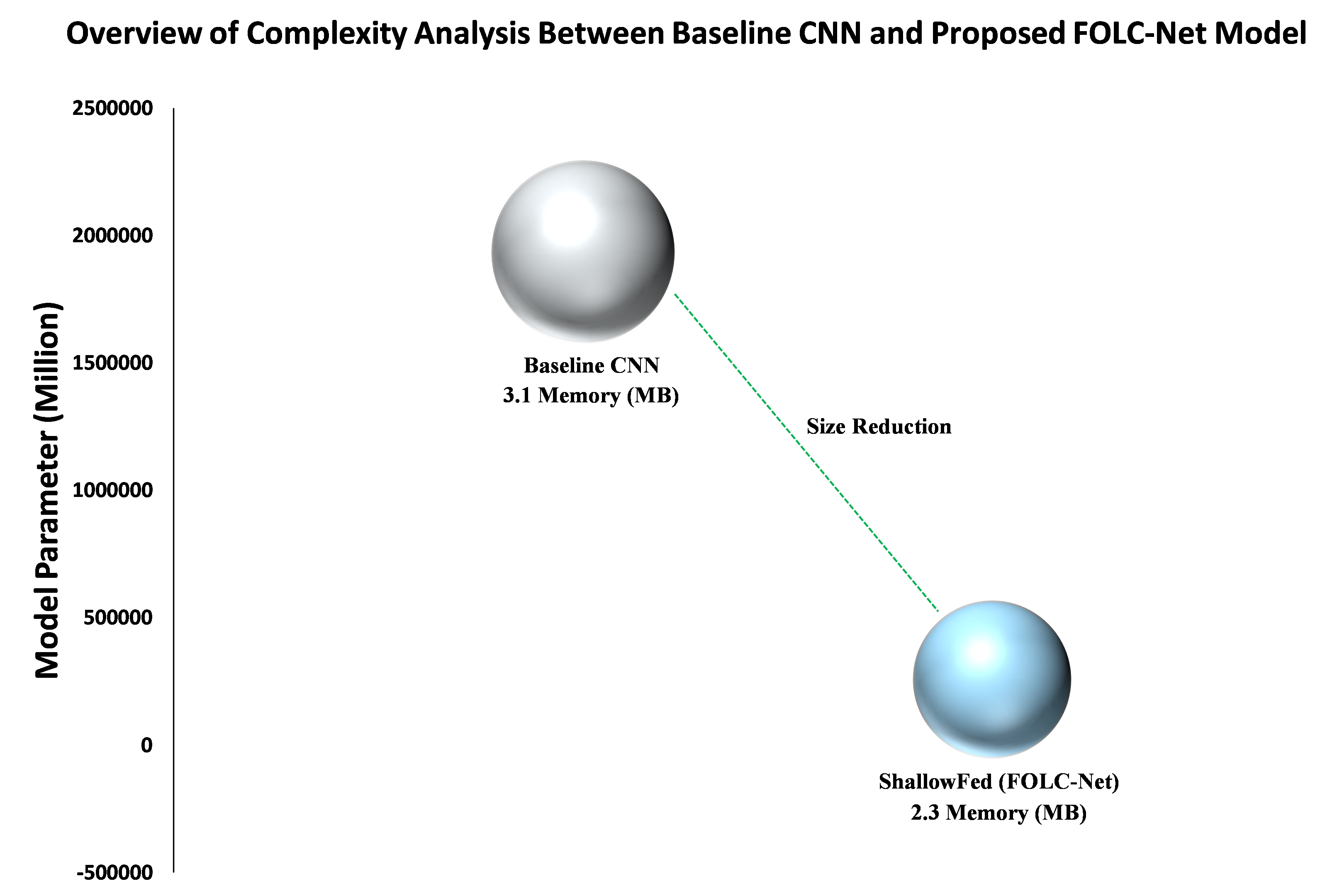}
    \caption{An overview of the complexity analysis for Baseline CNN, and proposed ShallowFed with ConvNeXt (FOLC-Net) }
    \label{fig:se.png}
\end{figure}
\subsection{Additional test}
To evaluate the generalization ability and performance of the proposed FOLC-Net framework on a variety of unseen data, this section discusses the use of an additional dataset for validation, which helps mitigate overfitting. The Brain MRI dataset available on Kaggle \footnote{https://www.kaggle.com/code/ahmedhamada0/brain-tumor-detection-br35h/data}, namely BR35H binary class \cite{hekmat2025brain}, was used in this study. Specifically, the Brain Tumor Detection 2020 dataset was utilized and introduced it. This dataset contains 1500 images of brain tumors and 1500 images of normal MRI, providing a comprehensive representation of brain anomalies. Table 9 presents further details on the distribution of samples between the tumor and normal classes. For our experiment, to maintain the fair evaluation dataset was split into initial predefined training, validation and testing sets to ensure an accurate evaluation of image diagnosis. The training set includes 1200 images from the tumor class and 1200 images from the normal class, while the testing set consists of 300 images from the tumor class and 300 images from the normal class. This balanced split ensures a fair representation of both benign and malignant cases during the training and testing phases. MRI images of both normal and tumor brains are shown in Fig 11, offering a visual overview of the dataset composition.

We applied the same preprocessing and augmentation settings as described in our primary kaggle multiclass dataset utilization to ensure consistency in data handling. These techniques, which include normalization, and resizing were crucial for enhancing the diversity of the training samples. By maintaining the same preprocessing pipeline, we aimed to guarantee that the ShallowFed model performance is evaluated under consistent conditions across all datasets.
\begin{figure}[h!]
    \centering
    \includegraphics[width = 8cm]{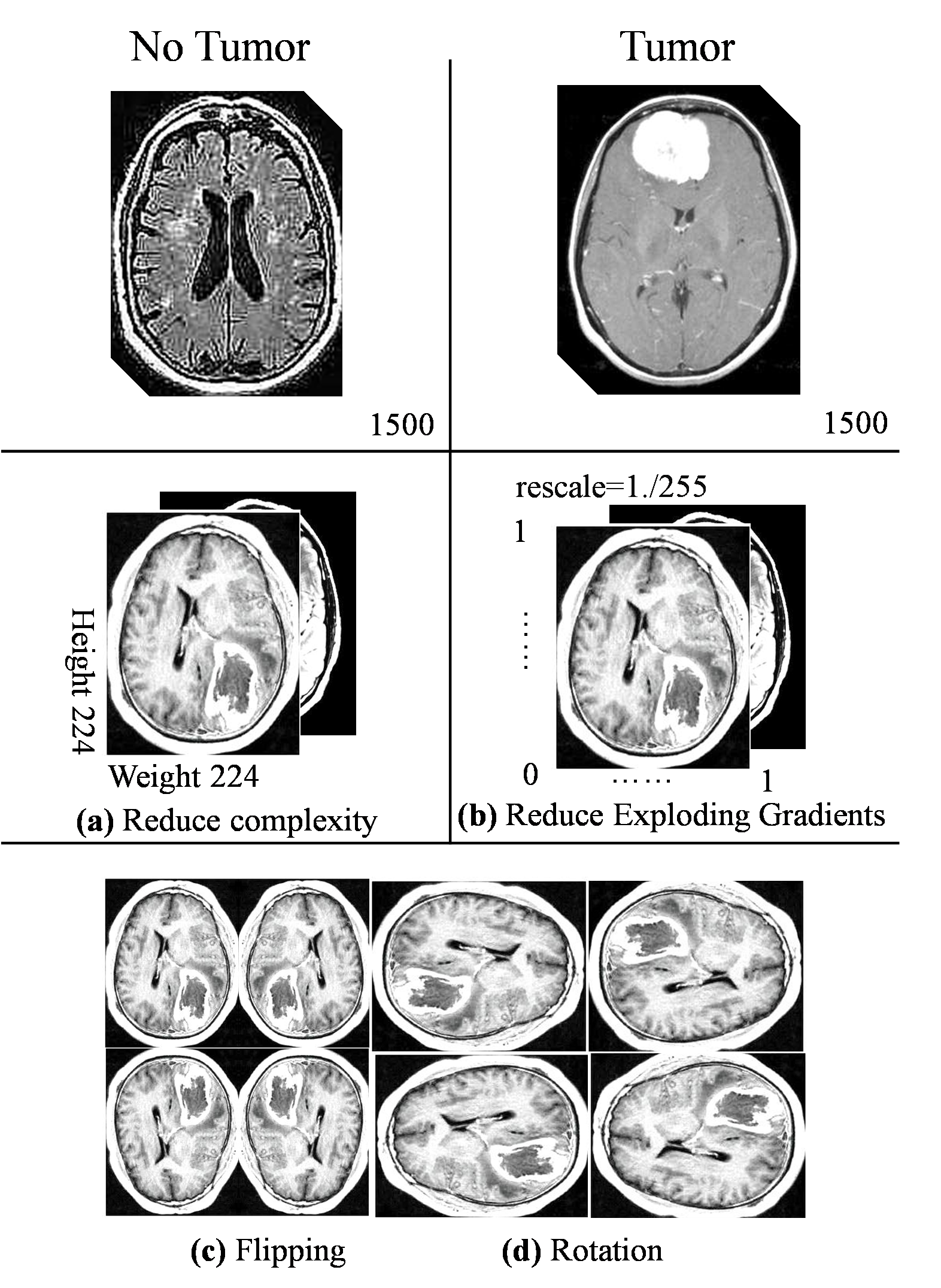}
    \caption{Sample images of each cancer type along Axial view (Additional dataset (BR35H)) overview}
    \label{fig:se.png}
\end{figure}
\begin{table}[ht]
\centering
\caption{Presents the distribution of class samples across four distinct image types available in the Kaggle BR35H dataset. These image types correspond to single anatomical views or angles from which the brain images were captured. Specifically, the axial view represents images taken from a top-down perspective of the brain only present here}
\begin{tabular}{ccccc}
\hline
\textbf{Class} & \textbf{Train} & \textbf{Validate} & \textbf{Testing} & \textbf{Total} \\ 
\hline
No Tumor & 1080 & 120 & 300 & 1500 \\
Tumor & 1080 & 120 & 300 & 1500 \\
\hline
\end{tabular}

\label{table:dataset_overview}
\end{table}

Table 10 compares the performance of the proposed ShallowFed with ConvNeXt (FOLC-Net) model and its baseline, ShallowFed, in terms of precision, recall, F1-score, and accuracy for brain tumor classification in the axial view (Single). Both models demonstrate high performance, with ShallowFed achieving an accuracy of 98.00\%, while the ShallowFed with ConvNeXt (FOLC-Net) shows a slight improvement, achieving an accuracy of 98.16\%. The precision, recall, and F1-scores for both models are also close, with the FOLC-Net slightly outperforming ShallowFed in the tumor class, where it achieves a precision of 0.99 compared to 0.98 for ShallowFed. This indicates that the incorporation of ConvNeXt into ShallowFed (FOLC-Net) marginally enhances the model performance in terms of tumor classification while maintaining similar overall effectiveness.

\begin{table}[h!]
\centering
\caption{Performance comparison of Baseline CNN, and proposed FOLC-Net model across axial view: Additional dataset BR35H}
\begin{tabular}{cccccc}
\hline
\textbf{Model} & \textbf{Class} & \textbf{Precision} & \textbf{Recall} & \textbf{F1-Score} & \textbf{Accuracy} \\ \hline
\multirow{2}{*}{ShallowFed} & No Tumor & 0.97 & 0.99 & 0.98 & 98.00 \\ 
  & Tumor & 0.98 & 0.97 & 0.98 & \\ \hline
\multirow{2}{*}{ShallowFed with ConvNeXt (FOLC-Net)}   & No Tumor & 0.97 & 0.99 & 0.98 & 98.16 \\ 
  & Tumor & 0.99 & 0.97 & 0.98 & \\ \hline
\end{tabular}
\end{table}

The confusion matrix comparisons between the proposed model (ShallowFed with ConvNeXt) and its baseline model (ShallowFed) show (Fig 12) a notable improvement in performance. In the baseline model (a), there are 297 \text{True Negatives} and 291 \text{True Positives}, with 9 \text{False Positives} and 3 \text{False Negatives}. In contrast, the proposed model (b) demonstrates an improvement with 298 \text{True Negatives}, 291 \text{True Positives}, 9 \text{False Positives}, and only 2 \text{False Negatives}. This indicates a slight reduction in \text{False Negatives}, suggesting the proposed model has enhanced accuracy, particularly in identifying the No Tumor class. The overall correct classification rate is higher in the proposed model compared to the baseline, reflecting a more reliable performance in tumor detection.
\begin{figure}[h!]
\centering
\begin{minipage}[]{0.40\textwidth}
  \centering
  \includegraphics[width = \textwidth]{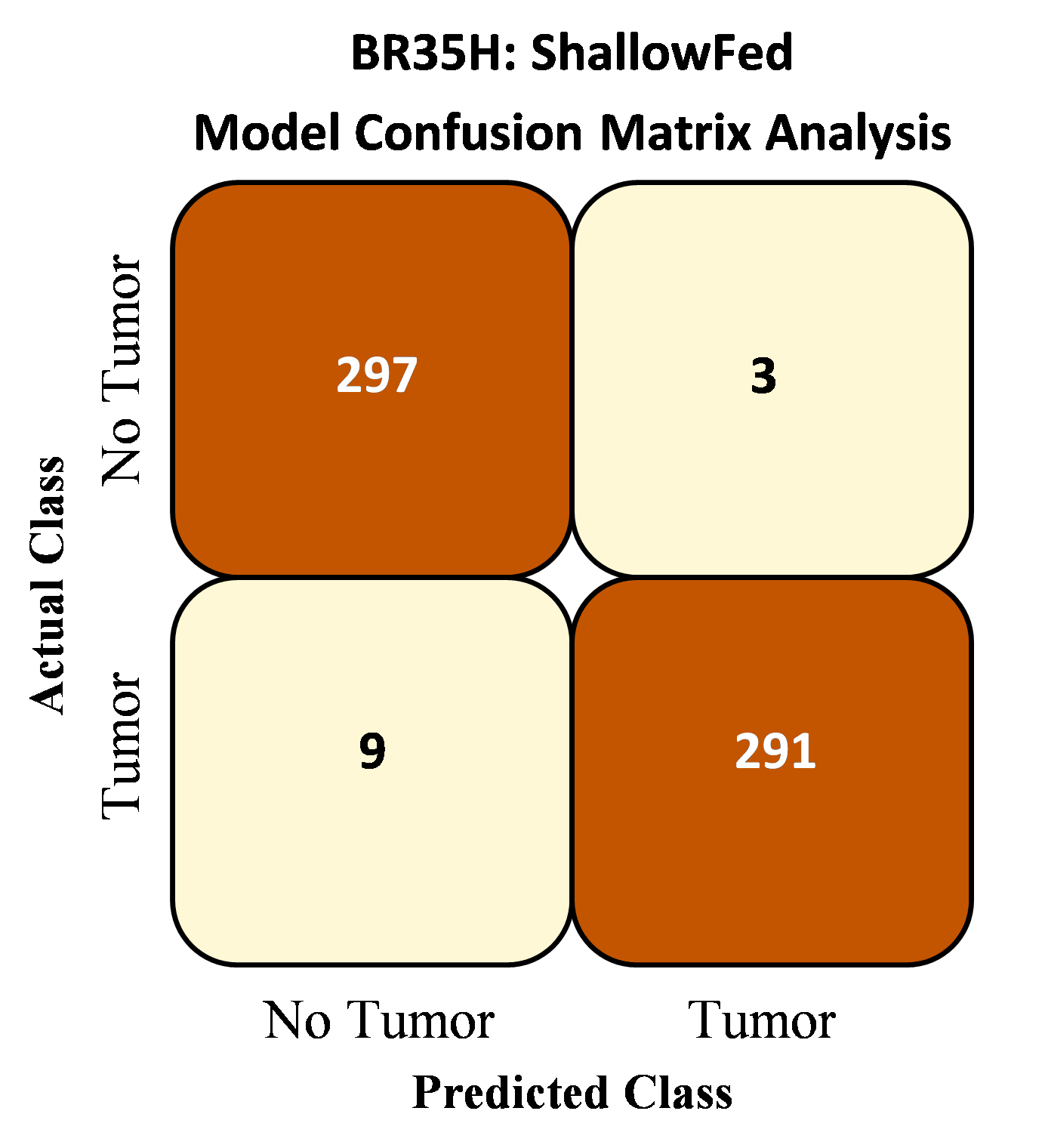}
  
    \begin{center}
     \textbf{a}    
     \end{center}
\end{minipage}
\begin{minipage}[]{0.40\textwidth}
  \centering
  \includegraphics[width = \textwidth]{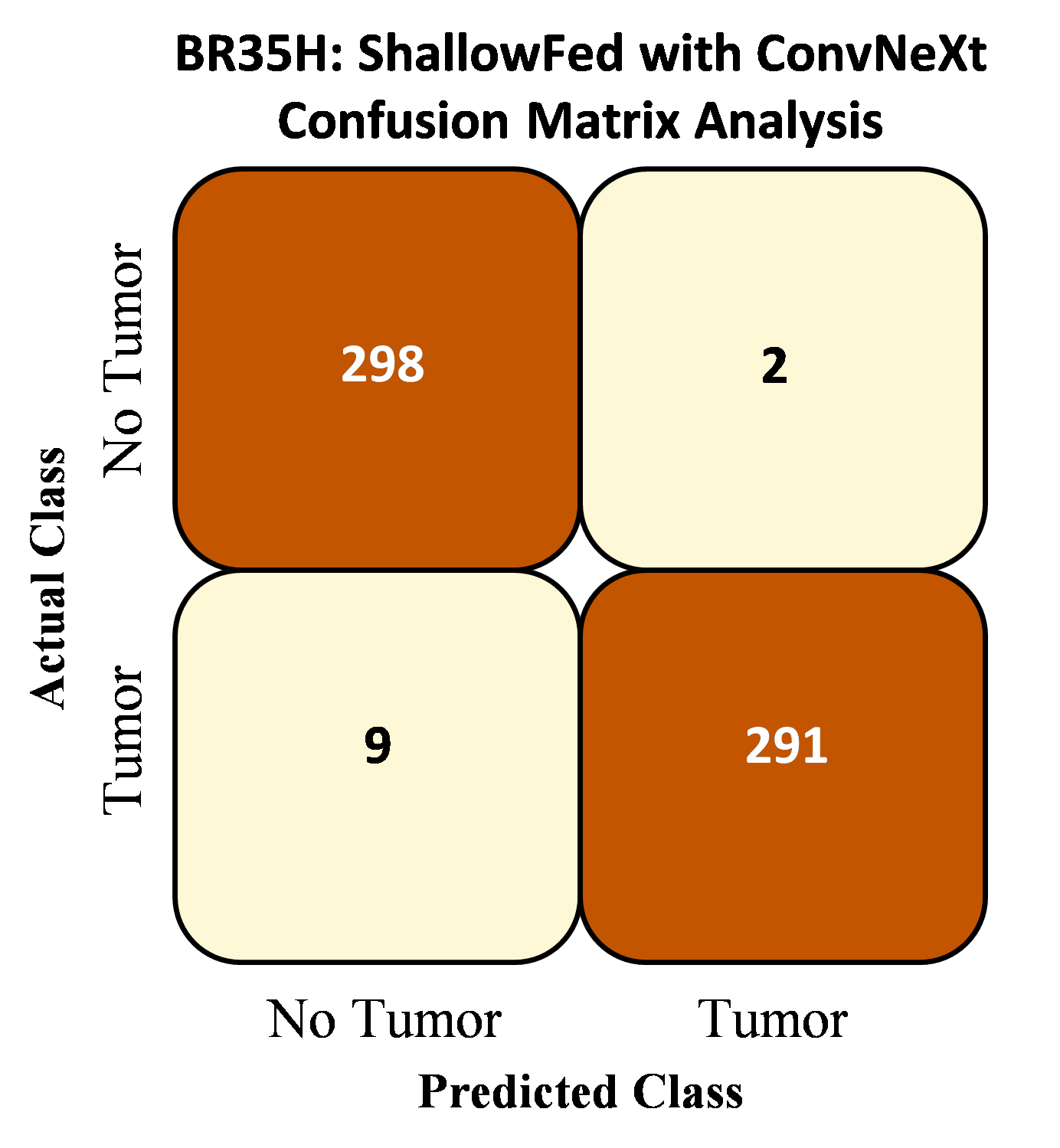}
 
     \begin{center}
     \textbf{b}    
     \end{center}
\end{minipage}
  \caption{An overview of the confusion matrices for Baseline CNN, and proposed ShallowFed with ConvNeXt (FOLC-Net): (a) Baseline CNN, and (b) ShallowFed with ConvNeXt (FOLC-Net) }
  \label{fig:11.png}
\end{figure}
In the fig 13, the ROC curve and PR curve illustrate the performance of the proposed model, ShallowFed with ConvNeXt, compared to the baseline model, ShallowFed CNN. The ROC curve shows that the proposed model significantly outperforms the baseline, achieving a higher AUC of 0.9956 compared to the baseline model AUC of 0.9880, indicating better overall classification performance. Similarly, the Precision-Recall curve shows a higher AP of 0.9972 for the proposed model, compared to the baseline model's AP of 0.9903, further reinforcing the superior performance of the proposed model in terms of both precision and recall.
\begin{figure}[!h]
\centering
\begin{minipage}[]{0.46\textwidth}
  \centering
  \includegraphics[width = \textwidth]{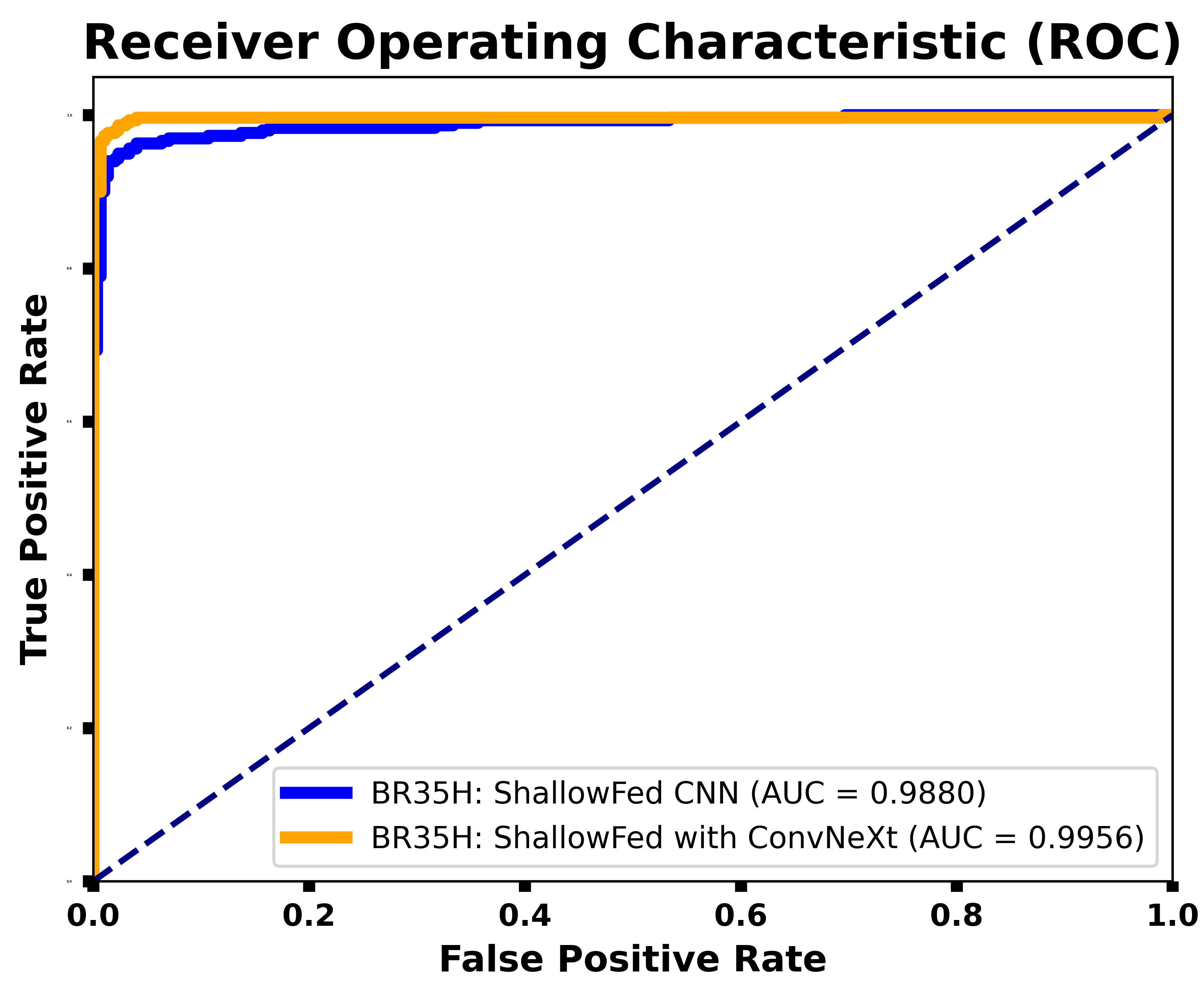}
  
    \begin{center}
     \textbf{a}    
     \end{center}
\end{minipage}
\begin{minipage}[]{0.46\textwidth}
  \centering
  \includegraphics[width = \textwidth]{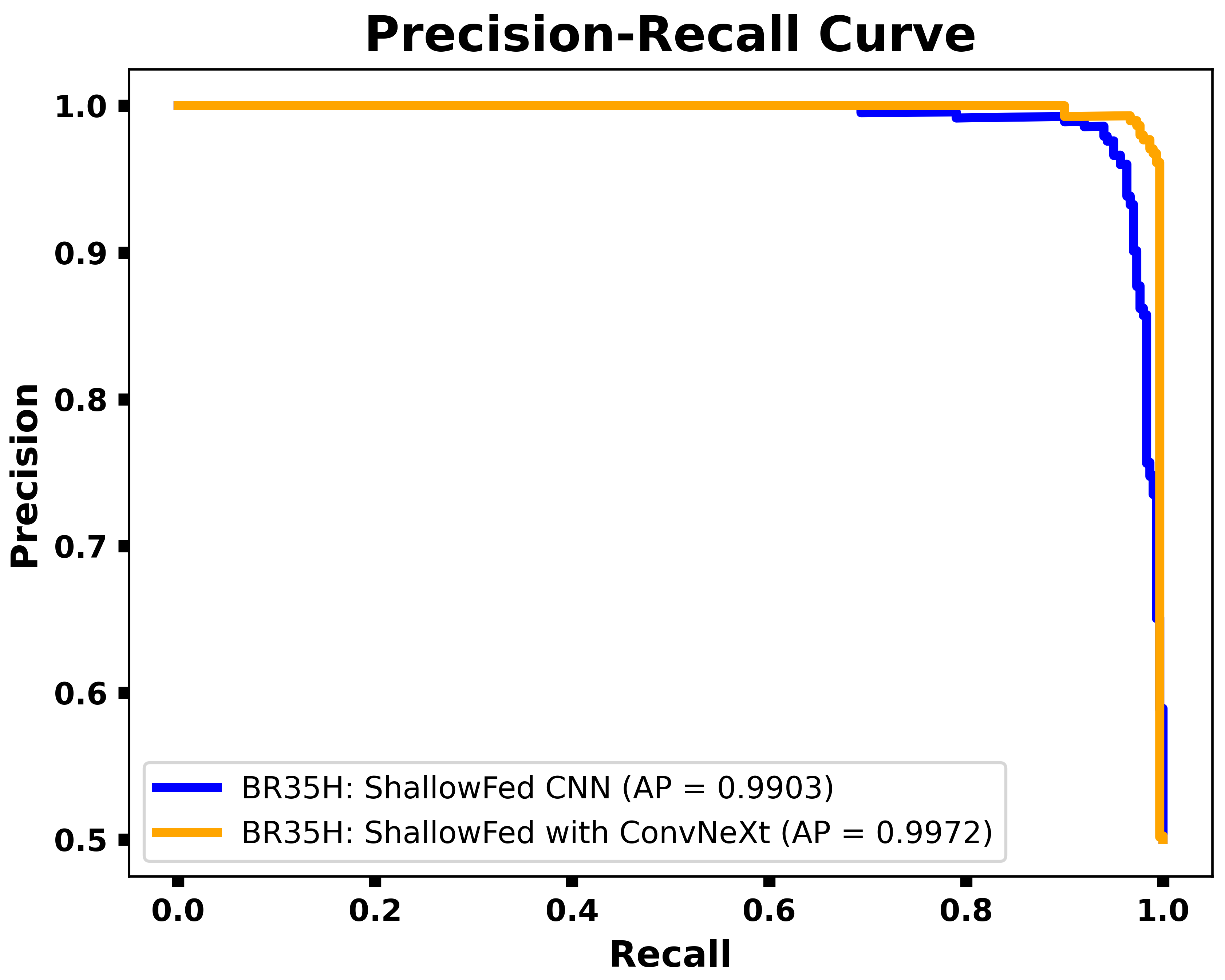}
 
     \begin{center}
     \textbf{b}    
     \end{center}
\end{minipage}
  \caption{An overview of the Baseline CNN and the proposed ShallowFed with ConvNeXt (FOLC-Net), evaluated using ROC and PR curves: (a) ROC, and (b) PR}
  \label{fig:11.png}
\end{figure}
The performance comparison between the proposed ShallowFed (FOLC-Net) model and existing CNN models reveals (Table 11) that ShallowFed (FOLC-Net) outperforms the other models across all evaluation metrics on additional dataset. With an accuracy of 98.16\%, precision of 98.19\%, recall of 98.16\%, and an F1-score of 98.16\%, ShallowFed (FOLC-Net) demonstrates superior performance compared to the other models. Among the existing CNN models, the highest performance was achieved by Base-DensNet121 and Base-MobileNetV1, both with an accuracy of 97.66\%, but they fall short in comparison to ShallowFed (FOLC-Net) metrics. Other models like ResNet50, ResNet101, and VGG16 perform significantly lower, with accuracy values ranging from 88.16\% to 97.50\%. Therefore, ShallowFed (FOLC-Net) demonstrates a clear robustness in terms of both precision and recall, showing its potential as a more effective model. In particular, ShallowFed (FOLC-Net) consistent performance across all metrics highlights its robustness and reliability. Furthermore, its ability to outperform other models, including those like ResNet and NASNet that are widely used in various applications \cite{khan2024hybrid, hekmat2025brain}, demonstrates its suitability for more complex tasks. The results indicate that ShallowFed (FOLC-Net) design effectively enhances classification performance.
\begin{table}[h!]
\centering
\caption{Performance comparison of ShallowFed (FOLC-Net) and existing CNN models: BR35H Binary}
\begin{tabular}{ccccc}
\hline
\textbf{Model} & \textbf{Accuracy} & \textbf{Precision} & \textbf{Recall} & \textbf{F1-Score} \\ \hline
Base- DensNet121 & 97.66 & 97.90 & 97.66 & 97.66 \\ 
Base- DensNet169 & 97.50 & 97.56 & 97.50 & 97.49 \\ 
Base- DensNet201 & 97.16 & 97.01 & 97.00 & 96.95 \\ 
Base- MobileNetV1 & 97.66 & 97.74 & 97.66 & 97.66 \\
Base- MobileNetV2 & 96.16 & 96.43 & 96.16 & 96.16 \\ 
Base- ResNet50 & 88.16 & 88.75 & 88.16 & 88.12 \\ 
Base- ResNet101 & 90.83 & 91.39 & 90.83 & 90.80 \\ 
Base- ResNet152 & 90.66 & 90.92 & 90.66 & 90.65 \\ 
Base-VGG16 & 97.16 & 97.23 & 97.16 & 97.16 \\ 
Base-VGG19 & 97.50 & 97.54 & 97.50 & 97.49 \\ 
Base- NASNetMobile & 97.00 & 97.03 & 97.00 & 96.99 \\ 
Base- NASNetLarge & 97.16 & 97.28 & 97.16 & 97.16 \\ 
ShallowFed (FOLC-Net) & 98.16 & 98.19 & 98.16 & 98.16 \\ \hline
\end{tabular}
\label{table:performance_comparison}
\end{table}
Fig 14 demonstrates the use of GRADCAM and GRADCAM++ for visualizing the attention areas of a model in brain MRI scans. The baseline model heatmaps highlight regions of the brain that are less defined and focused on areas that are not central to the region of interest. In contrast, the ShallowFed (FOLC-Net) model exhibits more accurate and focused heatmaps, highlighting the relevant brain regions with efficient precision, especially the abnormal areas indicative of brain lesions. This comparison suggests that the proposed model has superior performance in identifying critical features compared to the baseline model, as the heatmaps in the second row are more aligned with the pathological regions in the MRI scans.
\begin{figure}[h]
    \centering
    \includegraphics[width = 8cm]{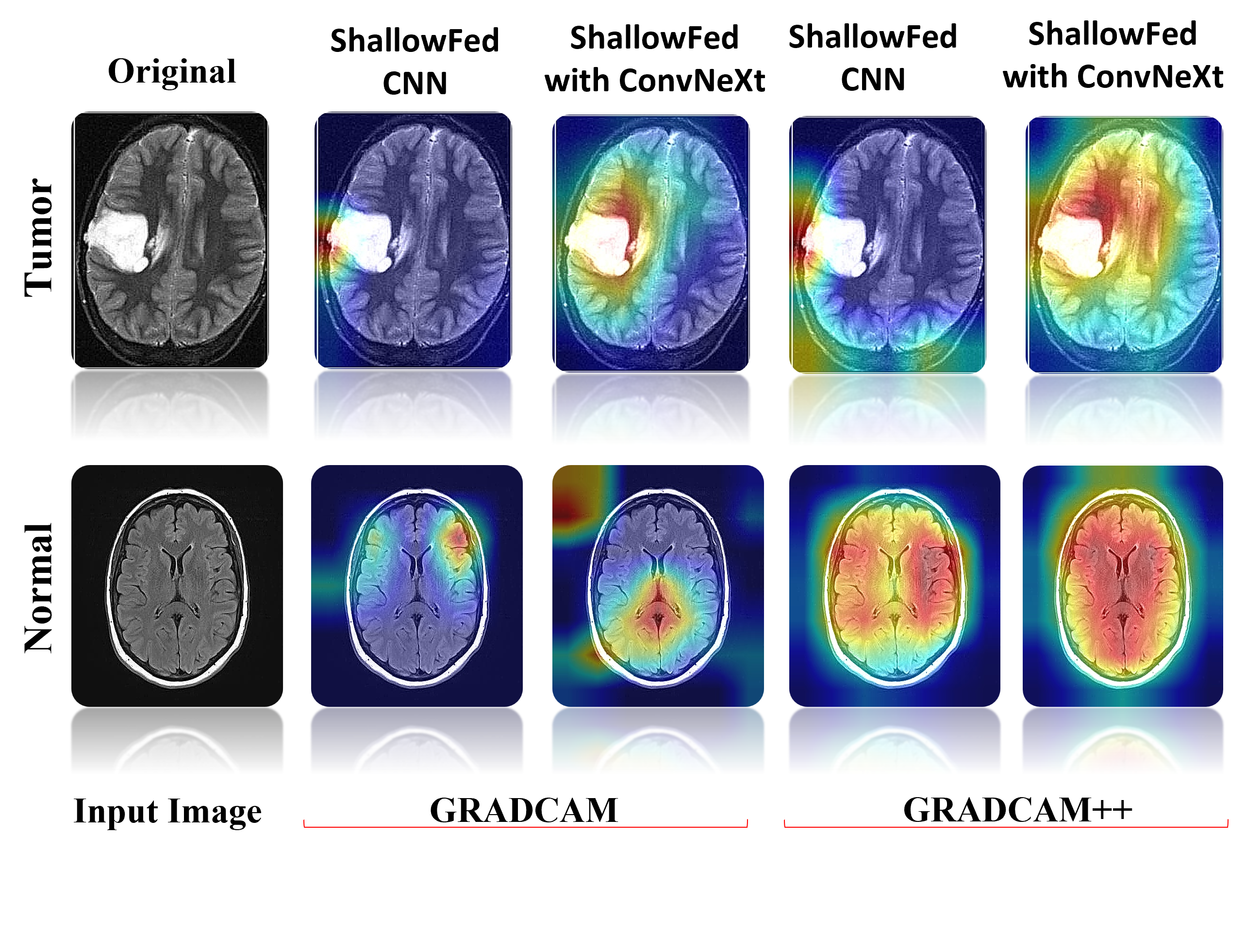}
    \caption{An overview of the Baseline CNN and the proposed ShallowFed with ConvNeXt (FOLC-Net), evaluated using GRADCAM and GRADCAM++: Additional BR35H dataset}
    \label{fig:se.png}
\end{figure}
In the fig 15 t-SNE visualization, the proposed model (ShallowFed with ConvNeXt) significantly improves the clustering and separation of classes compared to the baseline model (ShallowFed CNN). In the first plot (a), the baseline model shows a clear overlap between the No Tumor and Tumor classes, indicating a less distinct feature representation. In contrast, the second plot (b) demonstrates a better-defined separation between the two classes using the proposed ShallowFed (FOLC-Net) model, with more distinct clusters for both No Tumor and Tumor. This indicates that the ConvNeXt-enhanced ShallowFed model achieves a more effective and refined feature extraction, resulting in clearer differentiation between the two classes in the t-SNE space.
\begin{figure}[!h]
\centering
\begin{minipage}[]{0.46\textwidth}
  \centering
  \includegraphics[width = \textwidth]{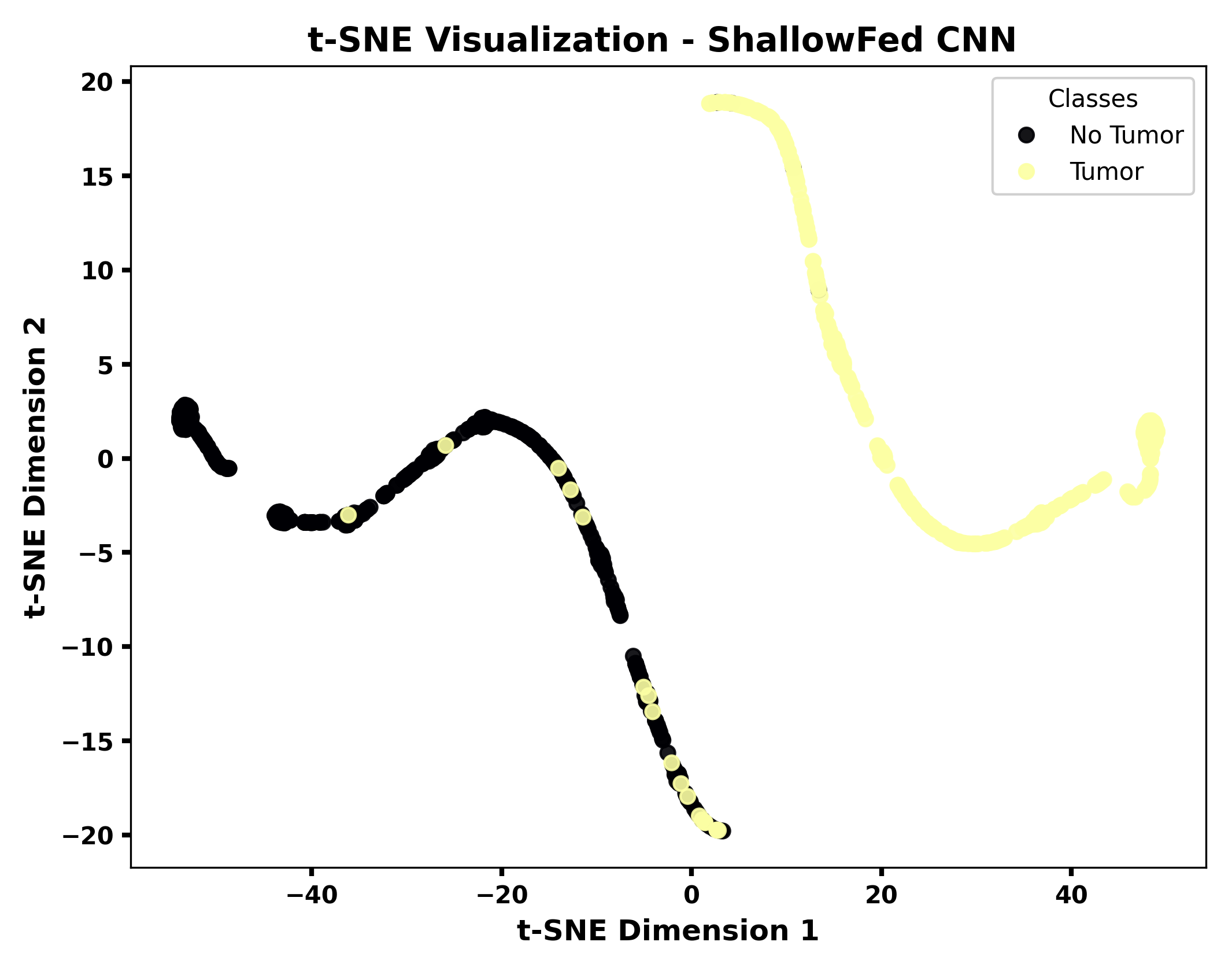}
  
    \begin{center}
     \textbf{a}    
     \end{center}
\end{minipage}
\begin{minipage}[]{0.46\textwidth}
  \centering
  \includegraphics[width = \textwidth]{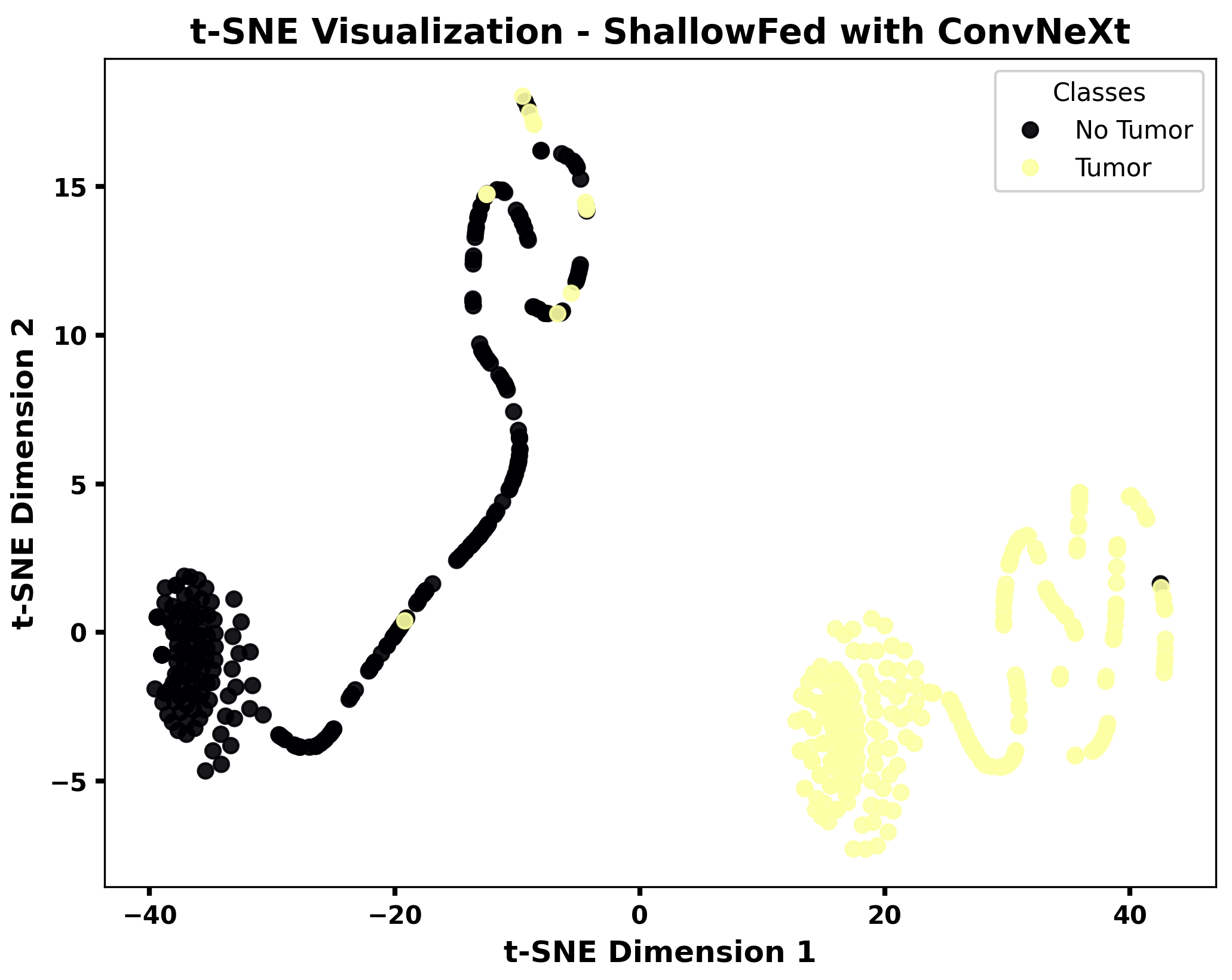}
 
     \begin{center}
     \textbf{b}    
     \end{center}
\end{minipage}
  \caption{An overview of the Baseline CNN (a), and proposed ShallowFed with ConvNeXt (FOLC-Net) (b) evaluated using t-SNE visualization: Additional BR35H dataset }
  \label{fig:11.png}
\end{figure}
\subsection{Comprehensive Disease Modality Testing}
Why is it important to test a ShallowFed (FOLC-Net) model on data that is entirely different from the primary training dataset? The reason lies in evaluating the model ability to generalize beyond the data it has seen during training. If a model is only tested on data similar as we did in additional test to its training set, it may simply memorize patterns, leading to overfitting. Testing on unseen, totally different data ensures that the model is not only performing well on the specific examples it has encountered but is also capable of making accurate predictions in real-world scenarios, where the data can vary significantly. This approach helps validate the ShallowFed (FOLC-Net) robustness, reliability, and true predictive power, ensuring it can adapt to new, previously unknown situations. 
In this study, we have utilized two publicly available distinct and previously unseen datasets, namely the Histopathological \footnote{(https://www.kaggle.com/datasets/andrewmvd/lung-and-colon-cancer-histopathological-images )} \cite{khan2025optimize} and MHC-CT \footnote{https://github.com/VS-EYE/KidneyDiseaseDetection.git } \cite{khan2025multilevel} datasets , to rigorously validate the robustness and generalizability of the proposed model (Fig 16). The MHC-CT dataset consists of 2,886 total samples, utilized 286 for testing. While this dataset is designed to evaluate the model performance in classifying binary CT-scan images, ensuring it can generalize well across unseen data. The Histopathological dataset consists of two distinct subsets: Colon-ACA (test set of 400 samples) and Colon-N (test set of 400 samples), totaling 800 samples. This dataset is primarily used for classifying colon cancer, with the Colon-ACA subset representing cancerous tissue and the Colon-N subset representing non-cancerous tissue. 

By testing the model on these diverse datasets, we aim to ensure that its performance is not limited to a specific type of data but can effectively handle a wide range of scenarios. This evaluation helps to further demonstrate the model ability to adapt to variations in data and confirms its reliability for real-world applications where data distribution may differ from the training set. The use of these datasets is crucial in assessing the model overall effectiveness and robustness. Figure presents the samples images visualization from both datasets.
\begin{figure}[h]
    \centering
    \includegraphics[width = 10cm]{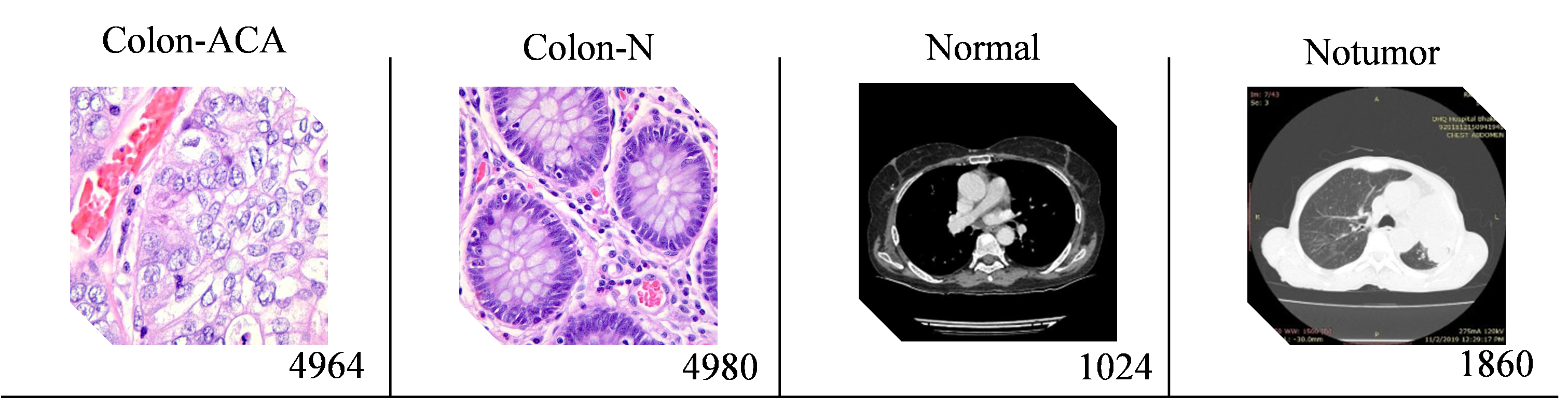}
    \caption{Domain-Shifted Dataset Samples: A Comprehensive Overview of Histogram and MHC-CT Images}
    \label{fig:se.png}
\end{figure}
Table 12 presents a performance comparison of the ShallowFed (FOLC-Net) model and a baseline CNN model on the Domain Shift Test using two different datasets: Histopathological and MHC-CT. For the Histopathological dataset, the ShallowFed CNN model shows strong performance, with precision, recall, F1-score, and accuracy all near 99\% for both Colon-ACA and Colon-N categories. The ShallowFed (FOLC-Net) model performs even better, achieving efficient precision and recall for the Colon-N class and slightly higher accuracy (99.75\%) for Colon-ACA. In the MHC-CT dataset, the ShallowFed CNN also performs well, with an accuracy of 99.30\% for the Tumor class and 99.46\% for the Notumor class. However, the ShallowFed (FOLC-Net) model outperforms the CNN model by achieving 100\% precision and recall for the Tumor class and slightly higher accuracy (99.65\%). Overall, the ShallowFed (FOLC-Net) model demonstrates superior performance, particularly in terms of precision and recall across both datasets.
\begin{table}[ht]
\centering
\caption{Performance comparison of Baseline CNN and ShallowFed (FOLC-Net) model: Domain Shift Test}
\begin{tabular}{ccccccc}
\hline
 \textbf{Method} & \textbf{Class} & \textbf{Precision} & \textbf{Recall} & \textbf{F1-Score} & \textbf{Accuracy} \\
\hline
 \multirow{2}{*}{ShallowFed CNN} & Colon-ACA & 99.50 & 99.10 & 99.30 & 99.30 \\
  & Colon-N & 99.10 & 99.50 & 99.30 & 99.30 \\

  \multirow{2}{*}{ShallowFed (FOLC-Net)} & Colon-ACA & 1.000 & 0.99 & 1.000 & 99.75 \\
  & Colon-N & 1.000 & 1.000 & 1.000 & 1.000 \\
\hline
 \multirow{2}{*}{ShallowFed CNN} & Tumor & 99.00 & 99.00 & 99.00 & 99.30 \\
  & Notumor & 99.46 & 99.46 & 99.46 & 99.46 \\

  \multirow{2}{*}{ShallowFed (FOLC-Net)} & Tumor & 99.01 & 1.000 & 99.50 & 99.65 \\
  & Notumor & 1.000 & 99.46 & 99.73 & 99.73 \\
\hline
\end{tabular}
\end{table}
The confusion matrices in fig 17 illustrate the performance of two models (ShallowFed and ShallowFed with ConvNeXt) on two datasets: one for colon cancer classification (Colon-ACA and Colon-N) and one for tumor classification (Tumor and Notumor). For the colon cancer dataset, the ShallowFed model (subfigure (a)) performs well but shows a slight misclassification, with 9 False positives for Colon-ACA and 5 False negatives for Colon-N. However, the ShallowFed with ConvNeXt model (subfigure (b)) improves slightly, misclassifying only 1 false positive for Colon-ACA and 4 false negatives for Colon-N.

In the MHC-CT tumor dataset, the ShallowFed CNN model (subfigure (c)) also shows an efficient performance with 99 True Positives and 1 False Positive for the Tumor class, and 185 True Negatives and 1 False Negative for Notumor. The ShallowFed with ConvNeXt model (subfigure (d)) performs even better, with 99 True Positives and 1 False Positive for Tumor, and 186 True Negatives and no False Negatives for Notumor. The confusion matrices indicate that while both models are performing well, the ConvNeXt-enhanced model provides a slight improvement in both datasets.
\begin{figure}[!h]
\centering
\begin{minipage}[]{0.40\textwidth}
  \centering
  \includegraphics[width = \textwidth]{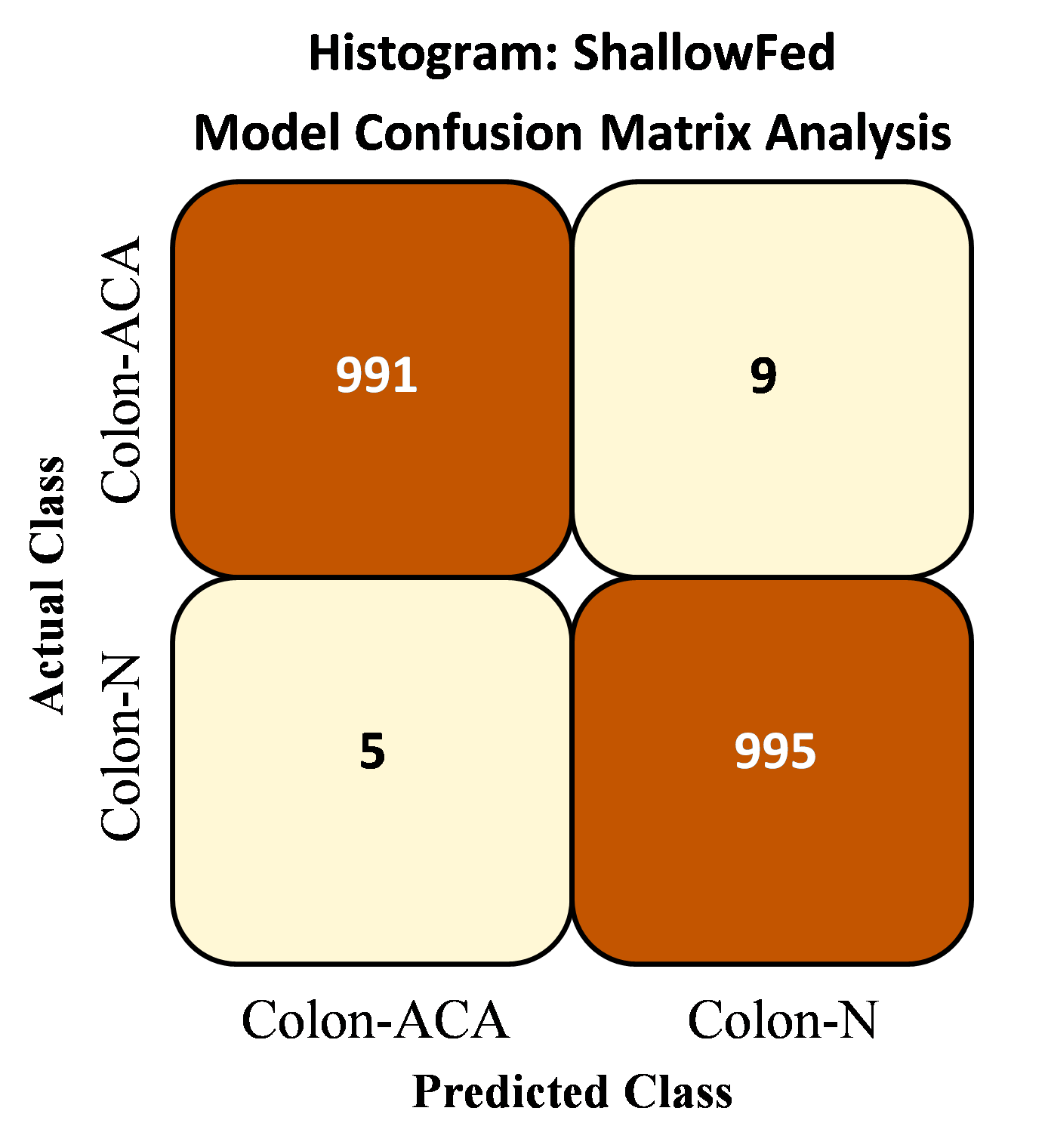}
  
    \begin{center}
     \textbf{a}    
     \end{center}
\end{minipage}
\begin{minipage}[]{0.40\textwidth}
  \centering
  \includegraphics[width = \textwidth]{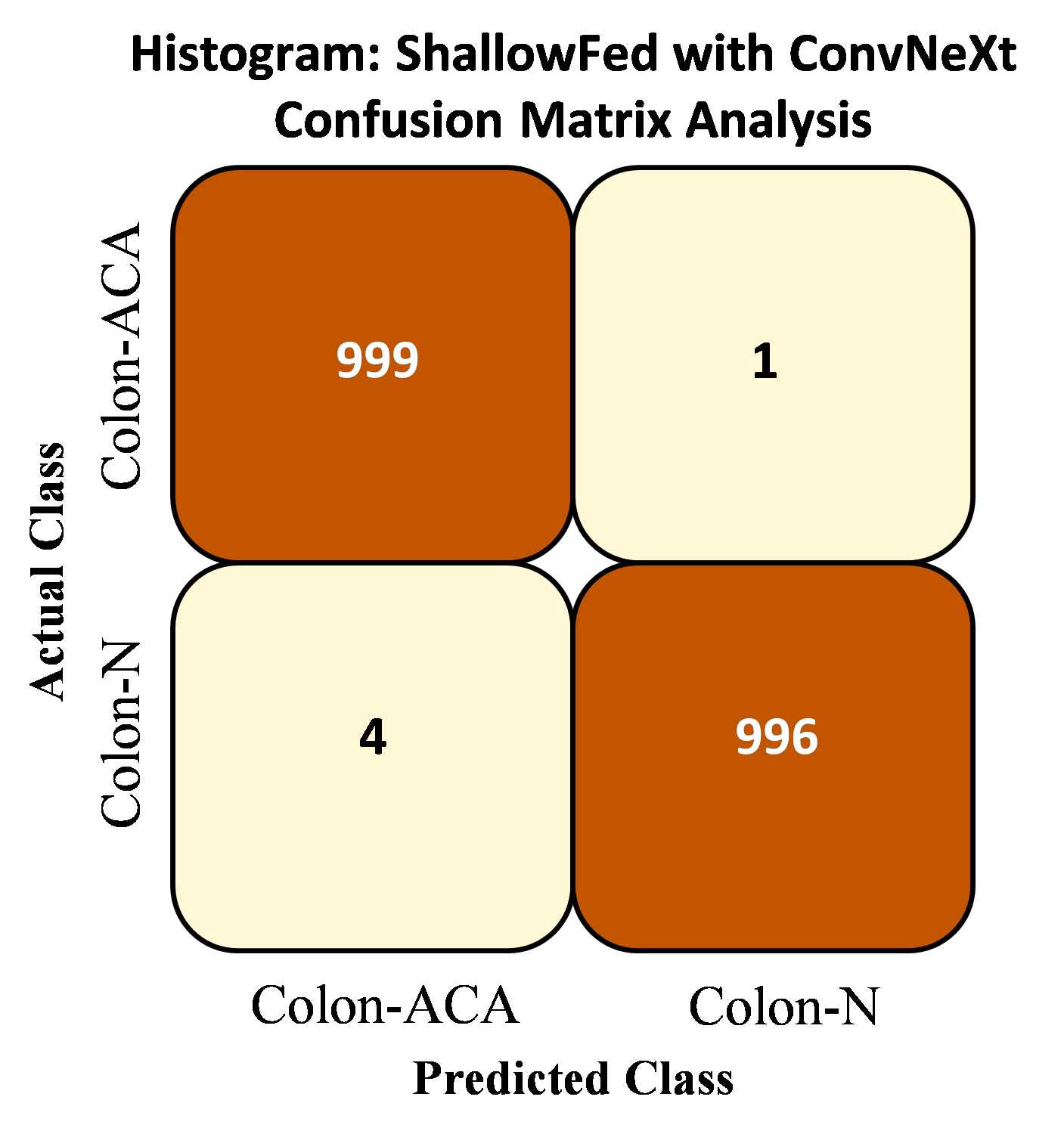}
 
     \begin{center}
     \textbf{b}    
     \end{center}
\end{minipage}
\begin{minipage}[]{0.40\textwidth}
  \centering
  \includegraphics[width = \textwidth]{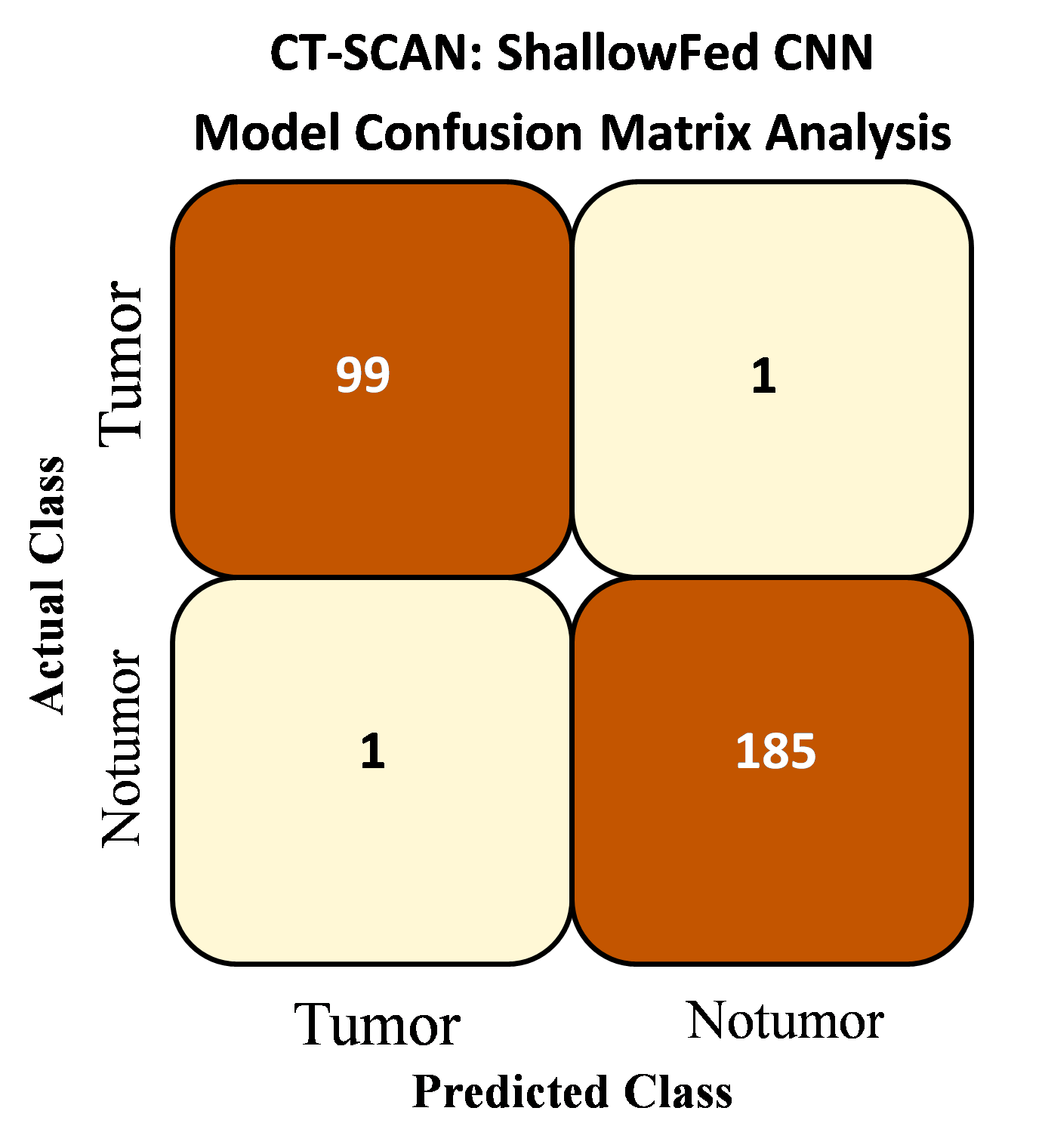}
  
    \begin{center}
     \textbf{c}    
     \end{center}
\end{minipage}
\begin{minipage}[]{0.40\textwidth}
  \centering
  \includegraphics[width = \textwidth]{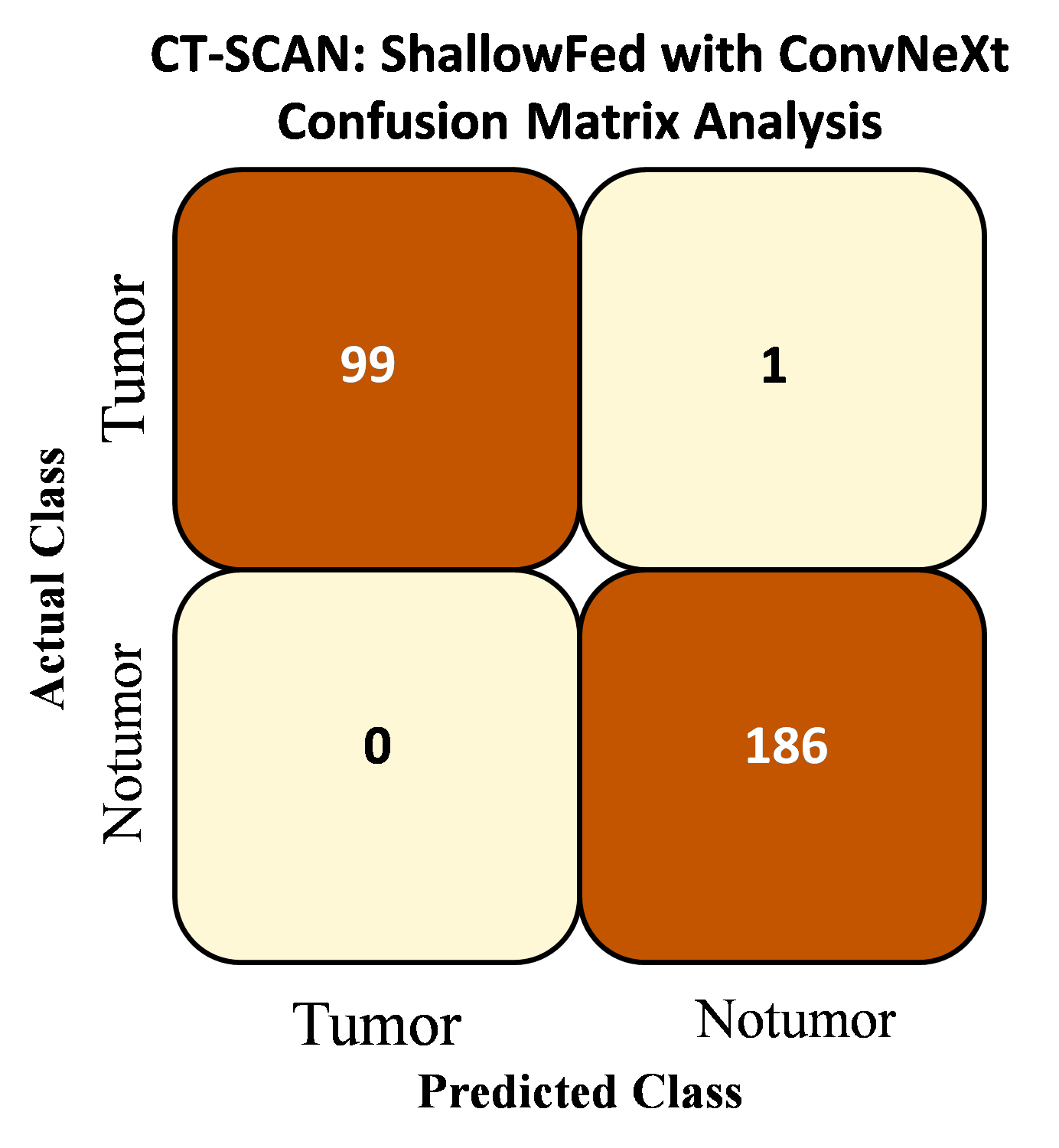}
 
     \begin{center}
     \textbf{d}    
     \end{center}
\end{minipage}
  \caption{An overview of the confusion matrices for Baseline CNN, and proposed ShallowFed with ConvNeXt (FOLC-Net): (a) Baseline CNN (Histogram), and (b) ShallowFed with ConvNeXt (FOLC-Net) (Histogram), (c) Baseline CNN (MHC-CT), and (d) ShallowFed with ConvNeXt (FOLC-Net) (MHC-CT)}
  \label{fig:11.png}
\end{figure}
The performance comparison between the ShallowFed (FOLC-Net) model and existing CNN models across two datasets Domain Shift Test (Table 13), Histopathological and MHC-CT, highlights the superior performance of ShallowFed (FOLC-Net). In the Histopathological dataset, proposed model achieves an accuracy of 99.75\%, significantly outperforming other models such as ResNet50 (87.35\%) and NASNetLarge (98.80\%). Similarly, for the MHC-CT dataset, ShallowFed (FOLC-Net) reaches an accuracy of 99.65\%, surpassing other models like MobileNetV2 (98.95\%) and ResNet101 (97.20\%). The model also excels in precision, recall, and F1-score, maintaining consistent performance across both datasets. These results demonstrate ShallowFed (FOLC-Net) robustness and ability to handle domain shifts effectively, outperforming conventional CNN architectures in both diagnostic accuracy and generalization to unseen data.

In terms of precision, recall, and F1-score, ShallowFed (FOLC-Net) stands out with an efficient 99.75\% score across all metrics in the Histopathological dataset, while in the MHC-CT dataset, it achieves 99.50\% precision, 99.73\% recall, and 99.61\% F1-score, showcasing its excellent capability in accurately detecting medical conditions with minimal errors. The model consistent high performance across both datasets ensures its generalization ability and adaptability to various medical imaging challenges, providing a reliable solution for real-world applications in healthcare diagnostics. This robust performance reinforces proposed potential in bridging domain gaps, making it a promising choice for clinical environments where precision and reliability are critical.
\begin{table}[h!]
\centering
\caption{Performance comparison of ShallowFed (FOLC-Net) and existing CNN models: Domain Shift Test}
\begin{tabular}{lllll}
\hline
\textbf{Model} & \textbf{Accuracy} & \textbf{Precision} & \textbf{Recall} & \textbf{F1-Score} \\
\hline
\multicolumn{5}{c}{\textbf{Histogram}} \\
\hline
Base- DensNet121 & 98.50 & 98.75 & 98.75 & 98.74 \\
Base- DensNet169 & 98.60 & 98.90 & 98.90 & 98.89 \\
Base- DensNet201 & 98.60 & 98.65 & 98.64 & 98.64 \\
Base- MobileNetV1 & 98.75 & 98.80 & 98.80 & 98.79 \\
Base- MobileNetV2 & 98.55 & 98.65 & 98.64 & 98.64 \\
Base- ResNet50 & 87.35 & 87.43 & 87.34 & 87.34 \\
Base- ResNet101 & 89.64 & 89.65 & 87.65 & 87.64 \\
Base- ResNet152 & 90.00 & 90.00 & 90.00 & 89.99 \\
Base-VGG16 & 98.50 & 98.70 & 98.75 & 98.74 \\
Base-VGG19 & 98.75 & 98.75 & 98.75 & 98.74 \\
Base- NASNetMobile & 98.80 & 98.40 & 98.80 & 98.79 \\
Base- NASNetLarge & 98.80 & 98.93 & 98.93 & 99.29 \\
ShallowFed (FOLC-Net) & 99.75 & 99.75 & 99.75 & 99.74 \\
\hline
\multicolumn{5}{c}{\textbf{MHC-CT}} \\
\hline
Base- DensNet121 & 98.25 & 98.73 & 98.50 & 98.61 \\
Base- DensNet169 & 98.60 & 98.76 & 98.50 & 98.65 \\
Base- DensNet201 & 96.15 & 95.04 & 97.04 & 95.86 \\
Base- MobileNetV1 & 98.95 & 98.73 & 98.73 & 98.99 \\
Base- MobileNetV2 & 98.95 & 98.73 & 99.50 & 98.99 \\
Base- ResNet50 & 98.25 & 98.69 & 97.50 & 98.05 \\
Base- ResNet101 & 97.20 & 97.64 & 96.23 & 96.87 \\
Base- ResNet152 & 98.60 & 98.46 & 98.46 & 98.46 \\
Base-VGG16 & 98.95 & 98.78 & 98.50 & 98.65 \\
Base-VGG19 & 98.60 & 98.59 & 98.65 & 98.55 \\
Base- NASNetMobile & 98.95 & 98.50 & 98.73 & 99.61 \\
Base- NASNetLarge & 98.60 & 98.23 & 98.23 & 98.23 \\
ShallowFed (FOLC-Net) & 99.65 & 99.50 & 99.73 & 99.61 \\
\hline
\end{tabular}
\label{tab:performance_comparison}
\end{table}
\section{Discussion: why FOLC-Net outperform Existing Models That Struggle in Some Views}
The performance of FL models is deeply influenced by both the architecture and the methods used to train these models, especially in decentralized settings with heterogeneous clients. Existing methodologies, while effective in handling combined multi-view representation, often experience significant performance degradation when tested on individual views. This phenomenon is particularly evident in the axial, coronal, and sagittal views, where the accuracy tends to drop compared to evaluations conducted on the combined views representation. These models typically struggle to adapt to the unique characteristics of individual views, as they rely on aggregated features or optimized primarily for multi-view representation, overlooking the specificity of each view challenges.

The FOLC-Net framework, however, addresses these limitations by incorporating Novel ShallowFed architectural which significantly enhance performance in both multi-view and individual view settings. The following phase contribute to FOLC-Net high performance, especially in scenarios where existing models underperform:
\begin{itemize}
    \item FOLC-Net uses MRFO to optimize the structure model by considering both global and local structure generation. Unlike existing SOTA, which uses a static single model structure, FOLC-Net adapts to individual view characteristics, ensuring optimal performance across all views.
    \item FOLC-Net scales efficiently by cloning the global model, distributing the training load across clients. This approach accelerates convergence and overcomes bottlenecks common in large-scale FL approaches, particularly for siloed data in axial, coronal, or sagittal views.
    \item By integrating ConvNeXt, FOLC-Net improves client device adaptability, especially in cases of sparse or view-specific data. It enhances performance across heterogeneous environments, addressing challenges in axial and sagittal views where SOTA models struggle.
    \item FOLC-Net employs a multi-view to single-view validation approach, evaluating both combined and individual view performance. Unlike traditional models, which overlook combined representation performance, FOLC-Net ensures a deeper understanding of the model's strengths and weaknesses across different views.
    \item FOLC-Net outperforms existing models on multi-views to single-view, particularly the challenging sagittal view. For instance, ShallowFed (FOLC-Net) achieves 92.44\% accuracy on the sagittal view, surpassing models like Khan et al. DL + Residual Learning (88.37\%) and Muneeb et al. DL (88.95\%). This is crucial for applications where specific views play varying roles.
\end{itemize}
\bmhead{Outcome: }
FOLC-Net overcomes the limitations of existing models that struggle with individual views by introducing mechanisms for efficient updates, model cloning for scalability, and the integration of ConvNeXt for enhanced client adaptability. Furthermore, its comprehensive validation approach ensures a more thorough understanding of its capabilities, enabling it to deliver superior performance across both multi-view and individual view scenarios. This makes FOLC-Net a robust solution for FL in diverse, real-world environments where client heterogeneity and view-specific challenges are common.
\section{Conclusion}\label{sec13}
In this work, the FOLC-Net framework is proposed as a novel solution to address the performance degradation observed in Federated Learning models when tested on individual medical imaging views, such as axial, coronal, and sagittal. By incorporating innovations like MRFO, global model cloning, and ConvNeXt integration, FOLC-Net significantly improves model performance across both multi-view and individual view scenarios. The experimental results demonstrate substantial accuracy improvements, particularly in the challenging sagittal view, showcasing the framework effectiveness in real-world medical image analysis tasks. FOLC-Net ability to enhance client adaptability and provide more tailored updates for individual views sets it apart from existing models. The comprehensive validation approach ensures that the performance under various conditions is thoroughly evaluated, making it a more reliable solution for decentralized healthcare applications. Additionally, FOLC-Net scalable architecture ensures that it can be efficiently deployed across heterogeneous clients in large-scale federated learning settings.

Future research will explore further optimization of FOLC-Net performance in more complex medical datasets and expand its application to other medical imaging modalities. Additionally, efforts will be directed towards enhancing model efficiency in large-scale federated environments to support real-time, edge-based medical diagnostics.

\backmatter

\section*{Declarations}
\bmhead{Ethics approval and consent to participate}
Approved and Not applicable
\bmhead{Consent for publication}
Not applicable
\bmhead{Data availability}
Data will be made available on request.
\bmhead{Funding}
No funding
\bmhead{Declaration of competing interest}
The authors declare that they have no known competing financial interests or personal relationships that could have appeared to influence the work reported in this paper.
\bmhead{CRediT authorship contribution statement}
Saif Ur Rehman Khan \& Muhammed Nabeel Asim: Conceptualization, Data curation, Methodology, Software, Validation, Writing original draft \& Formal analysis. Sebastian Vollmer: Conceptualization, Funding acquisition, Supervision. Andreas Dengel: review \& editing. 

\begin{appendices}

\section{Supplementary material}\label{secA1}

\begin{table}[h!]
\centering
\caption{Acronyms with Detail Form}
\begin{tabular}{ll}
\hline
\textbf{Acronyms} & \textbf{Detail Form} \\ \hline
TL  & Transfer Learning \\ 
CNN & Convolutional Neural Network \\ 
PR  & Precision-Recall \\ 
ML  & Machine Learning \\ 
FL  & Federated Learning \\ 
MRFO & Manta-ray foraging optimization \\ 
t-SNE & t-distributed Stochastic Neighbor Embedding \\ 
CHF & Chain Foraging \\ 
CYA & Cyclone Aging \\ 
CV  & Computer Vision \\ 
ROC & Receiver Operating Characteristic \\ 
AP  & Average Precision \\ \hline
\end{tabular}

\label{tab:acronyms}
\end{table}
\end{appendices}


\bibliography{sn-bibliography}

\end{document}